\documentclass{article} %
\usepackage{fix-cm}
\usepackage{iclr2027_conference,times}

\usepackage{amsmath,amsfonts,bm}

\def\eqref#1{equation~\ref{#1}}

\def\1{\bm{1}}

\DeclareMathAlphabet{\mathsfit}{\encodingdefault}{\sfdefault}{m}{sl}
\SetMathAlphabet{\mathsfit}{bold}{\encodingdefault}{\sfdefault}{bx}{n}

\usepackage{url}
\usepackage{graphicx}
\usepackage{booktabs}
\usepackage{float}
\usepackage{placeins}
\usepackage{tabularx}
\usepackage{xcolor}
\usepackage{wrapfig}
\usepackage{needspace}
\usepackage{colortbl}
\usepackage{comment}
\usepackage{longtable}
\usepackage{colortbl}

\newcommand{\sys}{CARVE}

\title{\sys{}: Breaking Data Barriers in Chip Placement by Harnessing Reusable Expertise}

\iclrfinalcopy

\author{Jiefu Zhang$^{1}$, Haixiang Sun$^{1}$, Yang Xu$^{1}$, Vaneet Aggarwal$^{1}$, Zishen Wan$^{2}$\\
$^{1}$Purdue University, West Lafayette, IN 47907, USA \\
$^{2}$Columbia University, New York, NY 10027, USA \\
\texttt{\{zhan4018, sun1321, xu1720, vaneet\}@purdue.edu} \\
\texttt{zishen.wan@columbia.edu}
}

\usepackage[T1]{fontenc}
\usepackage{xurl}
\usepackage{hyperref}
\usepackage{amsthm}
\usepackage{caption}
\DeclareCaptionFont{carvesmall}{\fontsize{7.5}{8.6}\selectfont}
\newtheorem{proposition}{Proposition}
\newtheorem{theorem}[proposition]{Theorem}
\begin{document}

\maketitle
\pagestyle{plain}

\begin{abstract}
Pretrained macro-placement policies can reduce repeated optimization across circuits, but deployment often exposes them to unfamiliar designs when the original training data are unavailable. Repeatedly fine-tuning a single serving model can overwrite earlier improvements, while simply saving checkpoints does not determine where they can be reliably reused. We introduce \underline{C}ontinual \underline{A}daptation through the \underline{R}euse of \underline{V}alidated \underline{E}xpertise (CARVE), a framework that represents accumulated expertise as a frozen base policy, immutable specialists, and task-specific credentials obtained through local validation. For a new task, CARVE first checks existing specialists and trains a new specialist from the frozen base only when none qualifies. Under fixed task distributions and validation rules that control cumulative error, we establish expected-performance guarantees for repeated reuse. For bounded losses, we also derive matching worst-case bounds on the local samples needed for reliable reuse. In macro placement, a reuse-first follow-up reduces recorded training time by 58.5\% (9.66 to 4.01 hours), while mean HPWL gain changes only from 8.41\% to 7.86\%. In a simulated receiving deployment, imported specialists are reused on six of seven new IBM circuits with no receiver-side training, achieving a 5.76\% mean HPWL gain. Navigation studies provide complementary evidence on repair retention and repeated adaptation.

\end{abstract}

\section{Introduction}
\label{sec:intro}

Macro placement in hardware design determines where large circuit blocks
are positioned in an application-specific integrated circuit (ASIC),
affecting wirelength and routing congestion. Conventional placement
uses commercial electronic design automation tools with input from
human experts and can require substantial time and
effort~\citep{lee2025chipdiffusion}. In contrast, learning-based approaches can reduce this effort by using circuit data and feedback on placement quality to train policies that select macro locations.
For example, \citet{mirhoseini2021graph} report generating a placement for a TPU block in six hours after pretraining, compared with several weeks for their human expert baseline. The learned placement also achieved shorter wirelength. However, realizing these efficiency gains in deployment depends on whether an organization can adapt and reuse the policy with the design data available to it.

Access to design data constrains how learned placement policies can be
reused. Circuit netlists and placement histories can contain proprietary
information, and licensing restrictions limit the data available for
learning~\citep{chai2022circuitnet}. Access to a pretrained placement
policy does not necessarily include access to its source training data.
For example, AlphaChip~\citep{google2024alphachip} provides a pretrained
checkpoint that other researchers have fine-tuned on their own circuit
benchmarks, despite incomplete availability of the original training
data~\citep{cheng2026assessment}. We consider deployments in which an
organization receives such a policy but cannot access the provider's
training circuits or placement histories. The organization can evaluate
and adapt the policy on its own circuits, but joint retraining or replay
requiring unavailable provider examples is not possible. Adaptation
using only new-task data nevertheless remains
possible~\citep{li2016lwf}. Within this data constraint, deployed policies
also face an out-of-distribution (OOD) challenge as new circuits arrive.
Differences in macro sizes, netlist connectivity, and placement
constraints may produce tasks poorly represented in the training data.
Such differences do not necessarily prevent a learned policy from
producing a good placement~\citep{chipformer}, but performance on earlier
circuits does not establish whether it will perform well on a new design.
An operator therefore needs to evaluate the available policy on each new
circuit and determine whether further adaptation is needed.
Methods such as ChiPFormer~\citep{chipformer} support adaptation through
additional placement trials and policy updates. However, their pretraining and fine-tuning procedure does not specify how to retain the
resulting adaptations, validate them for later circuits, or select among
saved models as requests arrive. The deployment problem therefore
extends beyond adapting to an individual circuit to accumulating local
improvements and reusing suitable models without the provider's training
data.

Together, restricted access to source data and unfamiliar circuits
motivate a continual-learning-based approach that adapts a pretrained
policy using local data while retaining earlier capabilities.
Specifically, placement trials on local circuits should support
adaptation to unfamiliar designs~\citep{chipformer}, while the approach
should also retain the resulting improvements for later reuse.
However, adaptation alone leaves several deployment requirements
unresolved. For instance, EWC~\citep{kirkpatrick2017ewc} and Synaptic
Intelligence~\citep{zenke2017si} use parameter importance statistics
from earlier learning that may not accompany pretrained weights.
Methods that freeze earlier models~\citep{rusu2016progressive} can
preserve their behavior, but freezing alone does not establish whether
those models can serve a new task without further training.
In addition, successively updating the shared model can change
parameters needed by earlier tasks, causing catastrophic
forgetting~\citep{kirkpatrick2017ewc}.
Although replay-based
approaches~\citep{lopez2017gradient} can mitigate
forgetting by revisiting earlier examples, provider training data are
unavailable in our setting and cannot be used for replay.
Another possible approach would be Learning without Forgetting, which
can operate without earlier training examples by matching the previous
model's predictions~\citep{li2016lwf} on current-task inputs.
However, placement states from a new circuit may poorly represent
conditions encountered on earlier circuits, leaving earlier behavior
insufficiently protected during adaptation. Finally, naively extended training
can lead to a higher risk of loss
of plasticity~\citep{dohare2024plasticity}, which impedes subsequent learning
even when new data are available~\citep{lyle2023plasticity}.
Neuron recycling, parameter resets, and regularization can mitigate
this decline using local data, but their effectiveness depends on the
network and training conditions. An intervention targeting one mechanism
may leave other causes
unresolved~\citep{dohare2024plasticity,lyle2025disentangling}.
A reliable solution therefore requires preserving accumulated expertise,
establishing where it can be reused, and supporting further learning
from local data.

Motivated by the above, we introduce Continual Adaptation through the
Reuse of Validated Expertise (CARVE), a framework that accumulates
adapted policies as specialists using a pretrained base and local task
data. We introduce a key concept called a specialist, which is
a copy of the pretrained policy adapted to a task using local data.
Each specialist is stored with task-specific authorizations, called
credentials, that record where it has passed evaluation against the base.
To retain earlier capabilities, CARVE keeps the base and accepted
specialists unchanged, so training for a new task cannot overwrite an
earlier specialist.
When an earlier task returns, its authorized specialist remains available
without reconstructing its training data. To avoid carrying accumulated
repair updates into later learning, each new specialist is trained from
a separate copy of the same frozen base. Its initial parameters therefore
remain independent of the sequence of earlier repairs, preventing
successive adaptations from progressively changing the starting point
for later learning. Training and validation use current-task data without
requiring provider examples, previous-task replay, or historical parameter
importance statistics.

To decide whether to reuse an existing specialist or train a new one,
CARVE first evaluates its saved specialists and the frozen base on
the same local cases for the current task.
A specialist that passes a predeclared paired comparison receives a
credential for that task under the chosen metric. The credential
authorizes subsequent use without changing the specialist, allowing
one model to serve several tasks after separate evaluations.
If no examined specialist qualifies, CARVE trains a new candidate
from the frozen base and accepts it only after it passes the same
comparison. Each request is served before the repair it triggers,
so new specialists and credentials affect later requests. Each task
receives at most one bounded training session per repair epoch, and
the base serves whenever no specialist is authorized.
An imported specialist follows the same validation procedure at the
receiving operator, where reuse requires local evaluation without access
to its source circuits or training histories. CARVE therefore combines
unchanged models with task-specific evidence for deciding whether to
reuse existing expertise or train a new specialist.
Our contributions are:
\begin{itemize}
\item \textbf{A continual-adaptation abstraction based on validated expertise.}
CARVE represents learned capability as immutable specialists with
task-specific authorizations, separating capability acquisition,
retention, reuse, and deployment decisions. It checks saved specialists
and starts bounded local training only when none qualifies, without
requiring provider training data or previous-task replay.
    \item \textbf{A statistical characterization of reliable reuse.} We establish conditions for preserving expected performance under repeated
reuse and matching worst-case sample bounds under bounded losses for a fixed library
of specialists.
    \item \textbf{Empirical evidence across macro placement and navigation.} We study repeated repair, cross-task reuse, task-local model selection, and simulated cross-operator transfer, quantifying both the optimization savings and the quality trade-offs that arise when previously learned expertise is reused instead of training a task-specific specialist.

\end{itemize}

\section{Continual adaptation with \sys{}}
\label{sec:system}

\begin{figure}[H]
\centering
\includegraphics[width=0.92\linewidth]{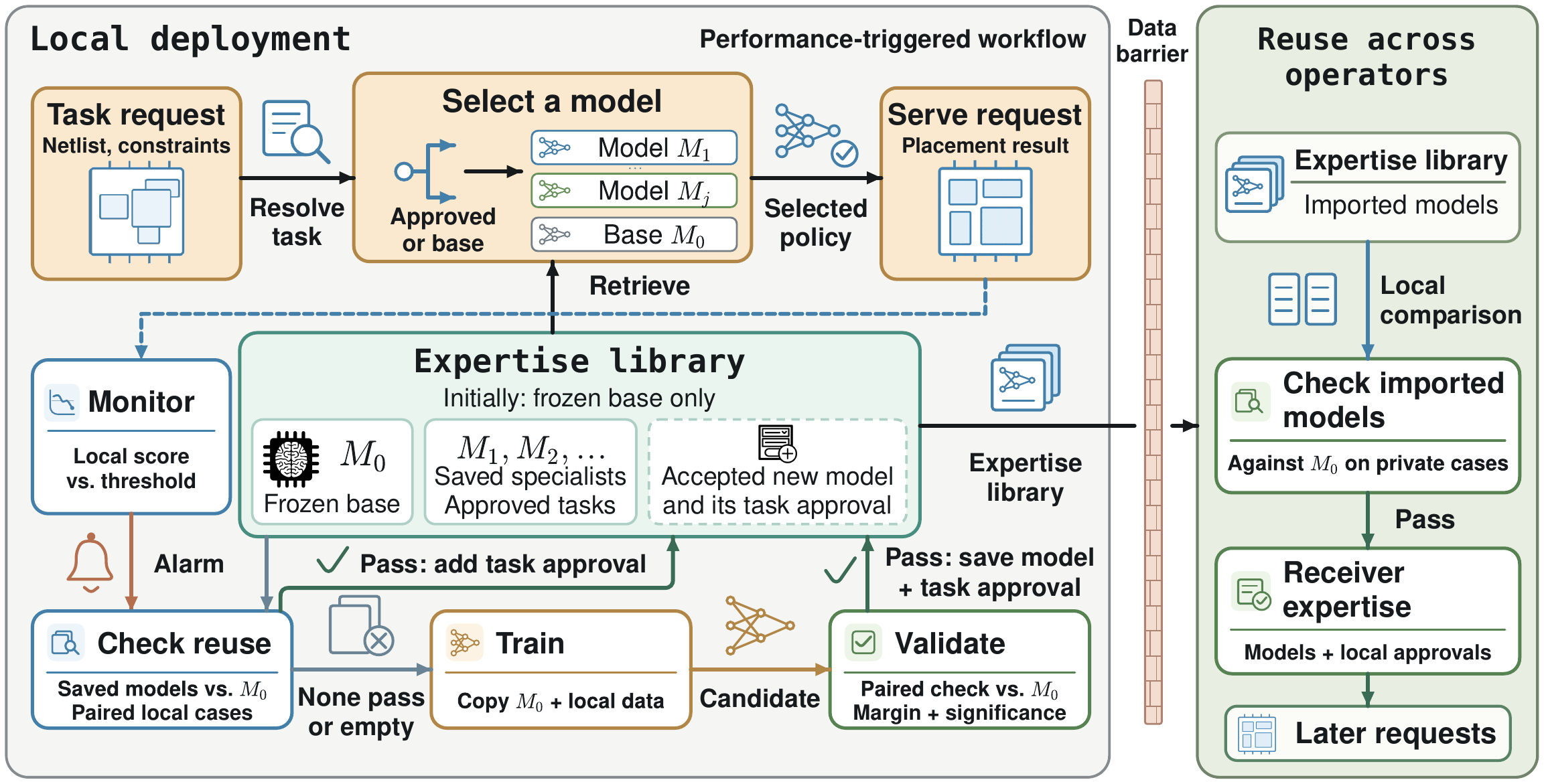}
\caption{Overview pipeline of \sys{}.}\vspace{-10pt}
\label{fig:state-machine}
\end{figure}

\sys{} is our approach for supporting continual adaptation in macro placement without provider
training data, shown in Figure~\ref{fig:state-machine}. Its library stores the
frozen base, unchanged specialists, and task approvals. Each request uses
an approved specialist or the base. When repair is triggered, local
evaluation can authorize saved specialists for later requests. If none
qualifies, training starts from the base, and only passing candidates
enter the library. Imported specialists undergo the same local checks,
enabling reuse across operators with current-task data alone.

\providecommand{\NavStudy}{Navigation streams}
\providecommand{\ChipStudy}{Chip streams}
\providecommand{\ReuseStudy}{Reuse before training study}
\providecommand{\TransferStudy}{Local transfer study}
\providecommand{\ForgetStudy}{A, B, A forgetting test}
\providecommand{\RecheckStudy}{Rejected model recheck}
\providecommand{\ReuseArm}{CARVE reuse first}
\providecommand{\FTCRTableStyle}{}
\renewcommand{\FTCRTableStyle}{\fontsize{7.5}{8.6}\selectfont
  \setlength{\tabcolsep}{3pt}\renewcommand{\arraystretch}{1.04}
  \setlength{\aboverulesep}{1.4pt}\setlength{\belowrulesep}{1.4pt}}
\providecommand{\FTCRDenseTableStyle}{}
\renewcommand{\FTCRDenseTableStyle}{\fontsize{7}{8.1}\selectfont
  \setlength{\tabcolsep}{2pt}\renewcommand{\arraystretch}{1.02}
  \setlength{\aboverulesep}{1.2pt}\setlength{\belowrulesep}{1.2pt}}

\FloatBarrier\vspace{-6pt}

\subsection{Retained models and permitted uses}
\label{sec:retained-models}

We consider a request $x_t$ that asks a policy to solve a task at arrival $t$.
For placement, it supplies a circuit's netlist, macro sizes, and placement
constraints, and the policy returns macro locations. Each request may contain several trials or episodes,
each called a case $z$. A family $f$ groups repeated requests for the
same task, and we denote $\mathcal{F}$ as the set that contains all the declared families.
Each placement family is one circuit with
its macro and netlist specifications.  The deterministic
resolver $\rho$ assigns $f_t=\rho(x_t)$, with $\bot$ denoting an unresolved
or out-of-scope request. A repair epoch fixes the base policy $M_0$,
resolver, metric, and contract $\mathcal{C}$, which specifies the trigger,
trainer, budget, evaluation rule, and model selector. We denote the expertise library as $\mathcal{M}_t=\{M_0,\ldots,M_{J_t}\}$ that contains the
frozen base $M_0$ and $J_t$ unchanged specialists, each a separately adapted
policy. The credential set $\Gamma_t$ records approvals
$(j,f)$ permitting $M_j$ to serve family $f$ under the epoch's base,
metric, and contract. Local evaluation for issuing a credential is
called admission. The fixed selector $\sigma(f,\Gamma_t)$ chooses
among approved specialists, and $g_t$ denotes the serving model.
Initially, the library contains only $M_0$ and has no credentials.
The set $\mathcal{T}_t$ records families whose new candidate received a
complete rejection. The set $\mathcal{X}_t$ records families whose entire
training session exhausted its allowed attempts without producing a candidate.
Eligible families are resolved families with no credential and no stopped
status. Monitoring cases are denoted
by $E_t$, and local admission cases by $D_f^{\mathrm{dev}}$.
The score $s_f(M,z)$ evaluates model $M$ on case $z$ and is oriented
so that larger values are better. Placement uses an equivalent rule
favoring lower half-perimeter wirelength (HPWL), an estimate of wiring length.

Given $x_t$, the resolver supplies the family key used to retrieve an
approved specialist. The selector returns $g_t$, which produces the
requested output before any resulting repair. Unresolved or stopped
families, and families without approval, would still use $M_0$. Returning tasks can
therefore reuse an unchanged specialist without reconstructing earlier
training data or repeating validation. Since selection is restricted to
evaluated pairs $(j,f)$, approval on one circuit does not
authorize service on another. After the base service for an eligible family, the monitor applies
$\mathcal{C}$ to $E_t$ and returns a repair decision. A trigger then starts an extension
exam, which compares saved specialists with $M_0$ on the same local
cases using the admission rule in Section~\ref{sec:admission}.
The contract fixes the comparison set, order, and multiplicity correction.
A pass adds $(j,f)$ to $\Gamma_t$ without changing weights, extending
earlier expertise to another family. Training begins when no examined
specialist passes or the comparison set is empty. Newly issued
credentials affect only later requests.
The main placement study instead trains each new circuit before logging
extensions, which neither avoid training nor change routing. The
follow-up evaluates checking before training. The receiver takes imported specialists and evaluates them against its
own base on private target cases. A pass creates a local credential
for later service, using the same extension procedure. Only the models need
to cross the data barrier, while source circuits and training histories remain
unnecessary. Source credentials cannot replace local evaluation, and
the receiver study performs no training and uses its base if none passes.

\subsection{Training and admitting a new repair}
\label{sec:candidate}\label{sec:admission}\label{sec:commit}

During training, for family $f$, let $D_f^{\mathrm{tr}}$ denote local training data,
$\mathsf{Train}_f$ the trainer with fixed settings $\xi_f$, and $C_f$
its candidate policy. A session permits at most $r$ training attempts.
Admission compares $C_f$ with $M_0$ on $D_f^{\mathrm{dev}}$ using observed
improvement $\widehat\Delta_f$ and required margin $m_f$. When a test is
used, $p_f$ denotes its $p$-value and $\alpha_f$ its declared threshold. Let $P_f$ be the fixed distribution of family $f$'s
cases, including task settings and execution randomness. A fixed loss
$\ell_f(M,z)\in[0,1]$ defines expected loss
$R_f(M)=\mathbb{E}_{Z\sim P_f}[\ell_f(M,Z)]$ and expected improvement
$\Delta_f(M)=R_f(M_0)-R_f(M)$. The score $s_f=1-\ell_f$ connects this
analysis to paired evaluation. Let $k$ index validation rounds,
$\mathcal{G}_{k-1}$ denote their prior history, and $\alpha_k$ their
error allowance chosen from that history. The event $B_k$ occurs when
round $k$ issues any credential with $\Delta_f(M_j)<0$; $\delta$ is
the total error tolerance.

For one target family, fix $J\ge1$ specialists before drawing $n$
independent cases $Z_1,\ldots,Z_n$ from $P_f$.
Write $\Delta_j=\Delta_f(M_j)$, $\Delta_0=0$, and
$\widehat\Delta_j$ for the sample mean paired improvement.
A validator selects $\widehat j\in\{0,\ldots,J\}$, where zero denotes
the base. Reliable reuse requires nonnegative selected improvement and
selection of a specialist whenever some specialist improves by at least
$\gamma>0$, with probability at least $1-\delta$. Let $n^\star$ be the
smallest fixed sample size meeting this requirement uniformly over all
libraries and target distributions with bounded losses. The population
threshold $\gamma$ differs from the empirical admission margin $m_f$.

\noindent\textbf{Train.}
Training computes $C_f=\mathsf{Train}_f(M_0,D_f^{\mathrm{tr}};\xi_f)$
when the reuse check finds no passing specialist. Each candidate starts
from a separate copy of the base, so earlier repairs leave its initial
parameters and all saved models unchanged. Each family receives at most
one bounded session per epoch. Retries are permitted only when no
candidate is produced, never after rejection. Training uses current-family
data without provider examples or replay from previous families.

\noindent\textbf{Validate and retain.}
The candidate and base receive the same admission cases, and their
scores determine whether the candidate passes the declared rule.
A contract using both a margin and a test requires
$\widehat\Delta_f\ge m_f$ and $p_f\le\alpha_f$, and a margin-only contract
omits the test, while the margins were selected during
development. The protocol requires separate training and admission data;
the navigation records do not establish disjointness.
A pass saves $C_f$ as $M_{J_t+1}$ together with credential $(J_t+1,f)$
and its evaluation evidence before later service. Rejection adds $f$ to
$\mathcal{T}_t$, while training exhaustion adds it to $\mathcal{X}_t$.
Both retain the base for the epoch.
Incomplete evidence authorizes neither service nor a performance rejection. Admission thus separates
learning a candidate from permitting its use, while preservation retains
each accepted repair. The following results explain how unchanged specialists support repeated
reuse and when local evaluation can authorize reuse before training.

\begin{proposition}[Expected performance under repeated reuse]
\label{prop:retained-risk}
Suppose the base, stored specialists, inference procedures, losses, and
family distributions remain fixed within an epoch. Start with no
credentials and issue each through validation. For $0<\delta<1$, assume
$\Pr(B_k\mid\mathcal{G}_{k-1})\le\alpha_k$ and
$\sum_{k\ge1}\alpha_k\le\delta$ almost surely. If $g_t$ is the base or
a specialist credentialed for $f_t$, then
\begin{equation}
 \Pr\!\left[
 R_{f_t}(g_t)\le R_{f_t}(M_0)
 \text{ for every in-scope request }t
 \right]\ge1-\delta.
 \label{eq:retained-risk}
\end{equation}
\end{proposition}

\noindent Proof sketch.
A conditional union bound gives
$\Pr(\bigcup_k B_k)\le\mathbb{E}[\sum_k\alpha_k]\le\delta$.
Otherwise every credential has nonnegative expected improvement.
Unchanged models and distributions preserve that comparison, and
selection uses the corresponding credential. Fallback gives equality.
Appendix~\ref{app:theory-retention} gives the proof.
New credentials consume the error budget; repeated use does not.
The guarantee concerns expected loss, allowing variation across cases
and between successive specialists. An unchanged family key alone does
not ensure an unchanged distribution.

\noindent\textbf{Local evaluation for reliable reuse.}
The requirement defined above is
\small
\begin{equation}
 \Pr\!\left[
 \Delta_{\widehat j}\ge0\ \text{and}\
 \left(\max_{1\le j\le J}\Delta_j\ge\gamma
       \ \Longrightarrow\ \widehat j\ne0\right)
 \right]\ge1-\delta.
 \label{eq:reliable-reuse}
\end{equation}
\normalsize
The validator learns the target distribution only through the paired
losses on these cases; the library and $n$ are fixed beforehand.

\begin{theorem}[Local evaluation for reliable reuse]
\label{thm:local-reuse}
For $J\ge1$ and $0<\gamma,\delta\le1/4$, we have
\small
\begin{equation}
 \frac{\log(J/\delta)}{64\gamma^2}
 \le n^\star \le
 \left\lceil\frac{8\log(2J/\delta)}{\gamma^2}\right\rceil.
 \label{eq:reuse-complexity}
\end{equation}
\normalsize
Thus $n^\star\!=\!\Theta(\gamma^{-2}\log(J/\delta))$, no source training
data or parameter updates needed for upper bound.
\end{theorem}

\noindent Proof sketch.
The paired estimate is
$\widehat\Delta_j=n^{-1}\sum_{i=1}^n
[\ell_f(M_0,Z_i)-\ell_f(M_j,Z_i)]$. Authorize a specialist when
$\widehat\Delta_j\ge\sqrt{2\log(2J/\delta)/n}$.
Hoeffding's inequality and a union bound control all paired estimates
simultaneously. At the stated sample size, every authorized specialist
has nonnegative expected improvement and every specialist improving by
at least $\gamma$ passes, with probability at least $1-\delta$.
Appendix~\ref{app:theory-reuse} shows  the lower bound follows by distinguishing unknown target distributions
that make different fixed specialists useful from a distribution where
all specialists are harmful.

Theorem~\ref{thm:local-reuse} explains how local evidence can authorize
saved or imported expertise before training. It counts cases, while
evaluation can require $(J+1)n$ executions. The selected specialist need
not be best, improve by $\gamma$, or match a newly trained model.
Both results require representative fresh validation and error control
across credential decisions. The experimental navigation and
median-HPWL gates do not directly instantiate the bounded-loss rule.
Appendix~\ref{app:theory-application} gives a concrete rule and its scope.

\section{Evaluation design}
\label{sec:eval}\label{sec:results}
\label{sec:eval-questions}\label{sec:eval-rules}\label{sec:eval-measures}

We first use crowd navigation as a controlled numerical study of CARVE, followed by our main chip-placement evaluation.
Both domains involve a pretrained policy, recurring tasks, and local
feedback for evaluating repairs. Navigation lets us vary scene conditions
and request order explicitly to examine retention, admission, and reuse.
Within each domain, methods share a pretrained base and the same recorded
request order, but maintain separate repair states. Requests are served
before repair, so updates affect later requests.
Appendix~\ref{app:domain-contracts} gives the domain-specific rules.

Prior work evaluates candidates using an empirical promotion threshold
\citep{silver2017mastering} or an explicit performance requirement with
statistical error control \citep{thomas2015hcpi}. We selected CARVE's
domain margins from development experiments. Navigation requires an
increase of at least 15 percentage points in swept success on 100 paired
cases. Placement requires at least a 5\% relative reduction in median
HPWL on 30 paired trials, together with the declared paired test.
The margins specify the minimum observed improvement required for
admission.

\subsection{Crowd navigation}
\label{sec:nav-study}\label{sec:eval-nav}\label{sec:nav-batches}
\begingroup
\setlength{\intextsep}{4pt plus 1pt minus 1pt}
\setlength{\abovecaptionskip}{3pt}
\setlength{\belowcaptionskip}{1pt}

Navigation instantiates Section~\ref{sec:system} with scene families in
place of circuits. A family $f$ specifies geometry and the pedestrian model,
while a request $x_t$ specifies scene size and pedestrian count.
A case $z$ also fixes episode randomness. The frozen base $M_0$ is the
pretrained SARL policy~\citep{chen2019crowd} in
SICNav~\citep{samavi2025sicnav}. New candidates use only current-family
simulations. Credentials $(j,f)$ authorize unchanged specialists for a
family, retaining the same selection, reuse, and fallback rules.

Each run contains 30 requests of 100 matched cases. Success requires
reaching the goal within 90 seconds without swept-volume contact.
Measured success triggers repair, and saved specialists are checked before
training. Figure~\ref{fig:nav-scenes} shows the scene families and the
first requests in the fixed-arrival sequence.
\begin{figure}[H]
\centering
\includegraphics[width=0.92\linewidth,trim=0 17bp 0 3bp,clip]{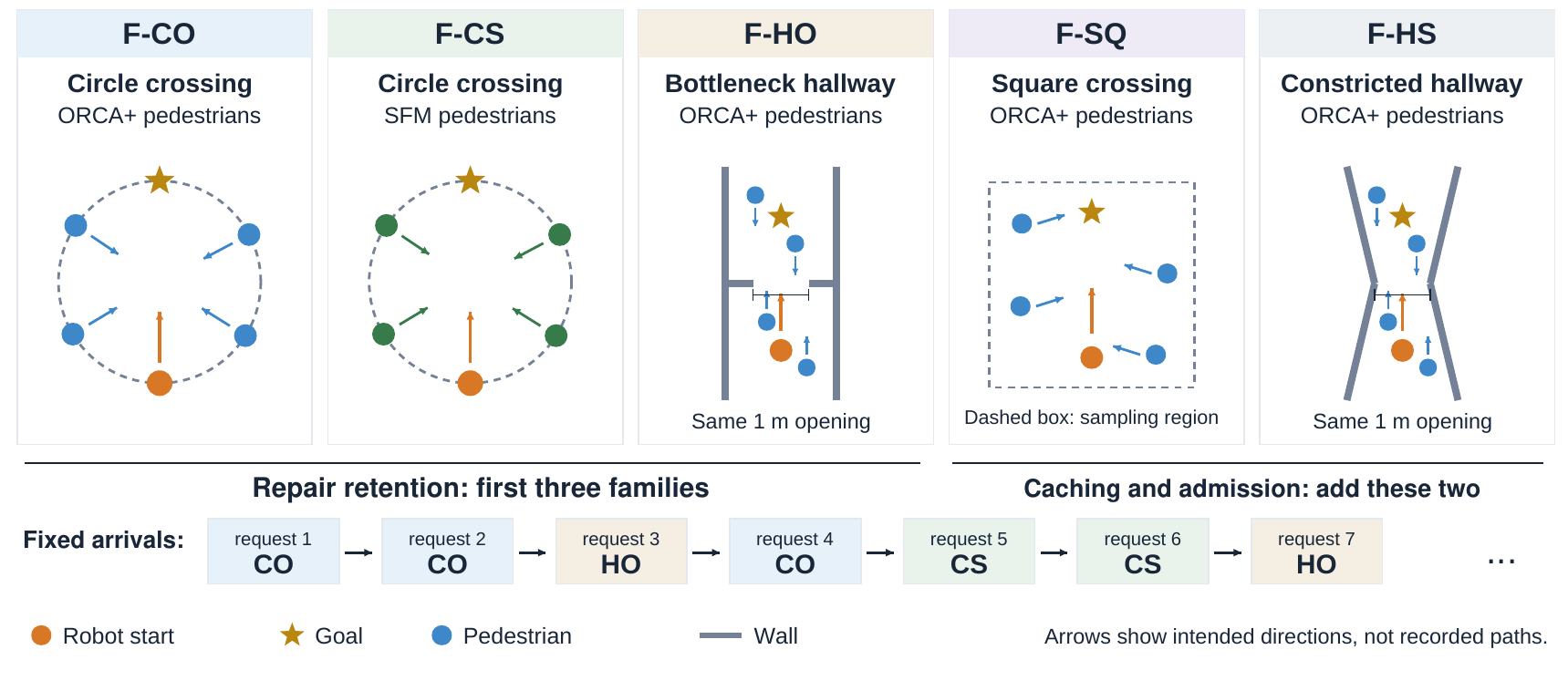}
\caption{Navigation scene families and the fixed-arrival request stream.}
\label{fig:nav-scenes}
\end{figure}
The repair retention study uses circle crossing with ORCA+ or SFM
pedestrians~\citep{vandenberg2011orca,helbing1995social,zhang2026dontfreezedontcrash} and an ORCA+
bottleneck hallway. Frozen keeps the base unchanged, Overwrite repeatedly
fine-tunes and replaces one active policy, and \sys{} preserves accepted
specialists. Appendix~\ref{app:deployment-methods} gives the complete
rules. Fixed arrivals follow a predefined request list, while sampled
arrivals draw requests from the same pool of scene settings. The caching and admission study adds square crossing and a constricted
hallway. It samples request order and scene settings while fixing the
number of requests from each family. Cache-only retains a model for each
family without an admission check. Adapted CompoNet
\citep{malagon2024componet} provides a modular continual-learning comparison
on separate case blocks with the same arrival order and three root seeds.
Neither method was evaluated on the fixed-arrival or sampled-arrival
repair retention streams. Complete workload and comparison settings are in
Appendix~\ref{app:nav-stream-records}.
\begingroup
\setlength{\columnsep}{10pt}
\setlength{\intextsep}{4pt plus 1pt minus 1pt}
\setlength{\emergencystretch}{1em}
\Needspace{9\baselineskip}
\noindent
\begin{wraptable}{r}{0.60\textwidth}
\vspace{-5pt}
\centering
\begingroup
\caption{Repair retention with fixed and sampled arrivals.}
\label{tab:nav-retention}\label{tab:nav-main}
\fontsize{7}{8.2}\selectfont
\setlength{\tabcolsep}{1.0pt}
\renewcommand{\arraystretch}{1.04}
\setlength{\aboverulesep}{1.2pt}
\setlength{\belowrulesep}{1.2pt}
\def\FTCRHeader#1{{\fontsize{6.5}{7.5}\selectfont\shortstack{#1}}}
\begin{tabular*}{\linewidth}{@{\extracolsep{\fill}}l*{5}{r}@{\hspace{4pt}}*{5}{r}@{}}
\toprule
& \multicolumn{5}{c}{Fixed arrivals} & \multicolumn{5}{c}{Sampled arrivals}\\
\cmidrule(lr){2-6}\cmidrule(l){7-11}
Method
& \FTCRHeader{Success\\(\%)} & \FTCRHeader{Train\\count} & \FTCRHeader{Train\\(h)} & \FTCRHeader{Checks\\(h)} & \FTCRHeader{Repair\\(h)}
& \FTCRHeader{Success\\(\%)} & \FTCRHeader{Train\\count} & \FTCRHeader{Train\\(h)} & \FTCRHeader{Checks\\(h)} & \FTCRHeader{Repair\\(h)}\\
\midrule
Frozen & 35.5 & 0 & 0.00 & 0.00 & 0.00 & 34.3 & 0 & 0.00 & 0.00 & 0.00\\
Overwrite & 59.1 & 21 & 13.09 & 0.00 & 13.09 & 70.5 & 22 & 10.84 & 0.00 & 10.84\\
CARVE & \textbf{67.3} & 2 & 1.39 & 0.34 & \textbf{1.73} & \textbf{71.9} & 2 & 1.30 & 0.29 & \textbf{1.60}\\
\bottomrule
\end{tabular*}
\endgroup
\vspace{-3pt}
\end{wraptable}
\noindent\textbf{Repair retention.}
CARVE reduces repeated training and has higher success than Overwrite under both arrival constructions in Table~\ref{tab:nav-retention}. Table~\ref{tab:nav-arrivals-first15} shows that, with fixed arrivals, it saves separate circle and hallway specialists, retrieves the circle model when its task returns, and extends it to SFM without retraining. Repeated hallway
repairs account for most of Overwrite's cost, but CARVE's hallway success stays near the base. Repair time includes training and all admission and extension checks, including failures, but excludes ordinary service.
\par
\WFclear
\smallskip
\Needspace{10\baselineskip}
\noindent
\begin{wraptable}{r}{0.45\textwidth}

\centering
\begingroup
\caption{Caching and admission with fixed family counts.}
\label{tab:nav-caching-admission}
\fontsize{7}{8.2}\selectfont
\setlength{\tabcolsep}{1.0pt}
\renewcommand{\arraystretch}{1.04}
\setlength{\aboverulesep}{1.2pt}
\setlength{\belowrulesep}{1.2pt}
\def\FTCRHeader#1{{\fontsize{6.5}{7.5}\selectfont\shortstack{#1}}}
\begin{tabular*}{\linewidth}{@{\extracolsep{\fill}}l*{5}{r}rr@{}}
\toprule
Method & \FTCRHeader{Success\\(\%)} & \FTCRHeader{Train\\count} & \FTCRHeader{Train\\(h)} & \FTCRHeader{Checks\\(h)} & \FTCRHeader{Repair\\(h)}
& \FTCRHeader{Storage\\files / MiB} \\
\midrule
Frozen & 40.9 & 0 & 0.00 & 0.00 & 0.00 & --- \\
Overwrite & 57.9 & 25 & 14.16 & 0.00 & 14.16 & ---\\
CompoNet & 54.4 & 5 & 2.49 & 0.00 & 2.49 & ---\\
Cache-only & 57.9 & 5 & 2.85 & 0.00 & \textbf{2.85} & 5 / 1.872 \\
CARVE & \textbf{59.8} & 3 & 2.23 & 0.73 & 2.96 & \textbf{4 / 1.498}\\
\bottomrule
\end{tabular*}
\endgroup
\end{wraptable}
\noindent\textbf{Caching and admission.}
CARVE has higher success and lower checkpoint storage than Cache-only on the displayed stream, while Cache-only has lower repair time in Table~\ref{tab:nav-caching-admission}. Checks add evaluation work, and the additional arrival stream gives a different quality comparison. Cache-only uses 25\% more storage under the same six-entry limit. Extra storage is $100(B/B_{\mathrm{CARVE}}-1)$, using unrounded registered-file sizes. Storage measures checkpoint files, not RAM or GPU memory.
\par
\WFclear
\endgroup

\label{sec:nav-rq2}\label{sec:rq2}\label{sec:eval-rq2}

\begin{table}[h]
\centering
\begingroup
\caption{Fixed arrivals, requests 1 to 15 of 30. Success is out of 100 cases;
repair cost includes training and checks after each request.}
\label{tab:nav-arrivals-first15}
\fontsize{7}{8.2}\selectfont
\setlength{\tabcolsep}{1.4pt}
\renewcommand{\arraystretch}{1.15}
\setlength{\aboverulesep}{1.2pt}
\setlength{\belowrulesep}{1.2pt}
\definecolor{NavTraceCO}{HTML}{E7F0F8}
\definecolor{NavTraceCS}{HTML}{EAF3EC}
\definecolor{NavTraceHO}{HTML}{F4EDE3}
\definecolor{NavTraceBand}{HTML}{F3F5F8}
\begin{tabular}{@{}>{\raggedright\arraybackslash}p{.18\linewidth}*{15}{>{\centering\arraybackslash}p{\dimexpr(\linewidth-.18\linewidth-30\tabcolsep)/15\relax}}@{}}
\toprule
\textbf{Request} & \textbf{1} & \textbf{2} & \textbf{3} & \textbf{4} & \textbf{5} & \textbf{6} & \textbf{7} & \textbf{8} & \textbf{9} & \textbf{10} & \textbf{11} & \textbf{12} & \textbf{13} & \textbf{14} & \textbf{15}\\
\midrule
Family & \cellcolor{NavTraceCO}\textbf{CO} & \cellcolor{NavTraceCO}\textbf{CO} & \cellcolor{NavTraceHO}\textbf{HO} & \cellcolor{NavTraceCO}\textbf{CO} & \cellcolor{NavTraceCS}\textbf{CS} & \cellcolor{NavTraceCS}\textbf{CS} & \cellcolor{NavTraceHO}\textbf{HO} & \cellcolor{NavTraceCO}\textbf{CO} & \cellcolor{NavTraceCO}\textbf{CO} & \cellcolor{NavTraceCS}\textbf{CS} & \cellcolor{NavTraceHO}\textbf{HO} & \cellcolor{NavTraceCO}\textbf{CO} & \cellcolor{NavTraceCO}\textbf{CO} & \cellcolor{NavTraceHO}\textbf{HO} & \cellcolor{NavTraceCS}\textbf{CS}\\
Size $r[w]$ & 2 & 2.25 & 1.75 & 2 & 1.375 & 1.375 & 1.75 & 1.5 & 1.25 & 2.25 & 1.75 & 2.5 & 1.375 & 1.75 & 1.375\\
Pedestrians & 5 & 4 & 5 & 5 & 4 & 5 & 5 & 5 & 2 & 4 & 4 & 3 & 5 & 5 & 4\\
\midrule
\rowcolor{NavTraceBand}
Frozen success & 20 & 11 & 27 & 22 & 80 & 74 & 26 & 35 & 51 & 36 & 39 & 7 & 46 & 28 & 78\\
Overwrite success & 20 & 97 & 18 & 20 & 96 & 91 & 28 & 69 & 89 & 92 & 26 & 88 & 45 & 13 & 94\\
\rowcolor{NavTraceBand}
\sys{} success & 20 & 97 & 27 & 81 & 80 & 96 & 28 & 64 & 94 & 95 & 40 & 92 & 65 & 29 & 98\\
\midrule
Overwrite cost (min) & 20.2 & 0.0 & 63.0 & 20.8 & 0.0 & 0.0 & 64.5 & 21.0 & 0.0 & 0.0 & 55.3 & 14.8 & 21.2 & 65.4 & 0.0\\
\sys{} cost (min) & 26.2 & 0.0 & 76.4 & 0.0 & 1.5 & 0.0 & 0.0 & 0.0 & 0.0 & 0.0 & 0.0 & 0.0 & 0.0 & 0.0 & 0.0\\
\bottomrule
\end{tabular}
\endgroup
\end{table}\vspace{-5pt}
Table~\ref{tab:nav-arrivals-first15} shows the circle specialist serving
later circle requests after hallway adaptation and receiving approval for
SFM without retraining. Navigation therefore tests the retention and reuse
decisions that the placement studies next examine on circuit workloads.

\endgroup
\subsection{Chip macro placement}
\label{sec:chip-study}\label{sec:eval-chip}\label{sec:results-chip}\label{sec:chip-rq1}
\begingroup
\setlength{\intextsep}{4pt plus 1pt minus 1pt}
\setlength{\abovecaptionskip}{3pt}
\setlength{\belowcaptionskip}{1pt}

Our main application tests whether CARVE can retain and reuse local
placement repairs without the provider's training data. We use the released
ChiPFormer checkpoint~\citep{chipformer} and adapt it using only the current
circuit, without using the provider's training dataset or placement histories.
The experiments follow how a local repair is retained when its circuit
returns, reused for a different circuit, and then used by a simulated
receiving operator. We measure whether each form of reuse improves placement with less new
optimization, while respecting the source-data restriction.

\paragraph{Adapting locally and retaining repairs.}
We first test whether local adaptation can improve later placement without
repeating the optimization whenever a circuit returns. The circuit revisit
study contains 24 requests over eight circuits, each appearing three times.
Three circuits are in the reported pretraining corpus and five are not.
Each request provides macro sizes, connections, and a placement region;
the policy assigns macro positions. Figure~\ref{fig:chip-layouts-stream}
illustrates the placement task and the arrival sequence, including a circuit
returning after other circuits have been processed.

Frozen, Overwrite, RS, Cache-only, and \sys{} start from the same checkpoint
and process the same order with 30 matched service seeds per request.
Overwrite updates one active policy; Cache-only saves a model for each
circuit; RS checks candidates but uses one active model across circuits.
\sys{} retrieves a saved model for a circuit on which it has passed
evaluation. Each request is served before any resulting repair, so a new
model can improve only later requests. In the main study, \sys{} gives
each new circuit its own training session. Extensions recorded afterward
do not skip training; the next experiment tests that use of extension.
Appendix~\ref{app:domain-contracts} gives the full deployment rules. Half-perimeter wirelength (HPWL) estimates wirelength. Gain is the percentage reduction in a request's median HPWL relative to
Frozen. Mean gain and its standard deviation summarize the 24 requests; mean rank
averages each method's placement in the per-request HPWL ranking.
Repair time includes training and all checks, including failures, but
excludes ordinary service. 

\begin{figure}[h]
\centering
\includegraphics[width=.94\linewidth,trim=6bp 5bp 6bp 0bp,clip]{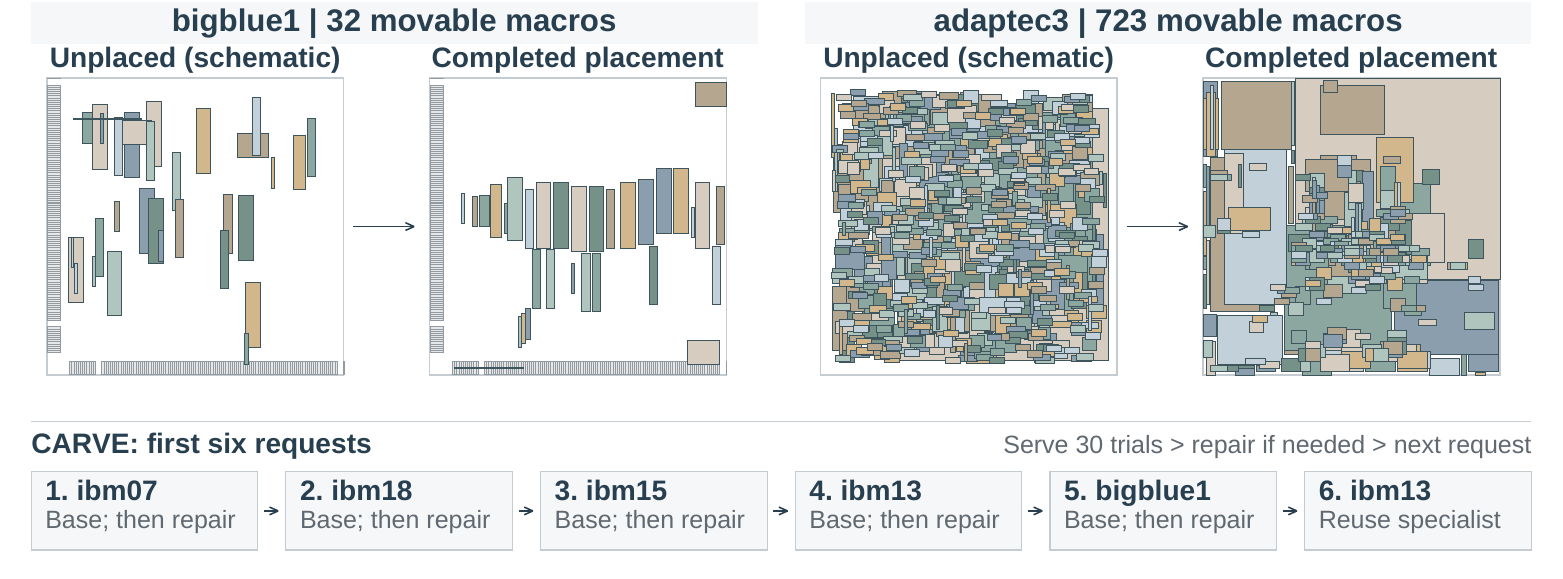}
\caption{Macro-placement examples and the first six requests in the circuit revisit study.
Scattered layouts are schematic; completed layouts are archived frozen-base
outputs, not \sys{} repair results.}
\label{fig:chip-layouts-stream}
\end{figure}

\begingroup
\setlength{\columnsep}{10pt}
\setlength{\intextsep}{4pt plus 1pt minus 1pt}
\setlength{\emergencystretch}{1em}
\Needspace{10\baselineskip}
\noindent
\begin{wraptable}{r}{0.58\textwidth}
\vspace{-5pt}
\centering
\begingroup
\FTCRDenseTableStyle
\caption{Placement quality and repair time over 24 requests.}
\label{tab:chip-main}\label{tab:chip-work-compact}
\fontsize{7}{8.1}\selectfont
\setlength{\tabcolsep}{2pt}
\renewcommand{\arraystretch}{1.08}
\begin{tabular*}{\linewidth}{@{\extracolsep{\fill}}lrrrrr@{}}
\toprule
\textbf{Method} & \shortstack{\bfseries HPWL gain (\%)\\\bfseries mean $\pm$ SD}
& \shortstack{\bfseries Mean\\\bfseries rank} & \shortstack{\bfseries Train\\\bfseries (h)}
& \shortstack{\bfseries Checks\\\bfseries (min)} & \shortstack{\bfseries Repair\\\bfseries (h)}\\
\midrule
\textbf{Frozen} & $0.00\pm0.00$ & 2.917 & 0.00 & 0.00 & 0.00\\
\textbf{Overwrite} & $-7.63\pm13.72$ & 3.917 & 19.56 & 0.00 & 19.56\\
\textbf{RS} & $0.39\pm10.48$ & 2.792 & 31.93 & 24.58 & 32.34\\
\textbf{Cache-only} & $7.04\pm9.03$ & 1.917 & 10.53 & 0.00 & 10.53\\
\textbf{\sys{}} & $8.41\pm10.05$ & 1.750 & 9.66 & 42.26 & 10.36\\
\bottomrule
\end{tabular*}
\endgroup
\vspace{-3pt}
\end{wraptable}
\noindent\textbf{Keeping a repair useful when its circuit returns.}
\sys{} obtains the highest mean HPWL gain and lowest mean rank among the
five methods on the recorded stream, with less repair work than Overwrite
and RS (Table~\ref{tab:chip-main}). Cache-only also avoids repeated
training and is close in aggregate quality and repair cost, so the
benefit of preserving models is shared by both methods. The comparison
therefore concerns how retained models are used as well as whether they
are stored.

Table~\ref{tab:chip-stream-first12-main} connects the aggregate result to
that use. After repairing \texttt{ibm13}, \sys{} processes other circuits
and retrieves the earlier \texttt{ibm13} specialist on both revisits,
without new training. The rejected \texttt{adaptec3} candidate incurs
repair cost but adds no serving model, and the next \texttt{ibm13} request
still uses its saved repair. Local learning can thus accumulate useful
repairs without recovering source training data or overwriting earlier
models. The request records show earlier optimization benefiting later service,
without another training job for each return.
\par
\WFclear
\endgroup

\begin{table}[H]
\centering
\begingroup
\fontsize{7}{8.2}\selectfont
\setlength{\tabcolsep}{1.0pt}
\renewcommand{\arraystretch}{1.15}
\setlength{\aboverulesep}{1.2pt}
\setlength{\belowrulesep}{1.2pt}
\definecolor{ChipTraceBand}{HTML}{F3F5F8}
\caption{Circuit revisits, requests 1 to 12. Repair cost includes training and checks; storage follows each request. Complete records and timing boundaries are in Appendix~\ref{app:chip-stream-accounting}.}
\label{tab:chip-stream-first12-main}
\begin{tabular}{@{}>{\raggedright\arraybackslash}p{.16\linewidth}*{12}{>{\centering\arraybackslash}p{\dimexpr(\linewidth-.16\linewidth-24\tabcolsep)/12\relax}}@{}}
\toprule
\textbf{Request} & \textbf{1} & \textbf{2} & \textbf{3} & \textbf{4} & \textbf{5} & \textbf{6} & \textbf{7} & \textbf{8} & \textbf{9} & \textbf{10} & \textbf{11} & \textbf{12}\\
\midrule
Circuit & ibm07 & ibm18 & ibm15 & ibm13 & bigblue1 & ibm13 & ibm18 & bigblue1 & ibm07 & ibm10 & adaptec3 & ibm13\\
Arrival & 1 & 1 & 1 & 1 & 1 & 2 & 2 & 2 & 2 & 1 & 1 & 3\\
\midrule
\multicolumn{13}{@{}l}{\textbf{HPWL gain over Frozen (\%)}}\\
\rowcolor{ChipTraceBand}
Overwrite & 0.0 & $-0.6$ & $0.8$ & $4.5$ & $4.8$ & $-7.0$ & $-11.5$ & $3.4$ & $-10.3$ & $8.0$ & $-37.0$ & $4.9$\\
RS & 0.0 & 0.0 & 0.0 & $8.2$ & $6.5$ & $0.4$ & $-20.7$ & $9.3$ & $-7.6$ & $8.0$ & $-15.4$ & $3.9$\\
\rowcolor{ChipTraceBand}
Cache-only & 0.0 & 0.0 & 0.0 & 0.0 & 0.0 & $17.0$ & $-2.8$ & $9.2$ & $11.2$ & 0.0 & 0.0 & $9.9$\\
\sys{} & 0.0 & 0.0 & 0.0 & 0.0 & 0.0 & $20.7$ & $17.6$ & $9.5$ & $16.3$ & 0.0 & 0.0 & $16.0$\\
\midrule
\multicolumn{13}{@{}l}{\textbf{Repair cost and \sys{} storage}}\\
\rowcolor{ChipTraceBand}
Overwrite cost (min) & 89.8 & 98.4 & 80.1 & 80.1 & 18.9 & 80.1 & 80.1 & 19.3 & 82.0 & 80.1 & 3.4 & 80.1\\
\sys{} cost (min) & 90.0 & 85.5 & 85.4 & 85.5 & 23.9 & 0.0 & 0.0 & 0.0 & 0.0 & 85.6 & 83.1 & 0.0\\
\rowcolor{ChipTraceBand}
Storage (MiB) & 37.3 & 74.6 & 112.0 & 149.3 & 186.6 & 186.6 & 186.6 & 186.6 & 186.6 & 223.9 & 223.9 & 223.9\\
\bottomrule
\end{tabular}
\endgroup
\end{table}

\paragraph{Reusing a repair for a new circuit.}
\label{sec:eval-e2x}\label{sec:results-e2x}
To make accumulated expertise useful on new designs, we test whether
a saved repair can replace further training without replaying earlier
circuit data. We compare training for each new circuit with checking
saved specialists first, using the same request order and service seeds.
The base serves each circuit's first request before any check or training, and a
passing specialist serves later requests unchanged.

\begin{figure}[htbp]
\centering
\includegraphics[width=0.95\linewidth]{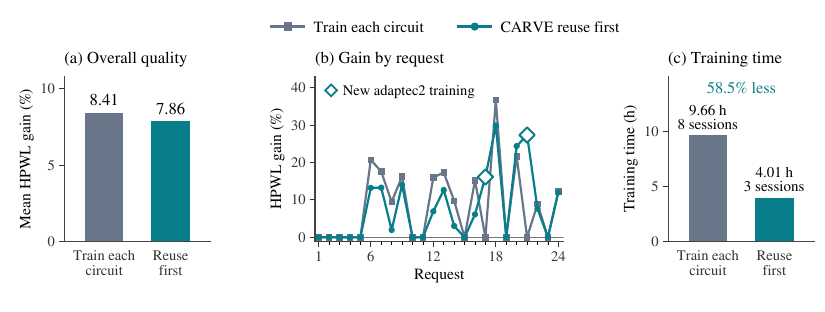}
\caption{Reuse across 24 circuit requests. Gain is the median HPWL reduction versus Frozen on 30 matched trials. }
\label{fig:chip-reuse}
\end{figure}

Figure~\ref{fig:chip-reuse} shows that local validation extends one
\texttt{ibm07} repair to five new circuits, reducing training work.
On the five reused circuits, reuse improves over Frozen but gives lower
mean gains than separate training;
new \texttt{adaptec2} training partly offsets the difference overall.
The result supports reusing accumulated expertise, with a quality trade-off.
Different training seeds and integration code limit causal attribution, as shown in Appendix~\ref{app:chip-reuse-details}.

\begingroup
\setlength{\columnsep}{10pt}
\setlength{\intextsep}{4pt plus 1pt minus 1pt}
\setlength{\emergencystretch}{1em}
\Needspace{12\baselineskip}
\noindent
\begin{wrapfigure}{r}{0.40\textwidth}
\vspace{-5pt}
\centering
\includegraphics[width=\linewidth]{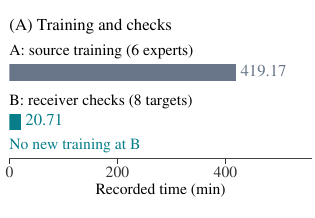}
\par\vspace{-8pt}
\caption{Training and checks.}
\label{fig:chip-receiver}
\vspace{-10pt}
\end{wrapfigure}
\noindent\textbf{Reusing another operator's models without its data.}
\label{sec:eval-transfer}\label{sec:results-transfer}
To test whether an existing library helps an operator without its source
data, a simulated operator B imports six specialists from operator A.
It checks them on seven new IBM circuits and an \texttt{adaptec1} control
using 30 paired trials. The earliest source-admitted passing model serves
on 30 separate seeds, with Frozen as fallback. Figure~\ref{fig:chip-receiver} shows checking the full library costs 20.71 minutes across eight targets,
with no receiver training. Overall,
\sys{} achieves a 5.76\% mean HPWL gain over Frozen across all eight
targets, averaging each target's median gain and including both base
fallbacks. Six new circuits use imports, showing useful reuse
without source training data. The 419.17-minute source training cost was
paid once. Per-target results, uncertainty, and timing definitions remain in
Appendix~\ref{app:chip-receiver-details}.
\par
\WFclear
\endgroup

\begingroup
\setlength{\columnsep}{10pt}
\setlength{\intextsep}{4pt plus 1pt minus 1pt}
\setlength{\emergencystretch}{1em}
\Needspace{11\baselineskip}
\noindent
\begin{wraptable}{r}{0.40\textwidth}
\vspace{-5pt}
\centering
\begingroup
\fontsize{8}{9.2}\selectfont
\setlength{\tabcolsep}{3pt}
\renewcommand{\arraystretch}{1.06}
\setlength{\aboverulesep}{1.4pt}
\setlength{\belowrulesep}{1.4pt}
\caption{Median HPWL gain over Frozen (\%) on 30 matched service trials. Higher is better.}
\label{tab:chip-ablation-compact}\label{tab:chip-scope-check}
\begin{tabular*}{\linewidth}{@{\extracolsep{\fill}}rlrr@{}}
\toprule
Request & Circuit & RS & \sys{}\\
\midrule
6 & \texttt{ibm13} & 0.39 & \textbf{20.71}\\
7 & \texttt{ibm18} & -20.73 & \textbf{17.57}\\
8 & \texttt{bigblue1} & 9.33 & \textbf{9.48}\\
9 & \texttt{ibm07} & -7.61 & \textbf{16.27}\\
\bottomrule
\end{tabular*}
\endgroup
\vspace{-3pt}
\end{wraptable}
\noindent\textbf{Selecting a model checked for the arriving circuit.}
\label{sec:chip-components-review}\label{sec:chip-rq3}
Saved repairs need evidence of suitability for the arriving circuit.
In four consecutive requests, RS reuses its \texttt{bigblue1} specialist,
while \sys{} retrieves each circuit's validated repair, as shown in Table~\ref{tab:chip-scope-check}. Gains are similar on \texttt{bigblue1}.
On \texttt{ibm18} and \texttt{ibm07}, RS worsens HPWL while \sys{} improves
it. Different models and training histories prevent isolating the check's
causal effect. Full HPWL values, model identities, and the complete RS
trace, including beneficial uses across circuits, remain in
Appendix~\ref{app:chip-gates}. The chip studies show how retained models support reuse when circuits
return, on new circuits, and across operators. Local evaluation extends
their use without source design data and can reduce new training.

\par
\WFclear
\endgroup

\FloatBarrier
\endgroup
\FloatBarrier

\section{Related Works}
\label{sec:related}

Continual World evaluates retention and transfer during sequential
reinforcement learning~\citep{wolczyk2021continualworld}. Progressive
Neural Networks preserve earlier columns~\citep{rusu2016progressive},
PackNet allocates parameters through pruning~\citep{mallya2018packnet},
and CompoNet retains and combines policy modules
\citep{malagon2024componet}. CARVE uses model isolation and studies the
deployment decisions around saved models. The experiments compare
whether a candidate is checked, whether its task scope is recorded,
and whether the appropriate specialist is used when a task returns.
Cache-only and RS test those choices with the same domain trainer. Learning without Forgetting uses new-task data to retain earlier behavior
\citep{li2016lwf}, while Tent adapts selected parameters using unlabeled
test inputs~\citep{wang2021tent}. Source-data restrictions therefore do
not rule out all existing adaptation methods. CARVE studies episodic
tasks with local evaluation and separates generating a candidate from
allowing it to serve a particular family. The proposed plan and
base-penalty studies examine that distinction for navigation. High Confidence Policy Improvement provides probabilistic guarantees
under its stated assumptions~\citep{thomas2015hcpi}. ChiPFormer supplies the pretrained placement model and adaptation
procedure~\citep{chipformer}. Our placement studies compare deployment
rules around that procedure, including repeated use, extension to new
circuits, and local evaluation of imported models.

\section{Conclusion}
\label{sec:conclusion}

CARVE keeps a repair as a separate model and records the tasks on which
it has passed evaluation. In the evaluated navigation and placement
streams, retaining specialists reduces repeated training and preserves
the earlier model identities. Service quality is less uniform, with
useful rejections and unfavorable navigation results alongside gains.
Local extension tests allow some new tasks to reuse a saved model without
training, although reuse can improve quality less than training for the
target. The component studies show why retention, admission, and task
scope should be evaluated separately. The resulting evidence concerns
an automated repair procedure with measured costs and outcomes, rather
than a guarantee that every repair or transfer improves future service.

\FloatBarrier

\bibliographystyle{iclr2027_conference}
\begingroup
\raggedright
\setlength{\emergencystretch}{2em}
\bibliography{refs}
\endgroup

\providecommand{\FTCRTableStyle}{}
\renewcommand{\FTCRTableStyle}{\fontsize{7.5}{8.6}\selectfont
  \setlength{\tabcolsep}{3pt}\renewcommand{\arraystretch}{1.04}
  \setlength{\aboverulesep}{1.4pt}\setlength{\belowrulesep}{1.4pt}}
\providecommand{\FTCRDenseTableStyle}{}
\renewcommand{\FTCRDenseTableStyle}{\fontsize{7}{8.1}\selectfont
  \setlength{\tabcolsep}{2pt}\renewcommand{\arraystretch}{1.02}
  \setlength{\aboverulesep}{1.2pt}\setlength{\belowrulesep}{1.2pt}}

\clearpage
\appendix
\begingroup
\setcounter{secnumdepth}{2}
\setcounter{tocdepth}{2}
\renewcommand{\thetable}{\thesection\arabic{table}}
\renewcommand{\thefigure}{\thesection\arabic{figure}}
\providecommand{\theHtable}{\arabic{table}}
\providecommand{\theHfigure}{\arabic{figure}}
\renewcommand{\theHtable}{appendix.\thesection.\arabic{table}}
\renewcommand{\theHfigure}{appendix.\thesection.\arabic{figure}}
\setlength{\emergencystretch}{2em}
\setlength{\intextsep}{5pt plus 1pt minus 1pt}
\setlength{\textfloatsep}{7pt plus 2pt minus 1pt}
\setlength{\abovecaptionskip}{3pt}
\setlength{\belowcaptionskip}{2pt}
\providecommand{\FTCRAppTableStyle}{}
\renewcommand{\FTCRAppTableStyle}{\FTCRTableStyle}
\providecommand{\FTCRAppDenseStyle}{}
\renewcommand{\FTCRAppDenseStyle}{\FTCRDenseTableStyle}
\providecommand{\NavStudy}{Navigation streams}
\providecommand{\ChipStudy}{E2}
\providecommand{\ReuseStudy}{E2-X}
\providecommand{\TransferStudy}{E-T}
\providecommand{\ForgetStudy}{A--B--A forgetting test}
\providecommand{\RecheckStudy}{Sampled-arrival recheck}
\providecommand{\ReuseArm}{CARVE reuse-first}
\renewcommand{\NavStudy}{Navigation streams}
\renewcommand{\ChipStudy}{E2}
\renewcommand{\ReuseStudy}{E2-X}
\renewcommand{\TransferStudy}{E-T}
\renewcommand{\ForgetStudy}{A--B--A forgetting test}
\renewcommand{\RecheckStudy}{Sampled-arrival recheck}
\renewcommand{\ReuseArm}{CARVE reuse-first}

\FloatBarrier
\section{Main experiment records}
\setcounter{table}{0}\setcounter{figure}{0}
\label{app:evidence-details}

This appendix reports benchmark workloads, service records, and repair costs.
Reuse and transfer appear in Appendix~\ref{app:reuse-records}, and component
studies appear in Appendix~\ref{app:ablation-records}. Measurement definitions
are in Appendix~\ref{app:measurement-details}. Methods, domain rules, and terms
are collected in Appendices~\ref{app:deployment-methods},
\ref{app:domain-contracts}, and~\ref{app:terminology}.
B1 denotes Frozen, B3 denotes Cache-only, and OV denotes Overwrite.
Archived chip slots start at zero, while displayed request numbers start at one.

\FloatBarrier
\subsection{Navigation workloads and repair work}
\label{app:nav-work-details}\label{app:nav-stream-records}

The \emph{repair retention study} compares Frozen, Overwrite, and CARVE
on circle crossing with ORCA+ (CO) or SFM (CS) pedestrians and an ORCA+
bottleneck hallway (HO). Fixed and sampled arrivals use the same pool of
scene settings. The \emph{caching and admission study} adds Cache-only,
square crossing (SQ), and a constricted hallway (HS). It samples request
order and scene settings while fixing the number of requests per family.
We show one such stream in detail and report the additional completed
stream below.

Each stream starts from the same frozen base with an empty repair state.
The method set and workload differ between the two studies, so their
results do not isolate admission alone. Table~\ref{tab:nav-scope} gives
the family counts in the three detailed streams.

\begin{table}[H]
\centering
\FTCRTableStyle
\caption{Workload composition of the three detailed streams. Each request evaluates 100 matched cases.}
\label{tab:appendix-A1}
\label{tab:nav-scope}
\begin{tabularx}{\linewidth}{@{}>{\raggedright\arraybackslash}X l l rrr@{}}
\toprule
& & & \multicolumn{2}{c}{Repair retention} & Caching and admission\\
\cmidrule(lr){4-5}\cmidrule(l){6-6}
Geometry & Pedestrians & Family & \shortstack{Fixed\\arrivals} & \shortstack{Sampled\\arrivals} & \shortstack{Fixed family\\counts}\\
\midrule
Circle crossing & ORCA+ & F-CO & 14 & 16 & 8\\
Circle crossing & SFM & F-CS & 7 & 7 & 5\\
Bottleneck hallway & ORCA+ & F-HO & 9 & 7 & 6\\
Square crossing & ORCA+ & F-SQ & --- & --- & 6\\
Constricted hallway & ORCA+ & F-HS & --- & --- & 5\\
\midrule
\multicolumn{3}{@{}l}{Request positions} & 30 & 30 & 30\\
\multicolumn{3}{@{}l}{Original-batch methods} & 3 & 3 & 4\\
\multicolumn{3}{@{}l}{Original-batch service episodes} & 9,000 & 9,000 & 12,000\\
\bottomrule
\end{tabularx}
\end{table}

\paragraph{Scope of the caching and admission comparison.}
\label{app:additional-arrival}
The two completed streams use the same family counts but different
sampled orders and scene settings. The detailed display was chosen during
editing without a predeclared exclusion rule. In the other completed
stream, CARVE achieves 57.2\% success, compared with 60.7\% for Cache-only
and 62.9\% for Overwrite. CARVE trains fewer times than both methods in
each stream, but its success advantage does not hold across both streams.
Its total repair time is lower than Overwrite's and higher than
Cache-only's in both streams. The full records remain in the archive.

\paragraph{Fixed and sampled arrivals for repair retention.}
Fixed arrivals follow the predefined list of 30 requests. Sampled arrivals
draw 30 requests from 43 allowed scene settings, comprising 25 CO, nine CS,
and nine HO settings. Settings can repeat, but the sampler redraws a
setting if it has already appeared four times. Sampling changes both
the family counts and the arrival order (Table~\ref{tab:nav-scope}).
Each request evaluates 100 cases, and repeated settings use the next
consecutive block of 100 cases.

\paragraph{Sampled arrivals for caching and admission.}
Both streams contain eight CO, five CS, six HO, six SQ, and five
HS requests. The generator first shuffles this family list, then draws
scene settings for each request. SQ uses widths 3, 4, and 5 with three to
five pedestrians. HS uses the recorded radius field 1.25, 1.5, or 1.75
with three to five pedestrians. A setting is redrawn if it has no eligible
case block.

For each setting, the generator excludes blocks already used in the
repair retention study or the SQ/HS census. It takes the next eligible
100-case block in ascending order without reusing a block within that
stream. Admission and service use different case blocks;
Appendix~\ref{sec:audit-blocks} describes their relationship to training.
A case is defined by its family, scene settings, and index, so the same
index with different settings does not identify the same case.

\paragraph{Reproducing the arrival histories.}
The archived request lists retain each stream's order, scene settings,
and generator configuration. Arrival randomization is separate from
model-training randomness.

Circle and square outlines mark sampling regions, not walls; square pedestrians start and finish on opposite sides. Both hallway openings are 1 m within a 2 m corridor, but the bottleneck uses a transverse wall and HS uses sloping walls. The recorded \path{circle_radius} is not the opening width. The square shim repairs missing environment references without adding walls.

\begin{table}[H]
\centering\FTCRAppTableStyle
\caption{Fixed navigation execution settings. Case allocation is described in Appendix~\ref{sec:audit-blocks}.}
\label{tab:appendix-A2}\label{tab:nav-settings-details}
\begin{tabularx}{\linewidth}{@{}l>{\raggedright\arraybackslash}X@{}}
\toprule
Setting & Value\\
\midrule
Base and simulator & Frozen SARL, archived CrowdSimPlus\\
Service and episode limit & 100 episodes per request, 90 seconds per episode\\
Alarm & Two-sided 95\% Clopper--Pearson lower endpoint below .80; equivalently at most 88 successes\\
Local training & 60,000 full fine-tuning steps, learning rate $2.5\times10^{-5}$, batch size 64\\
Training data & Current family only; no replay from earlier families\\
Admission and extension & Same 100 cases for candidate and base; at least 15 additional swept successes\\
Observed service blocks & 100 matched cases per request; separate from admission\\
Cache limit & Four entries in repair retention; six entries for both cached methods in caching and admission\\
\bottomrule
\end{tabularx}
\end{table}

Across fixed arrivals, sampled arrivals, and caching and admission,
Overwrite performs 3, 3, and 5 first-family updates and 18, 19, and 20
later updates, for totals of 21, 22, and 25. CARVE runs 2, 2, and 3
training sessions, passes 1, 1, and 2 extension exams, and records 27, 21,
and 20 cache hits, respectively. Cache-only trains five models in the
caching and admission stream. Later Overwrite updates are not individually
classified as parameter forgetting.

The three detailed streams have 11 first and 57 subsequent Overwrite updates. Exact CARVE phase totals are in Table~\ref{tab:appendix-D1}.

Retaining models reduces repeated work. Cache-only has lower total repair cost in the displayed caching and admission stream. With fixed arrivals, hallway repairs dominate Overwrite's cost, while CARVE's hallway success stays close to the base. SFM reuse avoids training but still incurs an extension check.

\begin{table}[H]
\centering\FTCRAppDenseStyle
\caption{Family breakdown for fixed arrivals over all 30 requests. Repair includes training and all checks. Frozen has no repair work.}
\label{tab:appendix-A5}\label{tab:nav-s1-family-work}
\begin{tabular*}{\linewidth}{@{\extracolsep{\fill}}lrrrrrrrr@{}}
\toprule
& & \multicolumn{3}{c}{Success (\%)} & \multicolumn{2}{c}{Trainings} & \multicolumn{2}{c}{Repair (min)}\\
\cmidrule(lr){3-5}\cmidrule(lr){6-7}\cmidrule(l){8-9}
Family & Requests & Frozen & Overwrite & CARVE & Overwrite & CARVE & Overwrite & CARVE\\
\midrule
Circle, ORCA+ & 14 & 27.6 & 65.6 & 77.1 & 11 & 1 & 212.9 & 26.2\\
Circle, SFM & 7 & 57.9 & 92.6 & 93.4 & 1 & 0 & 18.4 & 1.5\\
Bottleneck hallway & 9 & 30.4 & 23.0 & 31.7 & 9 & 1 & 553.9 & 76.4\\
\bottomrule
\end{tabular*}
\end{table}

\paragraph{Complete request records.}
Tables~\ref{tab:appendix-A6}, \ref{tab:appendix-A7}, and~\ref{tab:appendix-A9} give one column per request in arrival order, in two panels of 15 requests. Success is out of 100 for every method. Repair minutes are charged after that request, so they affect only later service. A model is named for its training family, not its current authorized target. The records preserve first arrivals, rejections, and unfavorable service. The size row gives radius (width for SQ), and the pedestrian row gives the pedestrian count. After-service actions are T+ (train/admit), T- (train/reject), E+ (extend), or keep (no new permission). The original archive, which also retains the additional arrival, preserves exact blocks, model identities, phase times, and decisions in \path{all_420_method_requests.csv}; \path{all_120_requests.csv} has one workload row per request.

\begin{table}[H]
\centering
\begingroup
\caption{Fixed arrivals: all 30 requests, with each panel following arrival order. Success is out of 100 cases; repair time is charged after service.}
\label{tab:appendix-A6}
\fontsize{7}{8.2}\selectfont
\setlength{\tabcolsep}{1.4pt}
\renewcommand{\arraystretch}{1.15}
\setlength{\aboverulesep}{1.2pt}
\setlength{\belowrulesep}{1.2pt}
\definecolor{NavTraceCO}{HTML}{E7F0F8}
\definecolor{NavTraceCS}{HTML}{EAF3EC}
\definecolor{NavTraceHO}{HTML}{F4EDE3}
\definecolor{NavTraceHS}{HTML}{EFE9F5}
\definecolor{NavTraceSQ}{HTML}{E4F2F1}
\definecolor{NavTraceBand}{HTML}{F3F5F8}

\begin{tabular}{@{}>{\raggedright\arraybackslash}p{.18\linewidth}*{15}{>{\centering\arraybackslash}p{\dimexpr(\linewidth-.18\linewidth-30\tabcolsep)/15\relax}}@{}}
\toprule
\textbf{Request} & \textbf{1} & \textbf{2} & \textbf{3} & \textbf{4} & \textbf{5} & \textbf{6} & \textbf{7} & \textbf{8} & \textbf{9} & \textbf{10} & \textbf{11} & \textbf{12} & \textbf{13} & \textbf{14} & \textbf{15}\\
\midrule
Family & \cellcolor{NavTraceCO}\textbf{CO} & \cellcolor{NavTraceCO}\textbf{CO} & \cellcolor{NavTraceHO}\textbf{HO} & \cellcolor{NavTraceCO}\textbf{CO} & \cellcolor{NavTraceCS}\textbf{CS} & \cellcolor{NavTraceCS}\textbf{CS} & \cellcolor{NavTraceHO}\textbf{HO} & \cellcolor{NavTraceCO}\textbf{CO} & \cellcolor{NavTraceCO}\textbf{CO} & \cellcolor{NavTraceCS}\textbf{CS} & \cellcolor{NavTraceHO}\textbf{HO} & \cellcolor{NavTraceCO}\textbf{CO} & \cellcolor{NavTraceCO}\textbf{CO} & \cellcolor{NavTraceHO}\textbf{HO} & \cellcolor{NavTraceCS}\textbf{CS}\\
Size $r[w]$ & 2 & 2.25 & 1.75 & 2 & 1.375 & 1.375 & 1.75 & 1.5 & 1.25 & 2.25 & 1.75 & 2.5 & 1.375 & 1.75 & 1.375\\
Pedestrians & 5 & 4 & 5 & 5 & 4 & 5 & 5 & 5 & 2 & 4 & 4 & 3 & 5 & 5 & 4\\
\midrule
\rowcolor{NavTraceBand}
Frozen success & 20 & 11 & 27 & 22 & 80 & 74 & 26 & 35 & 51 & 36 & 39 & 7 & 46 & 28 & 78\\
Overwrite success & 20 & 97 & 18 & 20 & 96 & 91 & 28 & 69 & 89 & 92 & 26 & 88 & 45 & 13 & 94\\
\rowcolor{NavTraceBand}
\sys{} success & 20 & 97 & 27 & 81 & 80 & 96 & 28 & 64 & 94 & 95 & 40 & 92 & 65 & 29 & 98\\
\midrule
\rowcolor{NavTraceBand}
\sys{} route & base & CO & base & CO & base & CO & HO & CO & CO & CO & HO & CO & CO & HO & CO\\
After service & T+ & keep & T+ & keep & E+ & keep & keep & keep & keep & keep & keep & keep & keep & keep & keep\\
\midrule
\rowcolor{NavTraceBand}
Overwrite repair (min) & 20.2 & 0.0 & 63.0 & 20.8 & 0.0 & 0.0 & 64.5 & 21.0 & 0.0 & 0.0 & 55.3 & 14.8 & 21.2 & 65.4 & 0.0\\
\sys{} repair (min) & 26.2 & 0.0 & 76.4 & 0.0 & 1.5 & 0.0 & 0.0 & 0.0 & 0.0 & 0.0 & 0.0 & 0.0 & 0.0 & 0.0 & 0.0\\
\bottomrule
\end{tabular}
\par\vspace{5pt}
\begin{tabular}{@{}>{\raggedright\arraybackslash}p{.18\linewidth}*{15}{>{\centering\arraybackslash}p{\dimexpr(\linewidth-.18\linewidth-30\tabcolsep)/15\relax}}@{}}
\toprule
\textbf{Request} & \textbf{16} & \textbf{17} & \textbf{18} & \textbf{19} & \textbf{20} & \textbf{21} & \textbf{22} & \textbf{23} & \textbf{24} & \textbf{25} & \textbf{26} & \textbf{27} & \textbf{28} & \textbf{29} & \textbf{30}\\
\midrule
Family & \cellcolor{NavTraceCO}\textbf{CO} & \cellcolor{NavTraceHO}\textbf{HO} & \cellcolor{NavTraceCO}\textbf{CO} & \cellcolor{NavTraceCS}\textbf{CS} & \cellcolor{NavTraceHO}\textbf{HO} & \cellcolor{NavTraceCO}\textbf{CO} & \cellcolor{NavTraceCO}\textbf{CO} & \cellcolor{NavTraceHO}\textbf{HO} & \cellcolor{NavTraceCS}\textbf{CS} & \cellcolor{NavTraceCO}\textbf{CO} & \cellcolor{NavTraceHO}\textbf{HO} & \cellcolor{NavTraceCO}\textbf{CO} & \cellcolor{NavTraceCS}\textbf{CS} & \cellcolor{NavTraceHO}\textbf{HO} & \cellcolor{NavTraceCO}\textbf{CO}\\
Size $r[w]$ & 2 & 1.25 & 1.25 & 2.25 & 1.75 & 1.75 & 1.5 & 1.75 & 1.375 & 2.25 & 1.75 & 1.25 & 2.25 & 1.25 & 2.5\\
Pedestrians & 5 & 5 & 5 & 5 & 4 & 2 & 5 & 5 & 5 & 2 & 4 & 4 & 2 & 5 & 5\\
\midrule
\rowcolor{NavTraceBand}
Frozen success & 22 & 24 & 36 & 33 & 38 & 37 & 29 & 24 & 76 & 15 & 40 & 51 & 28 & 28 & 4\\
Overwrite success & 74 & 27 & 52 & 87 & 31 & 93 & 60 & 18 & 94 & 80 & 28 & 60 & 94 & 18 & 71\\
\rowcolor{NavTraceBand}
\sys{} success & 81 & 32 & 65 & 93 & 47 & 96 & 73 & 26 & 97 & 93 & 34 & 67 & 95 & 22 & 92\\
\midrule
\rowcolor{NavTraceBand}
\sys{} route & CO & HO & CO & CO & HO & CO & CO & HO & CO & CO & HO & CO & CO & HO & CO\\
After service & keep & keep & keep & keep & keep & keep & keep & keep & keep & keep & keep & keep & keep & keep & keep\\
\midrule
\rowcolor{NavTraceBand}
Overwrite repair (min) & 20.6 & 64.0 & 21.1 & 18.4 & 55.4 & 0.0 & 21.2 & 64.6 & 0.0 & 12.8 & 54.5 & 19.1 & 0.0 & 67.2 & 20.1\\
\sys{} repair (min) & 0.0 & 0.0 & 0.0 & 0.0 & 0.0 & 0.0 & 0.0 & 0.0 & 0.0 & 0.0 & 0.0 & 0.0 & 0.0 & 0.0 & 0.0\\
\bottomrule
\end{tabular}
\endgroup
\end{table}

\begin{table}[H]
\centering
\begingroup
\caption{Sampled arrivals: all 30 requests, with each panel following arrival order. Success is out of 100 cases; repair time is charged after service.}
\label{tab:appendix-A7}
\fontsize{7}{8.2}\selectfont
\setlength{\tabcolsep}{1.4pt}
\renewcommand{\arraystretch}{1.15}
\setlength{\aboverulesep}{1.2pt}
\setlength{\belowrulesep}{1.2pt}
\definecolor{NavTraceCO}{HTML}{E7F0F8}
\definecolor{NavTraceCS}{HTML}{EAF3EC}
\definecolor{NavTraceHO}{HTML}{F4EDE3}
\definecolor{NavTraceHS}{HTML}{EFE9F5}
\definecolor{NavTraceSQ}{HTML}{E4F2F1}
\definecolor{NavTraceBand}{HTML}{F3F5F8}

\begin{tabular}{@{}>{\raggedright\arraybackslash}p{.18\linewidth}*{15}{>{\centering\arraybackslash}p{\dimexpr(\linewidth-.18\linewidth-30\tabcolsep)/15\relax}}@{}}
\toprule
\textbf{Request} & \textbf{1} & \textbf{2} & \textbf{3} & \textbf{4} & \textbf{5} & \textbf{6} & \textbf{7} & \textbf{8} & \textbf{9} & \textbf{10} & \textbf{11} & \textbf{12} & \textbf{13} & \textbf{14} & \textbf{15}\\
\midrule
Family & \cellcolor{NavTraceCO}\textbf{CO} & \cellcolor{NavTraceCO}\textbf{CO} & \cellcolor{NavTraceCS}\textbf{CS} & \cellcolor{NavTraceCO}\textbf{CO} & \cellcolor{NavTraceCS}\textbf{CS} & \cellcolor{NavTraceHO}\textbf{HO} & \cellcolor{NavTraceHO}\textbf{HO} & \cellcolor{NavTraceHO}\textbf{HO} & \cellcolor{NavTraceCO}\textbf{CO} & \cellcolor{NavTraceHO}\textbf{HO} & \cellcolor{NavTraceCS}\textbf{CS} & \cellcolor{NavTraceCS}\textbf{CS} & \cellcolor{NavTraceCO}\textbf{CO} & \cellcolor{NavTraceCS}\textbf{CS} & \cellcolor{NavTraceCO}\textbf{CO}\\
Size $r[w]$ & 2.5 & 1.5 & 2.25 & 1.5 & 1.375 & 1.25 & 1.75 & 1.5 & 2 & 1.25 & 2.25 & 1.375 & 1.375 & 2.25 & 2\\
Pedestrians & 2 & 3 & 4 & 2 & 5 & 5 & 5 & 5 & 3 & 5 & 2 & 2 & 4 & 5 & 4\\
\midrule
\rowcolor{NavTraceBand}
Frozen success & 13 & 47 & 36 & 41 & 74 & 24 & 27 & 25 & 13 & 28 & 28 & 67 & 45 & 33 & 20\\
Overwrite success & 13 & 85 & 82 & 94 & 90 & 15 & 41 & 36 & 81 & 22 & 69 & 100 & 78 & 95 & 91\\
\rowcolor{NavTraceBand}
\sys{} success & 13 & 85 & 36 & 98 & 92 & 24 & 27 & 25 & 92 & 28 & 94 & 97 & 70 & 91 & 84\\
\midrule
\rowcolor{NavTraceBand}
\sys{} route & base & CO & base & CO & CO & base & base & base & CO & base & CO & CO & CO & CO & CO\\
After service & T+ & keep & E+ & keep & keep & T- & keep & keep & keep & keep & keep & keep & keep & keep & keep\\
\midrule
\rowcolor{NavTraceBand}
Overwrite repair (min) & 13.5 & 15.4 & 16.2 & 0.0 & 0.0 & 65.4 & 63.5 & 67.5 & 14.7 & 66.7 & 12.6 & 0.0 & 18.6 & 0.0 & 0.0\\
\sys{} repair (min) & 18.5 & 0.0 & 5.0 & 0.0 & 0.0 & 72.2 & 0.0 & 0.0 & 0.0 & 0.0 & 0.0 & 0.0 & 0.0 & 0.0 & 0.0\\
\bottomrule
\end{tabular}
\par\vspace{5pt}
\begin{tabular}{@{}>{\raggedright\arraybackslash}p{.18\linewidth}*{15}{>{\centering\arraybackslash}p{\dimexpr(\linewidth-.18\linewidth-30\tabcolsep)/15\relax}}@{}}
\toprule
\textbf{Request} & \textbf{16} & \textbf{17} & \textbf{18} & \textbf{19} & \textbf{20} & \textbf{21} & \textbf{22} & \textbf{23} & \textbf{24} & \textbf{25} & \textbf{26} & \textbf{27} & \textbf{28} & \textbf{29} & \textbf{30}\\
\midrule
Family & \cellcolor{NavTraceCO}\textbf{CO} & \cellcolor{NavTraceCO}\textbf{CO} & \cellcolor{NavTraceCO}\textbf{CO} & \cellcolor{NavTraceCO}\textbf{CO} & \cellcolor{NavTraceCO}\textbf{CO} & \cellcolor{NavTraceHO}\textbf{HO} & \cellcolor{NavTraceHO}\textbf{HO} & \cellcolor{NavTraceCO}\textbf{CO} & \cellcolor{NavTraceCS}\textbf{CS} & \cellcolor{NavTraceHO}\textbf{HO} & \cellcolor{NavTraceCO}\textbf{CO} & \cellcolor{NavTraceCO}\textbf{CO} & \cellcolor{NavTraceCO}\textbf{CO} & \cellcolor{NavTraceCS}\textbf{CS} & \cellcolor{NavTraceCO}\textbf{CO}\\
Size $r[w]$ & 1.25 & 2.25 & 1.25 & 2.5 & 2 & 1.5 & 1.5 & 2 & 1.75 & 1.5 & 1.25 & 1.5 & 1.75 & 2.25 & 2\\
Pedestrians & 2 & 2 & 2 & 2 & 3 & 5 & 4 & 2 & 4 & 3 & 2 & 2 & 4 & 4 & 4\\
\midrule
\rowcolor{NavTraceBand}
Frozen success & 51 & 15 & 57 & 11 & 18 & 19 & 30 & 19 & 43 & 44 & 63 & 55 & 30 & 35 & 17\\
Overwrite success & 87 & 88 & 86 & 83 & 89 & 16 & 42 & 57 & 87 & 44 & 89 & 87 & 87 & 96 & 86\\
\rowcolor{NavTraceBand}
\sys{} success & 87 & 97 & 86 & 97 & 95 & 19 & 30 & 99 & 92 & 44 & 94 & 93 & 83 & 97 & 89\\
\midrule
\rowcolor{NavTraceBand}
\sys{} route & CO & CO & CO & CO & CO & base & base & CO & CO & base & CO & CO & CO & CO & CO\\
After service & keep & keep & keep & keep & keep & keep & keep & keep & keep & keep & keep & keep & keep & keep & keep\\
\midrule
\rowcolor{NavTraceBand}
Overwrite repair (min) & 13.9 & 12.8 & 13.7 & 12.9 & 0.0 & 66.4 & 55.1 & 13.1 & 16.4 & 43.9 & 0.0 & 13.1 & 17.7 & 0.0 & 17.4\\
\sys{} repair (min) & 0.0 & 0.0 & 0.0 & 0.0 & 0.0 & 0.0 & 0.0 & 0.0 & 0.0 & 0.0 & 0.0 & 0.0 & 0.0 & 0.0 & 0.0\\
\bottomrule
\end{tabular}
\endgroup
\end{table}

\begin{table}[H]
\centering
\begingroup
\caption{Caching and admission: all 30 requests, with each panel following arrival order. Success is out of 100 cases; repair time is charged after service.}
\label{tab:appendix-A9}
\fontsize{7}{8.2}\selectfont
\setlength{\tabcolsep}{1.4pt}
\renewcommand{\arraystretch}{1.15}
\setlength{\aboverulesep}{1.2pt}
\setlength{\belowrulesep}{1.2pt}
\definecolor{NavTraceCO}{HTML}{E7F0F8}
\definecolor{NavTraceCS}{HTML}{EAF3EC}
\definecolor{NavTraceHO}{HTML}{F4EDE3}
\definecolor{NavTraceHS}{HTML}{EFE9F5}
\definecolor{NavTraceSQ}{HTML}{E4F2F1}
\definecolor{NavTraceBand}{HTML}{F3F5F8}

\begin{tabular}{@{}>{\raggedright\arraybackslash}p{.18\linewidth}*{15}{>{\centering\arraybackslash}p{\dimexpr(\linewidth-.18\linewidth-30\tabcolsep)/15\relax}}@{}}
\toprule
\textbf{Request} & \textbf{1} & \textbf{2} & \textbf{3} & \textbf{4} & \textbf{5} & \textbf{6} & \textbf{7} & \textbf{8} & \textbf{9} & \textbf{10} & \textbf{11} & \textbf{12} & \textbf{13} & \textbf{14} & \textbf{15}\\
\midrule
Family & \cellcolor{NavTraceCS}\textbf{CS} & \cellcolor{NavTraceCS}\textbf{CS} & \cellcolor{NavTraceCO}\textbf{CO} & \cellcolor{NavTraceCO}\textbf{CO} & \cellcolor{NavTraceCS}\textbf{CS} & \cellcolor{NavTraceCO}\textbf{CO} & \cellcolor{NavTraceHO}\textbf{HO} & \cellcolor{NavTraceHS}\textbf{HS} & \cellcolor{NavTraceSQ}\textbf{SQ} & \cellcolor{NavTraceHO}\textbf{HO} & \cellcolor{NavTraceHO}\textbf{HO} & \cellcolor{NavTraceCO}\textbf{CO} & \cellcolor{NavTraceCO}\textbf{CO} & \cellcolor{NavTraceCO}\textbf{CO} & \cellcolor{NavTraceHS}\textbf{HS}\\
Size $r[w]$ & 2.25 & 1.375 & 2.5 & 2 & 1.75 & 1.5 & 1.25 & 1.75 & 4 & 1.75 & 1.25 & 1.75 & 1.25 & 1.25 & 1.75\\
Pedestrians & 5 & 2 & 3 & 4 & 5 & 3 & 5 & 4 & 5 & 3 & 4 & 2 & 3 & 3 & 4\\
\midrule
\rowcolor{NavTraceBand}
Frozen success & 33 & 64 & 11 & 22 & 60 & 48 & 23 & 31 & 33 & 51 & 42 & 33 & 55 & 59 & 34\\
Overwrite success & 33 & 100 & 95 & 91 & 96 & 87 & 21 & 22 & 36 & 28 & 37 & 72 & 82 & 88 & 21\\
\rowcolor{NavTraceBand}
Cache-only success & 33 & 100 & 11 & 87 & 96 & 86 & 23 & 31 & 33 & 2 & 57 & 87 & 86 & 72 & 52\\
\sys{} success & 33 & 100 & 11 & 91 & 96 & 87 & 23 & 31 & 33 & 51 & 42 & 98 & 86 & 86 & 52\\
\midrule
\rowcolor{NavTraceBand}
\sys{} route & base & CS & base & CS & CS & CS & base & base & base & base & base & CS & CS & CS & HS\\
After service & T+ & keep & E+ & keep & keep & keep & T- & T+ & E+ & keep & keep & keep & keep & keep & keep\\
\midrule
\rowcolor{NavTraceBand}
Overwrite repair (min) & 18.8 & 0.0 & 0.0 & 0.0 & 0.0 & 15.7 & 68.3 & 51.8 & 20.9 & 45.0 & 57.7 & 13.2 & 16.4 & 16.6 & 52.1\\
Cache-only repair (min) & 18.7 & 0.0 & 16.0 & 0.0 & 0.0 & 0.0 & 65.1 & 50.7 & 20.8 & 0.0 & 0.0 & 0.0 & 0.0 & 0.0 & 0.0\\
\rowcolor{NavTraceBand}
\sys{} repair (min) & 23.9 & 0.0 & 5.9 & 0.0 & 0.0 & 0.0 & 77.5 & 66.8 & 3.7 & 0.0 & 0.0 & 0.0 & 0.0 & 0.0 & 0.0\\
\bottomrule
\end{tabular}
\par\vspace{5pt}
\begin{tabular}{@{}>{\raggedright\arraybackslash}p{.18\linewidth}*{15}{>{\centering\arraybackslash}p{\dimexpr(\linewidth-.18\linewidth-30\tabcolsep)/15\relax}}@{}}
\toprule
\textbf{Request} & \textbf{16} & \textbf{17} & \textbf{18} & \textbf{19} & \textbf{20} & \textbf{21} & \textbf{22} & \textbf{23} & \textbf{24} & \textbf{25} & \textbf{26} & \textbf{27} & \textbf{28} & \textbf{29} & \textbf{30}\\
\midrule
Family & \cellcolor{NavTraceCO}\textbf{CO} & \cellcolor{NavTraceSQ}\textbf{SQ} & \cellcolor{NavTraceHO}\textbf{HO} & \cellcolor{NavTraceHS}\textbf{HS} & \cellcolor{NavTraceHO}\textbf{HO} & \cellcolor{NavTraceSQ}\textbf{SQ} & \cellcolor{NavTraceHS}\textbf{HS} & \cellcolor{NavTraceCS}\textbf{CS} & \cellcolor{NavTraceCO}\textbf{CO} & \cellcolor{NavTraceHS}\textbf{HS} & \cellcolor{NavTraceHO}\textbf{HO} & \cellcolor{NavTraceSQ}\textbf{SQ} & \cellcolor{NavTraceSQ}\textbf{SQ} & \cellcolor{NavTraceCS}\textbf{CS} & \cellcolor{NavTraceSQ}\textbf{SQ}\\
Size $r[w]$ & 1.25 & 3 & 1.25 & 1.75 & 1.25 & 5 & 1.25 & 2.25 & 1.5 & 1.5 & 1.5 & 4 & 5 & 1.375 & 3\\
Pedestrians & 5 & 5 & 4 & 4 & 3 & 3 & 3 & 4 & 5 & 4 & 5 & 4 & 3 & 4 & 4\\
\midrule
\rowcolor{NavTraceBand}
Frozen success & 47 & 32 & 37 & 41 & 39 & 37 & 57 & 35 & 38 & 39 & 26 & 41 & 44 & 72 & 44\\
Overwrite success & 45 & 44 & 22 & 42 & 41 & 68 & 56 & 79 & 58 & 27 & 33 & 74 & 85 & 97 & 58\\
\rowcolor{NavTraceBand}
Cache-only success & 57 & 43 & 43 & 54 & 47 & 77 & 53 & 98 & 62 & 44 & 19 & 54 & 83 & 96 & 51\\
\sys{} success & 55 & 42 & 37 & 54 & 39 & 75 & 53 & 98 & 68 & 44 & 26 & 59 & 77 & 96 & 52\\
\midrule
\rowcolor{NavTraceBand}
\sys{} route & CS & CS & base & HS & base & CS & HS & CS & CS & HS & base & CS & CS & CS & CS\\
After service & keep & keep & keep & keep & keep & keep & keep & keep & keep & keep & keep & keep & keep & keep & keep\\
\midrule
\rowcolor{NavTraceBand}
Overwrite repair (min) & 22.1 & 20.8 & 58.5 & 51.8 & 48.5 & 14.8 & 44.3 & 16.4 & 22.1 & 52.8 & 69.1 & 18.2 & 15.3 & 0.0 & 18.3\\
Cache-only repair (min) & 0.0 & 0.0 & 0.0 & 0.0 & 0.0 & 0.0 & 0.0 & 0.0 & 0.0 & 0.0 & 0.0 & 0.0 & 0.0 & 0.0 & 0.0\\
\rowcolor{NavTraceBand}
\sys{} repair (min) & 0.0 & 0.0 & 0.0 & 0.0 & 0.0 & 0.0 & 0.0 & 0.0 & 0.0 & 0.0 & 0.0 & 0.0 & 0.0 & 0.0 & 0.0\\
\bottomrule
\end{tabular}
\endgroup
\end{table}

\FloatBarrier
\subsection{Complete fixed-case forgetting sequence}
\label{app:forgetting-details}

To distinguish shared-parameter forgetting from changing request difficulty, one policy is evaluated on the same fixed circle A and hallway B case blocks after each A--B--A update. Both 100-case blocks lie within their respective training ranges; B training uses no A replay. Circle swept success falls after B training and recovers after retraining A. The historical strict score is retained separately, not substituted for swept success.

\begin{table}[H]
\centering\FTCRAppTableStyle
\caption{The same 100 cases per family after each update. Swept success is primary; strict success is a historical diagnostic.}
\label{tab:appendix-A10}\label{tab:forgetting-complete}
\begin{tabular}{@{\extracolsep{\fill}}lrrrr@{}}
\toprule
Stage & A swept & B swept & A strict & B strict\\
\midrule
Base & 16 & 29 & 19 & 70\\
After learning A & 80 & 15 & 100 & 32\\
After learning B & 22 & 30 & 23 & 92\\
After learning A again & 75 & 27 & 99 & 57\\
\bottomrule
\end{tabular}
\end{table}

The paired A comparison after learning B contains 61 losses and three gains, giving exact McNemar $p=4.74\times10^{-15}$~\citep{mcnemar1947note}. This describes the fixed cases, not independent training chains. CARVE retention is checked separately using later serving-model identities; no constant CARVE score curve is inferred.

\FloatBarrier
\subsection{Chip request order, work, and stored specialists}
\label{app:chip-stream-accounting}\label{app:admission-rules}

E2 follows eight circuits through three arrivals each. The reported pretraining corpus includes adaptec2, adaptec3, and bigblue1; the other circuits are ibm07, ibm10, ibm13, ibm15, and ibm18. Five methods process 30 matched trials per request (3,600 service trials). A request supplies macro sizes, connections, and a placement region. Repair uses only the current circuit, with at most three attempts per session, a 4,800-second budget per attempt, and checkpoints every 600 seconds.

E2 serves before repairing a new family. Later requests retrieve an admitted specialist; rejection consumes repair work without adding a serving file. Post-admission extensions are logged but do not avoid E2 training. The separate E2-X study instead tests reuse first. Admission and extension rules are stated once below; quality and timing definitions are centralized in Appendix~\ref{app:measurement-details}.

\paragraph{Recorded admission and extension rule.}
A candidate needs at least 5\% lower median HPWL and a one-sided paired signed-rank $p<0.05$ on 30 matched trials. For extension, the implementation sorts comparisons by p, applies successive Holm thresholds, and stops when either the current threshold or the effect requirement fails. The set contains eight targets per E2 specialist, available specialists per E2-X target, or six imports per E-T receiver target. This is the executed combined rule, not a substituted procedure or a lower bound on future gain.

\paragraph{Request continuation.}
Tables~\ref{tab:chip-stream-first12} and~\ref{tab:appendix-A11} give requests 1--12 and 13--24, respectively. Displayed request r is archive slot $r-1$; splits do not reset the system. CARVE's storage after a request counts registered files only. E2 extension credentials add no model copy; each admitted specialist is 39,129,374 bytes. Service is timed separately from repair.

\begin{table}[H]
\centering
\begingroup
\fontsize{7}{8.2}\selectfont
\setlength{\tabcolsep}{1.0pt}
\renewcommand{\arraystretch}{1.15}
\setlength{\aboverulesep}{1.2pt}
\setlength{\belowrulesep}{1.2pt}
\definecolor{ChipTraceBand}{HTML}{F3F5F8}
\caption{Circuit revisits, requests 1 to 12. Quality uses 30 matched trials per method. Repair cost includes training and checks; storage follows each request.}
\label{tab:chip-stream-first12}
\begin{tabular}{@{}>{\raggedright\arraybackslash}p{.16\linewidth}*{12}{>{\centering\arraybackslash}p{\dimexpr(\linewidth-.16\linewidth-24\tabcolsep)/12\relax}}@{}}
\toprule
\textbf{Request} & \textbf{1} & \textbf{2} & \textbf{3} & \textbf{4} & \textbf{5} & \textbf{6} & \textbf{7} & \textbf{8} & \textbf{9} & \textbf{10} & \textbf{11} & \textbf{12}\\
\midrule
Circuit & ibm07 & ibm18 & ibm15 & ibm13 & bigblue1 & ibm13 & ibm18 & bigblue1 & ibm07 & ibm10 & adaptec3 & ibm13\\
Arrival & 1 & 1 & 1 & 1 & 1 & 2 & 2 & 2 & 2 & 1 & 1 & 3\\
\midrule
\multicolumn{13}{@{}l}{\textbf{HPWL gain over Frozen (\%)}}\\
\rowcolor{ChipTraceBand}
Overwrite & 0.0 & $-0.6$ & $0.8$ & $4.5$ & $4.8$ & $-7.0$ & $-11.5$ & $3.4$ & $-10.3$ & $8.0$ & $-37.0$ & $4.9$\\
RS & 0.0 & 0.0 & 0.0 & $8.2$ & $6.5$ & $0.4$ & $-20.7$ & $9.3$ & $-7.6$ & $8.0$ & $-15.4$ & $3.9$\\
\rowcolor{ChipTraceBand}
Cache-only & 0.0 & 0.0 & 0.0 & 0.0 & 0.0 & $17.0$ & $-2.8$ & $9.2$ & $11.2$ & 0.0 & 0.0 & $9.9$\\
\sys{} & 0.0 & 0.0 & 0.0 & 0.0 & 0.0 & $20.7$ & $17.6$ & $9.5$ & $16.3$ & 0.0 & 0.0 & $16.0$\\
\midrule
\multicolumn{13}{@{}l}{\textbf{Repair cost and \sys{} storage}}\\
\rowcolor{ChipTraceBand}
Overwrite cost (min) & 89.8 & 98.4 & 80.1 & 80.1 & 18.9 & 80.1 & 80.1 & 19.3 & 82.0 & 80.1 & 3.4 & 80.1\\
\sys{} cost (min) & 90.0 & 85.5 & 85.4 & 85.5 & 23.9 & 0.0 & 0.0 & 0.0 & 0.0 & 85.6 & 83.1 & 0.0\\
\rowcolor{ChipTraceBand}
Storage (MiB) & 37.3 & 74.6 & 112.0 & 149.3 & 186.6 & 186.6 & 186.6 & 186.6 & 186.6 & 223.9 & 223.9 & 223.9\\
\bottomrule
\end{tabular}
\endgroup
\end{table}

\begin{table}[H]
\centering
\begingroup
\fontsize{7}{8.2}\selectfont
\setlength{\tabcolsep}{1.0pt}
\renewcommand{\arraystretch}{1.15}
\setlength{\aboverulesep}{1.2pt}
\setlength{\belowrulesep}{1.2pt}
\definecolor{ChipTraceBand}{HTML}{F3F5F8}
\caption{Circuit revisits, requests 13 to 24, continuing Table~\ref{tab:chip-stream-first12}. Quality uses 30 matched trials per method. Repair cost includes training and checks; storage follows each request.}
\label{tab:appendix-A11}\label{tab:chip-stream-last12}
\begin{tabular}{@{}>{\raggedright\arraybackslash}p{.16\linewidth}*{12}{>{\centering\arraybackslash}p{\dimexpr(\linewidth-.16\linewidth-24\tabcolsep)/12\relax}}@{}}
\toprule
\textbf{Request} & \textbf{13} & \textbf{14} & \textbf{15} & \textbf{16} & \textbf{17} & \textbf{18} & \textbf{19} & \textbf{20} & \textbf{21} & \textbf{22} & \textbf{23} & \textbf{24}\\
\midrule
Circuit & ibm18 & ibm10 & adaptec2 & bigblue1 & adaptec2 & ibm15 & adaptec3 & ibm15 & adaptec2 & ibm10 & adaptec3 & ibm07\\
Arrival & 3 & 2 & 1 & 3 & 2 & 2 & 2 & 3 & 3 & 3 & 3 & 3\\
\midrule
\multicolumn{13}{@{}l}{\textbf{HPWL gain over Frozen (\%)}}\\
\rowcolor{ChipTraceBand}
Overwrite & $-18.1$ & $-3.6$ & $-3.4$ & $8.8$ & $2.1$ & $-7.9$ & $-44.0$ & $-13.3$ & $-1.6$ & $-24.9$ & $-20.1$ & $-17.0$\\
RS & $-9.8$ & $31.2$ & $-1.6$ & $0.8$ & $4.3$ & $11.3$ & $1.0$ & $-6.5$ & $-8.2$ & $-3.1$ & $8.4$ & $-11.0$\\
\rowcolor{ChipTraceBand}
Cache-only & $-7.1$ & $8.1$ & 0.0 & $14.8$ & $5.4$ & $31.2$ & $3.2$ & $24.0$ & $10.3$ & $15.5$ & $7.7$ & $11.5$\\
\sys{} & $17.3$ & $9.8$ & 0.0 & $15.2$ & 0.0 & $36.7$ & 0.0 & $21.7$ & 0.0 & $8.9$ & 0.0 & $12.3$\\
\midrule
\multicolumn{13}{@{}l}{\textbf{Repair cost and \sys{} storage}}\\
\rowcolor{ChipTraceBand}
Overwrite cost (min) & 80.1 & 81.8 & 3.9 & 19.0 & 0.0 & 10.5 & 0.0 & 0.0 & 0.0 & 96.1 & 0.0 & 89.7\\
\sys{} cost (min) & 0.0 & 0.0 & 82.7 & 0.0 & 0.0 & 0.0 & 0.0 & 0.0 & 0.0 & 0.0 & 0.0 & 0.0\\
\rowcolor{ChipTraceBand}
Storage (MiB) & 223.9 & 223.9 & 223.9 & 223.9 & 223.9 & 223.9 & 223.9 & 223.9 & 223.9 & 223.9 & 223.9 & 223.9\\
\bottomrule
\end{tabular}
\endgroup
\end{table}

Eight shared base files for the main extension sweep are charged once at slot 0; later sweeps add specialist files only. The allocation is retained in \path{timing_file_allocation.csv}. Timers exclude loading and setup as detailed in Appendix~\ref{app:cost-accounting}; these are not end-to-end latency or resident-memory measurements.

\FloatBarrier
\subsection{Complete placement medians and training work}
\label{app:chip-all-medians}

Table~\ref{tab:appendix-A12} consolidates the complete service-median and per-request training records. All 24 requests remain, including base first arrivals and harmful transfers. E2-X shares service seeds but changes training seeds and integration code; its column is a follow-up, not a sixth contemporaneous E2 arm. Values are rounded only for display; original CSV precision determines gains and ranks.

\begin{table}[H]
\centering\FTCRAppDenseStyle\setlength{\tabcolsep}{1.5pt}
\caption{Complete chip service medians and recorded training work. HPWL is in thousands and lower is better; training is in minutes and includes all recorded attempts.}
\label{tab:appendix-A12}\label{tab:chip-all-work}\label{tab:all-chip-slots}
\begin{tabular*}{\linewidth}{@{\extracolsep{\fill}}rlrrrrrrrrrrr@{}}
\toprule
& & & \multicolumn{6}{c}{Median HPWL ($\times10^3$)} & \multicolumn{4}{c}{Training (min)}\\\cmidrule(lr){4-9}\cmidrule(l){10-13}
Req. & Circuit & Visit & B1 & OV & RS & B3 & CARVE & E2-X & OV & RS & B3 & CARVE\\
\midrule
1 & ibm07 & 1 & 221.6 & 221.6 & 221.6 & 221.6 & 221.6 & 221.6 & 89.8 & 83.6 & 80.1 & 80.1\\
2 & ibm18 & 1 & 566.4 & 570.0 & 566.4 & 566.4 & 566.4 & 566.4 & 98.4 & 80.1 & 94.8 & 80.1\\
3 & ibm15 & 1 & 267.5 & 265.4 & 267.5 & 267.5 & 267.5 & 267.5 & 80.1 & 80.1 & 80.1 & 80.1\\
4 & ibm13 & 1 & 265.4 & 253.5 & 243.6 & 265.4 & 265.4 & 265.4 & 80.1 & 152.3 & 117.7 & 80.1\\
5 & bigblue1 & 1 & 329.9 & 314.2 & 308.5 & 329.9 & 329.9 & 329.9 & 18.9 & 18.5 & 19.0 & 18.9\\
6 & ibm13 & 2 & 275.2 & 294.4 & 274.1 & 228.3 & 218.2 & 238.9 & 80.1 & 235.1 & 0.0 & 0.0\\
7 & ibm18 & 2 & 561.9 & 626.8 & 678.4 & 577.6 & 463.2 & 487.4 & 80.1 & 80.1 & 0.0 & 0.0\\
8 & bigblue1 & 2 & 322.8 & 311.8 & 292.7 & 293.2 & 292.2 & 316.5 & 19.3 & 0.0 & 0.0 & 0.0\\
9 & ibm07 & 2 & 232.0 & 255.8 & 249.6 & 206.1 & 194.2 & 199.5 & 82.0 & 80.1 & 0.0 & 0.0\\
10 & ibm10 & 1 & 1346.4 & 1239.3 & 1239.3 & 1346.4 & 1346.4 & 1346.4 & 80.1 & 80.1 & 80.1 & 80.1\\
11 & adaptec3 & 1 & 3917.9 & 5366.3 & 4521.7 & 3917.9 & 3917.9 & 3917.9 & 3.4 & 80.1 & 80.2 & 80.2\\
12 & ibm13 & 3 & 261.0 & 248.1 & 250.8 & 235.1 & 219.2 & 242.9 & 80.1 & 0.0 & 0.0 & 0.0\\
13 & ibm18 & 3 & 551.7 & 651.8 & 605.5 & 590.9 & 456.1 & 481.9 & 80.1 & 187.9 & 0.0 & 0.0\\
14 & ibm10 & 2 & 1293.8 & 1340.1 & 889.8 & 1189.4 & 1167.2 & 1254.6 & 81.8 & 0.0 & 0.0 & 0.0\\
15 & adaptec2 & 1 & 10852.9 & 11217.7 & 11030.1 & 10852.9 & 10852.9 & 10852.9 & 3.9 & 80.1 & 80.1 & 80.1\\
16 & bigblue1 & 3 & 344.5 & 314.0 & 341.8 & 293.7 & 292.1 & 323.4 & 19.0 & 18.9 & 0.0 & 0.0\\
17 & adaptec2 & 2 & 11005.6 & 10770.2 & 10536.3 & 10409.9 & 11005.6 & 9229.9 & 0.0 & 80.1 & 0.0 & 0.0\\
18 & ibm15 & 2 & 266.7 & 287.9 & 236.5 & 183.6 & 168.9 & 187.0 & 10.5 & 31.0 & 0.0 & 0.0\\
19 & adaptec3 & 2 & 3930.7 & 5661.2 & 3892.4 & 3805.0 & 3930.7 & 3930.7 & 0.0 & 151.0 & 0.0 & 0.0\\
20 & ibm15 & 3 & 254.9 & 288.8 & 271.5 & 193.6 & 199.7 & 192.8 & 0.0 & 0.0 & 0.0 & 0.0\\
21 & adaptec2 & 3 & 10893.2 & 11066.3 & 11784.9 & 9769.4 & 10893.2 & 7916.0 & 0.0 & 104.3 & 0.0 & 0.0\\
22 & ibm10 & 3 & 1334.7 & 1666.4 & 1375.9 & 1128.5 & 1215.9 & 1232.8 & 96.1 & 80.1 & 0.0 & 0.0\\
23 & adaptec3 & 3 & 4347.5 & 5222.7 & 3984.2 & 4013.4 & 4347.5 & 4347.5 & 0.0 & 132.6 & 0.0 & 0.0\\
24 & ibm07 & 3 & 228.0 & 266.9 & 253.1 & 201.9 & 199.9 & 200.6 & 89.7 & 80.1 & 0.0 & 0.0\\
\bottomrule
\end{tabular*}
\end{table}

Table~\ref{tab:chip-attempts} reports placement regressions and training
outcomes. A regression is a request whose median HPWL exceeds Frozen
by more than $5\%$. Each session permits up to three attempts and ends
when an attempt produces a candidate. Rejection after evaluation does
not permit another attempt.

\begin{table}[H]
\centering\FTCRAppTableStyle
\caption{Placement regressions, training, and admission outcomes over 24 requests.}
\label{tab:appendix-A14}\label{tab:chip-attempts}
\begin{tabular*}{\linewidth}{@{\extracolsep{\fill}}lrrrrr@{}}
\toprule
Method & \shortstack{$>5\%$ worse\\(/24)} & Sessions & Attempts & Candidates & Admissions\\
\midrule
Frozen & 0 & 0 & 0 & 0 & ---\\
Overwrite & 11 & 19 & 32 & 16 & Not tested\\
RS & 7 & 20 & 33 & 18 & 13/18\\
Cache-only & 1 & 8 & 11 & 8 & Not tested\\
\sys{} & 0 & 8 & 8 & 8 & 6/8\\
\bottomrule
\end{tabular*}
\end{table}

Training totals include all recorded attempts, including those ending
without a candidate. A circuit-registration issue in our integration
affected some baseline attempts, so the comparison also reflects
implementation effects.

Candidate outcomes also vary: E2 rejects adaptec2 at 7.04\% gain
($p=.079$), while E2-X accepts another candidate at 21.22\%
($p=.0073$). E2 accepts ibm18 at 19.78\%, whereas three RS candidates
for ibm18 are rejected. These comparisons vary in models and sometimes
code, not seed alone. A rejection applies to the evaluated candidate,
not to every model that could be trained for that family.

\FloatBarrier
\subsection{Compared deployment methods}
\label{app:deployment-methods}

Table~\ref{tab:deployment-methods} gives the update and model-selection
rules. Repair retention uses Frozen, Overwrite, and CARVE.
Caching and admission adds Cache-only, and chip placement also includes RS.
Adapted CompoNet uses the same caching-and-admission arrival sequences
with separate case blocks. The table describes the main and
supplementary comparisons, not the earlier synthetic training protocol.

\begin{table}[H]
\centering\FTCRAppTableStyle
\caption{Compared deployment methods. Each method serves the triggering request
before training or admission. A repair can affect only later requests.}
\label{tab:deployment-methods}
\begin{tabularx}{\linewidth}{@{}p{.13\linewidth}>{\raggedright\arraybackslash}p{.41\linewidth}>{\raggedright\arraybackslash}X@{}}
\toprule
Method & Model update & Model used on later requests\\
\midrule
Frozen & Never trains or changes the base. & Always uses the frozen base.\\
\addlinespace[2pt]
Overwrite & Fine-tunes the currently active model and replaces it without an admission test. & Uses the latest updated model for every family. Earlier models are not retrieved.\\
\addlinespace[2pt]
Cache-only & Trains a fresh copy of the base once for each family and saves the candidate without testing it. & Retrieves that family's saved candidate. Uses the base before a candidate exists.\\
\addlinespace[2pt]
RS & Trains a fresh copy of the base and tests it against the base. Acceptance replaces the active model; rejection keeps the previous one. & Uses one active model across families, without a family cache.\\
\addlinespace[2pt]
CompoNet\newline(adapted) & Adds a PPO-trained module while freezing the base and earlier modules; no performance admission test. & Uses the family's module with its frozen prefix. Before repair, uses the frozen-base sampling wrapper.\\
\addlinespace[2pt]
CARVE & Keeps earlier specialists unchanged. Tests a new candidate against the base before saving it. Existing specialists may earn new families by evaluation. & Retrieves a specialist authorized for the requested family. Otherwise uses the base. A rejected family is not trained again within the stream.\\
\bottomrule
\end{tabularx}
\end{table}

\FloatBarrier
\subsection{Domain-specific contracts}
\label{app:domain-contracts}

\paragraph{Choice of admission margins.}
We selected the navigation and placement margins empirically from
experiments conducted during development. Navigation requires at least
15 additional swept successes on 100 paired cases, equivalent to an
absolute increase of 15 percentage points. Placement requires a relative
reduction of at least 5\% in median HPWL on 30 paired trials, together
with the stated paired test and multiplicity rule.
The placement effect margin is distinct from the test level $\alpha=0.05$.

Evaluating a candidate before adoption has precedent in both empirical
promotion rules and statistical policy improvement.
AlphaGo Zero uses a win rate above 55\% over 400 games against the
current best player as its promotion criterion~\citep{silver2017mastering}.
High Confidence Policy Improvement allows the user to specify a
performance requirement and statistical error tolerance
separately~\citep{thomas2015hcpi}.
The cited work supports the use of an explicit adoption criterion;
CARVE's numerical margins remain choices from our development experiments.
Our analysis holds the recorded margins fixed and does not establish an
optimal choice. Section~\ref{sec:system} and
Appendix~\ref{app:theory-application} state the assumptions required for
statistical guarantees and their relationship to the empirical rules.

Table~\ref{tab:domain-contracts} gives the task definitions and repair rules
used in each domain. In the navigation alarm, $k$ counts swept successes
among 100 episodes. The synthetic chip alarm belongs to the earlier study,
while E2 uses the model or family identity. The chip selector row describes
main E2. E2-X and E-T instead use the recorded earliest passing specialist
as described in Appendices~\ref{app:chip-reuse-details}
and~\ref{app:chip-receiver-details}. The budget $r=3$ permits at most three
attempts in one session, including the first attempt. Exact admission and
extension procedures remain in Appendix~\ref{app:admission-rules}.

\begin{table}[H]
\centering
\FTCRAppTableStyle
\caption{Domain-specific task definitions and repair rules.}
\label{tab:domain-contracts}
\begin{tabular}{>{\raggedright\arraybackslash}p{0.17\linewidth}>{\raggedright\arraybackslash}p{0.37\linewidth}>{\raggedright\arraybackslash}p{0.37\linewidth}}
\toprule
Component & Crowd navigation & Chip macro placement \\
\midrule
Family key & Environment geometry and crowd-behavior model & Circuit identity with macro and netlist specification \\
Governed score & Swept-contact success; higher is better & Half-perimeter wirelength (HPWL); lower is better \\
Alarm & Statistical: $100$-episode block; Clopper--Pearson rule equivalent to $k\leq88$ & Synthetic families: random-anchor normalized-improvement statistic below $0.7585$.  Real circuits: no statistical alarm fired; identity-triggered (no credential $\Rightarrow$ alarm), see Appendix~\ref{sec:audit-alarm} \\
Candidate & SARL specialist trained from the frozen base on the current family & ChiPFormer specialist fine-tuned from the frozen base on the current circuit; $4{,}800$\,s budget, at most $r{=}3$ training attempts \\
Admission & Paired $100$-episode block; at least $15/100$ swept successes over the base & Paired $30$-episode block; median HPWL improved by at least $5\%$ and one-sided Wilcoxon signed-rank $p<0.05$ \\
Extension exam & Before training, on the alarmed family's paired block, same $15$ rule; a pass routes the family & Before training (exam-routing arm) or after admission with credentials logged (main arm); Holm over the simultaneous decisions \\
Selector $\sigma$ & Own-family specialist first, then earliest credential & Own-family specialist first (the only case exercised) \\
Fallback & Byte-identical frozen SARL base & Byte-identical frozen ChiPFormer base \\
\bottomrule
\end{tabular}
\end{table}

\FloatBarrier
\subsection{Experimental terms and reported quantities}
\label{app:terminology}

Table~\ref{tab:terminology} brings together definitions used in the text,
tables, and logs. The definitions do not add new metrics or alter the
recorded measurements.

\begingroup
\FTCRAppTableStyle
\setlength{\tabcolsep}{4pt}
\renewcommand{\arraystretch}{1.07}
\setlength{\LTleft}{0pt}
\setlength{\LTright}{0pt}
\setlength{\LTcapwidth}{\linewidth}
\begin{longtable}{@{}>{\raggedright\arraybackslash}p{.22\linewidth}>{\raggedright\arraybackslash}p{\dimexpr.78\linewidth-2\tabcolsep\relax}@{}}
\caption{Terms used in the experiments. Requests, training attempts, and evaluation
cases are different counting units.}
\label{tab:terminology}\\
\toprule
Term & Meaning in the recorded experiments\\
\midrule
\endfirsthead
\toprule
Term & Meaning in the recorded experiments\\
\midrule
\endhead
\endfoot
\bottomrule
\endlastfoot
Family & A navigation geometry and pedestrian model, or a chip circuit with its macro and netlist specification. Circle crossing with ORCA+ and circle crossing with SFM are different families.\\
Scene setting and case & A setting fixes parameters within a navigation family, such as scene size and pedestrian count. A case also has an evaluation index. The index alone does not identify a case.\\
Request & In the main comparisons, one family arrival contains 100 navigation cases or 30 chip placement trials per method.\\
Stream and repair epoch & A stream is the ordered request list. Each main stream starts a separate repair epoch, with the same base and an empty repair state. A later stream does not continue an earlier one.\\
Arrival seed & The random seed that generates the request order and scene settings. It is distinct from the seed used to train a model. The displayed caching and admission stream uses fixed per-family request counts.\\
\midrule
Base and specialist & The base is the original frozen policy. A specialist is a separate trained policy. An CARVE specialist remains unchanged after admission.\\
Admission and credential & Admission evaluates a candidate against the base under the recorded rule. A credential records permission to use a specialist for a specific family and metric. Passing on one family does not authorize another.\\
Extension and reuse & An extension evaluates a saved specialist on another family without changing its weights. A pass adds permission for later requests. Reuse means serving a request with a saved model.\\
Rejection and fallback & Rejection excludes a candidate from CARVE service. CARVE serves the base for that family and records that it will not train again within the epoch. RS instead retains its previous active model.\\
Training session and attempt & A session is one repair training job. In main chip E2, it permits up to three attempts and ends when an attempt produces a candidate. An attempt that produces no candidate is distinct from a candidate rejected after evaluation. A session with no candidate after its allowed attempts is exhausted.\\
\midrule
Swept success & The robot reaches its goal within 90 seconds without swept-volume contact. Goal reaching without the contact condition is a different score.\\
HPWL and gain & HPWL is estimated wirelength, with lower values preferred. Request gain is $100(1-m/m_0)$, where $m$ and $m_0$ are the method and Frozen medians on matched trials. Positive gain means shorter estimated wiring.\\
Mean rank and worse requests & Rank is one plus the number of methods with strictly lower median HPWL on the same request. Ties share a rank. Mean rank averages the request ranks. The $>5\%$ worse count uses requests with median HPWL above $1.05m_0$.\\
Mean $\pm$ SD and pp & The main stream SD is variation across matched requests, not independent training runs. Navigation differences in pp are percentage-point differences in success rate. Receiver intervals use a separate definition in Appendix~\ref{app:paired-variation}.\\
Repair and service time & Repair time sums training and recorded admission and extension checks, including failures. Ordinary service is separate. Chip trial timers exclude model loading and do not measure full request latency.\\
Checkpoint storage & Logical sizes of final registered specialist files. It is not RAM or GPU memory. Duplicate extension copies are included in the cache comparison, with the full counting rules in Appendix~\ref{app:cost-accounting}.\\
\end{longtable}
\endgroup

\FloatBarrier
\subsection{Repair protocol details}
\label{app:protocol-details}

\paragraph{Scope and repair epochs.}
A repair epoch is one deployment run with fixed versions of the base,
family resolver, metric, and repair contract. Each experimental request
stream starts a separate epoch. Runs with the same versions still maintain
separate credentials, stopped-family records, and ledgers. Replacing the
base, changing the family definitions, or changing a decision rule starts
a new epoch. Earlier specialists and their evidence remain available.
An earlier credential applies to its original base and contract. A saved
specialist can obtain a credential under a new base by passing another
evaluation, without changing its weights.

\paragraph{Required interfaces.}
The operator supplies an executable frozen base $M_0$ and a deterministic
resolver $\rho(x)\in\mathcal{F}\cup\{\bot\}$. The resolver assigns request
descriptor $x$ to a declared family or reports that it is outside scope.
The operator also supplies a score evaluator, a candidate trainer, and
storage for model artifacts and evidence. The score $s_f(M,z)$ evaluates
model $M$ on instance $z$ from family $f$, with larger values preferred.
A quantity with smaller preferred values, such as HPWL, uses a fixed
orientation or an equivalent predicate for that metric. The trainer uses
the base and current-family data. No source training set, source replay
buffer, or samples from previous families are required.

The resolver uses observable task keys. Navigation keys specify geometry
and the pedestrian model. Placement keys specify the circuit and its
macro and netlist description. The experiments do not learn a resolver
for ambiguous or continuously changing conditions. An unresolved request
uses the base without approximate matching to a saved specialist.

\paragraph{State and model selection.}
At time $t$, the deployment maintains
\begin{equation}
\begin{gathered}
\mathcal{M}_t=\{M_0,M_1,\ldots,M_{J_t}\},\qquad
\Gamma_t\subseteq\{1,\ldots,J_t\}\times\mathcal{F},\\
\mathcal{T}_t\subseteq\mathcal{F},\qquad
\mathcal{X}_t\subseteq\mathcal{F},\qquad
\mathcal{L}_t=\text{append-only audit ledger}.
\end{gathered}
\label{eq:carve-state-details}
\end{equation}
The model library $\mathcal{M}_t$ contains immutable artifacts. A credential
$(j,f)\in\Gamma_t$ authorizes specialist $M_j$ for family $f$.
One specialist can serve several families, and several specialists can
qualify for the same family. The contract fixes a deterministic selector
$\sigma(f,\Gamma_t)$ for the latter case. Navigation and the main placement
study prefer a specialist trained on the requested family, then the
earliest-admitted eligible credential. The placement reuse follow-up selects
the earliest-admitted specialist among those that pass. The receiving
operator selects the passing import admitted earliest at the source,
as specified in Appendices~\ref{app:chip-reuse-details}
and~\ref{app:chip-receiver-details}.

The tombstone set $\mathcal{T}_t$ records families whose new candidate
received a complete, valid rejection. The exhausted set $\mathcal{X}_t$
records families whose training session ended without a candidate.
Exhaustion is a training failure rather than a performance verdict. Both
sets prevent another training session for that family in the epoch.
A failed extension exam alone does not create a tombstone, since a new
candidate may still be trained. An incomplete gate record cannot add an
admitted artifact, credential, or performance tombstone.

\paragraph{Serving and monitoring requests.}
The system serves each request before any repair triggered by that
request. For a credentialed family, the selector retrieves an authorized
unchanged specialist without training or another admission test.
For an unresolved, tombstoned, or exhausted family, the system serves the
exact frozen base. An eligible family without a credential also uses the
base for its current request. A repair can therefore improve only later
requests, as in the other repairing methods in Section~\ref{sec:eval}.

After base service for an eligible family, the adapter computes
\begin{equation}
a_t=\mathsf{Alarm}_{f_t}(M_0;E_t,\mathcal{C}),
\label{eq:carve-alarm-details}
\end{equation}
where $E_t$ is the monitoring block and $\mathcal{C}$ is the frozen
contract. The contract states the score, sampling unit, decision rule,
and validity checks. It also states the uncertainty calculation when one
is used. A complete, valid alarm opens a repair event. A no-alarm result
leaves the family eligible for monitoring at its next arrival.
Incomplete monitoring evidence commits no state change and leaves the
base route in place. Navigation uses a statistical alarm. The placement
experiments on benchmark circuits instead use the recorded model or
family identity rules, as documented in Appendix~\ref{sec:audit-alarm}.

If the contract monitors a credentialed family, a later alarm is logged
without another training session in that epoch. A credential is removed
only through an explicit logged revocation or an epoch change.

\paragraph{Extension exams and their order.}
For family $f$, the contract fixes an ordered set $\mathcal{H}_f$ of saved
specialists eligible for an extension exam. Each specialist is evaluated
against $M_0$ using the same paired admission rule as a new candidate.
A pass adds $(j,f)$ to $\Gamma_t$ without changing any weights.
The contract fixes the comparison order and the correction for multiple
comparisons before evaluation. Appendix~\ref{app:admission-rules} records
the executed placement rule, including its effect requirement and
successive Holm thresholds. The comparison set depends on the study.

Navigation and the placement reuse follow-up examine saved models before
training. A passing specialist can then serve later requests without new
training. If none passes, an eligible family receives a training session.
The main placement study instead gives each new circuit its own training
session and examines an admitted specialist on all eight circuits afterward.
Those additional credentials are recorded as transfer evidence and do not
avoid training or change routing in that study.
Appendix~\ref{app:chip-reuse-details} keeps the two schedules separate.

\paragraph{Models received from another operator.}
An imported specialist begins without permission to serve the receiver's
families. The receiver compares it with its own frozen base on a local
paired evaluation block. A pass creates a local credential, while a failed
exam leaves the base route available. A source credential does not replace
the receiver's evaluation. Requests assigned a different family require
another local exam even when they resemble a source task. Reuse therefore
requires local evaluation data but no source circuits, source training
histories, or weight updates. The receiving-operator study evaluates its
selected model on a separate service block
(Appendix~\ref{app:chip-receiver-details}).

\paragraph{Candidate generation and evaluation data.}
Each candidate starts from a byte-identical copy of the frozen base:
\begin{equation}
C_f=\mathsf{Train}_f(M_0,D_f^{\mathrm{tr}};\xi_f).
\label{eq:carve-training-details}
\end{equation}
The data $D_f^{\mathrm{tr}}$ come from the current family, and $\xi_f$
specifies training settings, resources, and random seeds. The trainer may
use full fine-tuning, parameter-efficient adaptation, online reinforcement
learning, or another local update. It must produce a serializable policy
and support evaluation under the contract. Candidates cannot serve while
training or evaluation is incomplete. Training leaves the base and every
saved specialist unchanged, so retaining earlier models requires no
previous-family examples.

The intended contract separates admission data $D_f^{\mathrm{dev}}$ from
candidate training. Its statistical guarantees require the validation
assumptions in Section~\ref{sec:system}. Appendix~\ref{sec:audit-blocks}
describes case allocation and the available evidence for the recorded
navigation runs.

\paragraph{Paired admission and statistical interpretation.}
For evaluation cases $z_1,\ldots,z_n$, candidate and base produce paired
differences
\begin{equation}
d_i=s_f(C_f,z_i)-s_f(M_0,z_i).
\label{eq:carve-paired-details}
\end{equation}
The contract fixes an effect estimator $\widehat{\Delta}_f$, a minimum
margin $m_f$, an uncertainty procedure when used, and an acceptance region.
A contract with both an effect requirement and a statistical test takes
the form
\begin{equation}
\mathsf{Admit}_f(C_f,M_0;D_f^{\mathrm{dev}})
=\mathbf{1}\{\widehat{\Delta}_f\ge m_f
\ \text{and}\ p_f\le\alpha_f\}.
\label{eq:carve-admission-details}
\end{equation}
Here $p_f$ is the test's $p$-value and $\alpha_f$ is its declared threshold.
A contract can instead specify a paired margin-only rule or an equivalent
confidence-bound rule. Score orientation, comparison unit, effect margin,
error criterion, evaluation block, and multiplicity policy are fixed
before examining the candidate.

The effect estimator can summarize paired differences or relative change
in medians. Placement measures one minus the ratio of the candidate and
base HPWL medians and applies a one-sided Wilcoxon signed-rank test to
paired outcomes. Navigation applies its paired improvement margin without
that test. Appendix~\ref{app:domain-contracts} gives the domain thresholds,
and Appendix~\ref{app:admission-rules} gives the executed placement rule.

A passing evaluation authorizes only the tested family and metric.
Under a valid level-$\alpha_f$ test and its sampling assumptions, requiring
rejection of the declared null controls erroneous rejection of that null
at the corresponding level, subject to the declared multiplicity rule.
An empirical effect margin alone supplies no such error control.
Neither rule establishes universal policy safety, optimality, or a minimum
gain on every future instance. In particular, the empirical margin is not
automatically a lower bound on expected improvement. Section~\ref{sec:system}
and Appendix~\ref{app:theory-application} separately state the conditions
for the bounded-loss guarantees and their relation to the experiments.

\paragraph{Training attempts and complete decisions.}
Each family receives at most one bounded training session in an epoch.
A session allows up to $r$ attempts under a predeclared retry schedule.
Only an attempt that ends without a candidate may be retried with the
next seed. Producing a candidate ends the session, and the candidate then
receives one gate decision. A performance rejection cannot start another
attempt. Using all attempts without producing a candidate exhausts the
session.

Acceptance serializes and hashes the candidate, marks it immutable, and
records its credential, contract identity, and complete evaluation evidence
in one atomic commit. Only then can the candidate serve requests.
The admitted specialist may obtain further credentials through extension
exams without changing its parameters. A complete failed gate adds the
family to $\mathcal{T}_t$ and leaves $M_0$ serving. The rejected artifact
may remain in an archive for analysis but cannot have an active credential.

An exhausted session adds the family to $\mathcal{X}_t$ and records each
attempt's failure evidence. No candidate was evaluated, and the family
uses $M_0$ for the rest of the epoch. Incomplete evaluation evidence cannot
authorize a model or record a performance rejection. The navigation
contract used $r=1$ and would halt the epoch after a training failure;
none occurred. Placement permits up to three attempts per session and distinguishes
sessions that produce no candidate from candidates rejected after evaluation.

\paragraph{Automation and provenance.}
The operator defines scope, family resolution, metrics, decision rules,
the trainer, resource budgets, retry limits, and epoch changes. Once the
contract is fixed, \sys{} executes the request and repair procedure.
It resolves and serves the request, then monitors the outcome under the
contract. When repair is triggered, it runs the specified extension or
training sequence and evaluates the candidate. It records acceptance,
rejection, or exhaustion before changing subsequent routing.
The validator and admission procedure control entry into service.

The ledger records request and family identifiers, base and candidate
hashes, contract and code versions, training and evaluation block
identifiers, and seeds. It also records per-instance and aggregate scores,
uncertainty calculations, decisions, selected routes, wall-clock costs,
and failure evidence. Model identities support operational rollback,
and evaluation records support recomputation of the reported decisions.

\FloatBarrier
\subsection{Construction properties and resource accounting}
\label{app:protocol-properties}

\paragraph{Preserving models and controlling service.}
Within an epoch, training a candidate cannot modify $M_0$ or a previously
admitted specialist. When inference settings are also fixed, later
training therefore leaves the saved policy unchanged. Deleting an artifact
is a storage failure. Compressing or replacing it creates a different
artifact that requires a new identity and authorization. Preserving
parameters does not keep realized performance constant across changing
evaluation cases or task distributions.

A candidate cannot serve before its credential is committed. An
out-of-scope, tombstoned, or exhausted family uses the exact frozen base.
The construction guarantees concern preservation and selection. They do
not establish adequate base performance or improvement by a specialist
on every request. Section~\ref{sec:system} gives conditions for preserving
expected performance relative to the base. Later alarms on a credentialed
family are logged without further training within the same epoch.

Without an explicit logged revocation or an epoch change, $\Gamma_t$ only
gains authorized pairs. An extension can therefore increase the recorded
scope of reuse without adding model weights. The original artifact and
its evaluation record remain available for earlier authorized uses.

\paragraph{Training work and storage.}
Let $K$ count distinct alarmed families in a finite epoch, let $J$ count
accepted newly trained specialists, and let $r$ be the maximum number of
training attempts per session. The one-session rule gives
\begin{equation}
N_{\mathrm{sessions}}\le K,\qquad
N_{\mathrm{runs}}\le rK,\qquad J\le K.
\label{eq:carve-budget-details}
\end{equation}
A method that retrains after each alarm can start several sessions for
the same family. Equation~\ref{eq:carve-budget-details} bounds sessions
and training attempts rather than extension exams or their cost.
Every permitted attempt remains subject to its declared resource budget.

For a local library containing the base and $J$ accepted repairs, with
one stored copy of each model, logical serving storage is
\begin{equation}
B_{\mathrm{serve}}=B(M_0)+\sum_{j=1}^{J}B(M_j)+O(|\Gamma|),
\label{eq:carve-storage-details}
\end{equation}
where $B(M)$ is model size and $|\Gamma|$ counts credentials.
Imports and duplicate physical files require additional storage.
Appendix~\ref{app:cost-accounting} defines the measured file totals.
Full checkpoints, adapters, compression, and loading policies have
different costs, and transformed artifacts require new identities and
authorization. The bound depends on the declared scope. The model library
can grow as the number of families grows.

\FloatBarrier
\section{Reuse and transfer records}
\setcounter{table}{0}\setcounter{figure}{0}
\label{app:reuse-records}

The records below distinguish checks that authorize reuse from later service
with the same saved model. Request counts do not represent independent
training runs.

\subsection{Navigation extension decisions}
\label{app:nav-extension-records}

The four displayed successful extensions in Table~\ref{tab:appendix-B1} add permissions without updating weights. They include ORCA+/SFM changes and square crossing. The subsequent-service summary uses the same retrospective subset outside the permitted training index range as Appendix~\ref{sec:audit-blocks}. The three detailed streams contain 35 own-family and 21 extension requests with unchanged model identities, not new training repetitions.

\begin{table}[H]
\centering\FTCRAppTableStyle
\caption{Successful navigation extension exams. Time covers these successful events only; full repair costs also include failed checks.}
\label{tab:appendix-B1}\label{tab:nav-extensions}
\begin{tabular*}{\linewidth}{@{\extracolsep{\fill}}lrlrrr@{}}
\toprule
Stream & Slot & Target family & Specialist & Base & Exam time (s)\\
\midrule
Fixed arrivals & 5 & F-CS & 96/100 & 79/100 & 88.5\\
Sampled arrivals & 3 & F-CS & 95/100 & 26/100 & 302.8\\
Caching and admission & 3 & F-CO & 96/100 & 5/100 & 351.9\\
Caching and admission & 9 & F-SQ & 56/100 & 30/100 & 223.6\\
\bottomrule
\end{tabular*}
\end{table}

\begin{table}[H]
\centering\FTCRAppTableStyle
\caption{Later service using retained specialists, relative to matched Frozen (pp). Ranges show observed requests, not confidence intervals.}
\label{tab:appendix-B2}
\begin{tabular*}{\linewidth}{@{\extracolsep{\fill}}lrrrrrr@{}}
\toprule
& \multicolumn{3}{c}{Own training family} & \multicolumn{3}{c}{Extension family}\\\cmidrule(lr){2-4}\cmidrule(l){5-7}
Stream & $n$ & Mean & Range & $n$ & Mean & Range\\
\midrule
Fixed arrivals & 17 & 38.18 & [\ensuremath{-}6, 88] & 6 & 41.50 & [20, 67]\\
Sampled arrivals & 14 & 58.29 & [25, 86] & 6 & 47.17 & [18, 66]\\
Caching and admission & 4 & 18.25 & [\ensuremath{-}4, 36] & 9 & 27.11 & [8, 65]\\
\bottomrule
\end{tabular*}
\end{table}

All 56 requests are from that retrospective subset. Earlier development exposure is not excluded.

\paragraph{Earlier navigation checks.}
\label{app:local-case-check}\label{app:engine-grid}
Development evaluations also compared one specialist on three case banks
and on ORCA+ and SFM pedestrian grids. They used fixed models and did not
authorize service or use the reserved final SFM exam. Their complete records
remain in \path{analysis/appendix_original_upload.tex}.

\FloatBarrier
\subsection{Placement reuse before training}
\label{app:chip-reuse-details}

The reuse study follows the main circuit order and service seeds, but checks
saved specialists before training. The base serves each first arrival.
A passing check permits the earliest admitted qualifying specialist to serve
later arrivals unchanged. Five circuits reuse the \texttt{ibm07} specialist,
while \texttt{ibm07}, \texttt{adaptec3}, and \texttt{adaptec2} receive training
sessions. Training seeds and circuit integration code differ between the
executions, so their comparison does not isolate extension timing.

\begin{table}[H]
\centering\FTCRAppDenseStyle
\caption{New-circuit checks and later outcomes. Jobs count training sessions;
service gains average the two revisits relative to Frozen.}
\label{tab:appendix-B5}\label{tab:e2x-routing}\label{tab:chip-reuse}
\begin{tabular*}{\linewidth}{@{\extracolsep{\fill}}lrcrrrlr@{}}
\toprule
& \multicolumn{2}{c}{Check saved model}
& \multicolumn{2}{c}{Train each circuit}
& \multicolumn{3}{c}{Check saved models first}\\
\cmidrule(lr){2-3}\cmidrule(lr){4-5}\cmidrule(l){6-8}
Circuit & \shortstack{Exam gain\\(\%)} & Result
& Jobs & \shortstack{Service gain\\(\%)}
& Jobs & Model & \shortstack{Service gain\\(\%)}\\
\midrule
ibm07    & ---  & None & 1 & 14.29 & 1 & ibm07    & 13.01\\
ibm18    & 14.1 & Pass & 1 & 17.44 & 0 & ibm07    & 12.96\\
ibm15    & 26.9 & Pass & 1 & 29.17 & 0 & ibm07    & 27.14\\
ibm13    & 10.0 & Pass & 1 & 18.36 & 0 & ibm07    & 10.08\\
bigblue1 & 10.1 & Pass & 1 & 12.35 & 0 & ibm07    & 4.05\\
ibm10    & 6.8  & Pass & 1 & 9.34  & 0 & ibm07    & 5.34\\
adaptec3 & $-3.1$ & Fail & 1 & 0.00 & 1 & Base     & 0.00\\
adaptec2 & $-2.6$ & Fail & 1 & 0.00 & 1 & adaptec2 & 21.73\\
\bottomrule
\end{tabular*}
\end{table}

Figure~\ref{fig:chip-reuse} includes all 24 requests, including first
arrivals and base fallbacks. For request $r$, the gain is $g_r=100(1-m_r/b_r)$,
where $m_r$ and $b_r$ are the method and Frozen service medians.
Mean gains across all 24 requests are 8.41\% for training each circuit and
7.86\% for checking saved models first. Recorded training times are 9.657
and 4.005 hours, respectively. Including checks gives 10.362 and 4.247 hours
under Appendix~\ref{app:cost-accounting}'s timing definitions.
The \texttt{adaptec2} gains at requests 17 and 21 come from new training.

On the ten revisits to the five reused circuits, mean gains are 17.33\% for
own-circuit training and 11.91\% for reuse. Reuse gives lower gains on nine
of those ten requests. Appendix Table~\ref{tab:appendix-D8} retains every
paired revisit. The results therefore support reduced training work with
a placement-quality trade-off.

\paragraph{Extension coverage and recorded predictions.}
\label{app:chip-transfer-coverage}
The main study's six saved specialists pass 23 model--family checks across
six targets, with no pass on \texttt{adaptec2} or \texttt{adaptec3}.
Those checks occurred after training and did not avoid any training session.
The reuse study did not meet its recorded prediction of mean rank no worse
than Cache-only. Its six-method mean rank was 2.375, compared with 2.208
for Cache-only. The complete coverage and prediction records remain in
\path{analysis/appendix_original_upload.tex}.

\FloatBarrier
\subsection{Receiving-operator selection and service}
\label{app:chip-receiver-details}

The receiver imports six admitted source checkpoints and tests them against
its base on seven new IBM targets and an \texttt{adaptec1} control. Each
comparison uses 30 paired exam trials. The earliest source-admitted passing
model is selected, and its service uses a separate block of 30 seeds.
The receiver trains no weights and uses no source circuit files or training
trajectories. The experiment separates operator task sets without enforcing
privacy isolation. A prior census evaluated \texttt{ibm06} through
\texttt{ibm18}; five IBM circuits entered the source study, and
\texttt{ibm06} was an integration probe, leaving seven receiver targets.

Training the six imported source specialists cost 419.17 minutes in total,
paid once per specialist. Receiver checks cost 20.71 minutes across all
eight targets, with no receiver training. Each target requires one base and
six import evaluations of 30 trials each. No matched receiver training arm
was run, so the source time does not estimate receiver training time or time
saved. Source training includes work inside the training process, while
receiver check timers exclude model loading and setup
(Appendix~\ref{app:cost-accounting}). The reported 5.76\% mean gain averages
all eight target gains, including both base fallbacks.

\begin{table}[H]
\centering\FTCRAppTableStyle
\caption{All 48 local exam gains (\%). Bold entries mark passing under the recorded combined effect-and-Holm rule; signs alone do not determine admission.}
\label{tab:appendix-B11}
\begin{tabular*}{\linewidth}{@{\extracolsep{\fill}}lrrrrrr@{}}
\toprule
Target & ibm07 & ibm18 & ibm15 & ibm13 & bigblue1 & ibm10\\
\midrule
ibm08 & $0.51$ & $-2.89$ & $-2.62$ & $-1.18$ & $-15.62$ & $-1.63$\\
ibm09 & $\mathbf{6.45}$ & $\mathbf{6.09}$ & $-5.64$ & $\mathbf{15.02}$ & $-1.38$ & $2.38$\\
ibm11 & $\mathbf{9.31}$ & $\mathbf{7.35}$ & $-1.00$ & $\mathbf{14.06}$ & $-3.91$ & $2.85$\\
ibm12 & $3.82$ & $3.47$ & $-1.09$ & $\mathbf{8.71}$ & $-5.81$ & $4.15$\\
ibm14 & $3.73$ & $\mathbf{5.54}$ & $-1.63$ & $1.74$ & $-2.76$ & $-0.09$\\
ibm16 & $\mathbf{14.14}$ & $\mathbf{8.48}$ & $1.15$ & $\mathbf{7.75}$ & $-1.32$ & $\mathbf{6.53}$\\
ibm17 & $\mathbf{10.87}$ & $\mathbf{15.07}$ & $-5.84$ & $8.22$ & $6.80$ & $5.98$\\
adaptec1 & $-13.05$ & $-5.70$ & $-3.03$ & $-8.35$ & $-8.60$ & $-6.37$\\
\bottomrule
\end{tabular*}
\end{table}

Table~\ref{tab:appendix-B11} contains exam effects and decisions.
Fourteen passing model--target pairs cover six targets. Table~\ref{tab:appendix-B12}
reports the selected routes and their later service, including both fallbacks.

\begin{table}[H]
\centering\FTCRAppDenseStyle
\caption{Local selection and later service. All base fallbacks and both intervals containing zero remain. Intervals condition on the selected model and target.}
\label{tab:appendix-B12}\label{tab:transfer}\label{tab:et-service-intervals}
\begin{tabular*}{\linewidth}{@{\extracolsep{\fill}}lllrrr@{}}
\toprule
Target & Passing imports & Selected & Gain (\%) & 95\% interval & Lower/equal/higher\\
\midrule
ibm08 & --- & base & $0.00$ & --- & ---\\
ibm09 & ibm13, ibm07, ibm18 & ibm07 & $3.61$ & $[-2.00, 7.09]$ & 18/0/12\\
ibm11 & ibm13, ibm07, ibm18 & ibm07 & $10.21$ & $[4.28, 14.91]$ & 22/0/8\\
ibm12 & ibm13 & ibm13 & $9.95$ & $[2.02, 15.51]$ & 21/0/9\\
ibm14 & ibm18 & ibm18 & $8.92$ & $[2.46, 14.21]$ & 23/0/7\\
ibm16 & ibm07, ibm13, ibm18, ibm10 & ibm07 & $8.06$ & $[3.66, 11.98]$ & 21/0/9\\
ibm17 & ibm18, ibm07 & ibm07 & $5.33$ & $[-3.36, 11.27]$ & 19/0/11\\
adaptec1 & --- & base & $0.00$ & --- & ---\\
\bottomrule
\end{tabular*}
\end{table}

\paragraph{Checks and later service.}
\label{app:transfer-check-followup}
Figure~\ref{fig:chip-check-service} compares the exam with later service for
all five reused circuits and all six selected receiver imports. Some later
gains fall below the 5\% required in the check, so passing does not ensure
the same gain on later cases. All selected receiver imports have positive
service point estimates, but the intervals for \texttt{ibm09} and
\texttt{ibm17} include zero.

On \texttt{ibm09}, the selected \texttt{ibm07} model improves HPWL by 6.45\%
in the exam and 3.61\% in service. The unselected \texttt{ibm13} import has a
15.02\% exam gain but no corresponding service evaluation, so the two
blocks cannot estimate selector regret. Appendix~\ref{app:paired-variation}
defines the conditional paired-bootstrap intervals.

\begin{figure}[H]
\centering
\begin{minipage}[t]{.495\linewidth}\centering
\includegraphics[width=\linewidth]{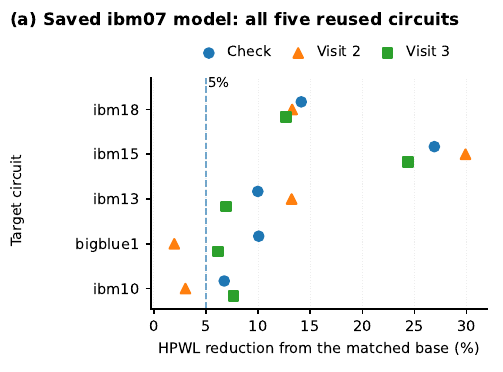}
\end{minipage}\hfill
\begin{minipage}[t]{.495\linewidth}\centering
\includegraphics[width=\linewidth]{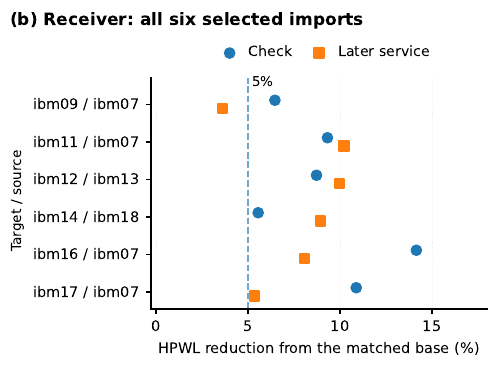}
\end{minipage}
\caption{Checking and later using the same saved model on the same target.
The left panel includes every circuit served by reuse of \texttt{ibm07};
each later marker represents one 30-trial revisit. The right panel includes
every selected import at the receiver, evaluated on a separate 30-trial block.
The dashed 5\% line applies to the check, not to later service.}
\label{fig:chip-check-service}
\end{figure}

\FloatBarrier
\section{Ablations and component records}
\setcounter{table}{0}\setcounter{figure}{0}
\label{app:ablation-records}

The following records examine admission decisions and model selection.
Comparisons using the same candidate separate deployment decisions from
training variation. Full traces retain both beneficial and harmful outcomes.

\FloatBarrier
\subsection{Navigation decisions and subsequent service}
\label{app:nav-gate-followup}

\paragraph{A matched candidate, with deployment or fallback.}
Saving a candidate does not establish that it should replace the base.
We examine the hallway candidate in the displayed caching and admission
stream. Its independently trained Cache-only and \sys{} model states
match. Cache-only serves the candidate, whereas \sys{} rejects it and
serves the base. We compare all five later hallway requests, each with
100 matched cases. Admission uses a different 100-case block and requires
at least 15 more successes than the base. The candidate gains 10 and
therefore fails this requirement.

\begin{table}[H]
\centering\begingroup
\fontsize{7.5}{8.8}\selectfont
\setlength{\tabcolsep}{4pt}
\renewcommand{\arraystretch}{1.09}
\caption{Serving the matched hallway candidate or retaining the base in the
displayed stream. Later scores pool five requests with 100 cases each.
The final column is \sys{} minus Cache-only in percentage points.}
\label{tab:nav-candidate-decisions}
\begin{tabular*}{\linewidth}{@{\extracolsep{\fill}}lrrrr@{}}
\toprule
Task & \shortstack{Candidate gain\\in check (pp)}
& \shortstack{Cache-only: use candidate\\Later successes /500}
& \shortstack{\sys{}: keep base\\Later successes /500}
& \shortstack{\sys{} minus\\Cache-only (pp)}\\
\midrule
Bottleneck hallway & $10$ & 168 & 195 & $5.4$\\
\bottomrule
\end{tabular*}
\endgroup
\end{table}

Table~\ref{tab:nav-candidate-decisions} compares serving the same candidate
with retaining the base. Rejection avoids a loss over these later hallway
requests. The result concerns this candidate and arrival history; it does
not establish that rejection always improves service. The later evaluation
below uses the retrospective subset outside the permitted training index
range (Appendix~\ref{sec:audit-blocks}), so its request counts and mean
gains differ from the five-request totals above.

We compare the seven admission decisions in the three detailed streams with later evaluation of the same candidates. Accepted states match CARVE service. The rejected caching-and-admission hallway state matches the independently trained Cache-only model. The stream with sampled arrivals has no Cache-only arm, so its rejected hallway candidate uses the executed recheck outside the permitted training index range. A fallback score is never substituted for the rejected candidate's performance. Table~\ref{tab:appendix-C1} reports all seven decisions and their later evaluations.

\begin{table}[H]
\centering\FTCRAppDenseStyle
\caption{Seven navigation gates and later own-family evaluation on the retrospective subset outside the permitted training index range. Gains are successes per 100; blocks do not represent independent training runs.}
\label{tab:appendix-C1}\label{tab:gate-followup}\label{tab:nav-reject-audit}
\begin{tabular*}{\linewidth}{@{\extracolsep{\fill}}llrlrrlr@{}}
\toprule
Stream & Family & Gate slot & Decision & Gate gain & Blocks & Evaluated via & Later mean gain\\
\midrule
Fixed arrivals & F-CO & 1 & Accept & $59$ & 12 & CARVE & $52.92$\\
Fixed arrivals & F-HO & 3 & Accept & $15$ & 5 & CARVE & $2.80$\\
Sampled arrivals & F-CO & 1 & Accept & $89$ & 14 & CARVE & $58.29$\\
Sampled arrivals & F-HO & 6 & Reject & $10$ & 1 & \RecheckStudy{} & $6.00$\\
Caching and admission & F-CS & 1 & Accept & $64$ & 2 & CARVE & $36.00$\\
Caching and admission & F-HO & 7 & Reject & $10$ & 4 & B3 & $-5.00$\\
Caching and admission & F-HS & 8 & Accept & $22$ & 2 & CARVE & $0.50$\\
\bottomrule
\end{tabular*}
\end{table}

In caching and admission, the rejected hallway model changes success by
$-49,+15,+6,+8,-7$ over its five later requests. The negative average
coexists with gains on three requests. On request 10, Frozen and CARVE
score 51/100, Cache-only scores 2/100, and Overwrite scores 28/100 on the
same recorded cases. The sampled-arrival recheck still shows a useful
candidate rejected by the fixed margin (Appendix~\ref{sec:audit-nav}).

\FloatBarrier
\subsection{Navigation evaluation scope}
\label{sec:audit-blocks}

The three detailed streams contain 300 service files and 30,000 episode
executions. Admission and service use different index blocks. The admission
block and some service indices lie within the permitted training range.
A case is identified by its family, scene settings, and index, so overlapping
indices alone do not establish that the same cases were used in training.

The subset outside the permitted training index range covers 76 shared
request positions and 24,900 method--episode executions. The other 5,100
executions fall within that range. The archive retains all 420 files,
including the additional arrival. Counts refer to executions, not unique
environments.

Table~\ref{tab:appendix-C3} applies the same retrospective filter to every
method while retaining the executed models and routes. Earlier requests
may have affected later decisions, and prior development exposure is not
excluded. This analysis describes existing service outcomes.

\begin{table}[H]
\centering\FTCRAppTableStyle
\caption{Service outcomes on the retrospective subset outside the permitted training index range. Original training and routing remain unchanged.}
\label{tab:appendix-C3}\label{tab:nav-existing-exam}
\begin{tabular*}{\linewidth}{@{\extracolsep{\fill}}lrrrrrr@{}}
\toprule
Stream & Requests & Episodes/arm & Frozen & Overwrite & B3 & CARVE\\
\midrule
Fixed arrivals & 26 & 2,600 & .366 & .631 & --- & .711\\
Sampled arrivals & 29 & 2,900 & .333 & .699 & --- & .712\\
Caching and admission & 21 & 2,100 & .425 & .577 & .554 & .576\\
\bottomrule
\end{tabular*}
\end{table}

The displayed subset contains 56 credential hits, with 35 on the five
accepted candidates' training families and 21 after extension. Every
accepted candidate has later own-family service with a matching model state.
The caching and admission subset retains the avoided loss on request 10,
but CARVE's full-stream advantage over Overwrite is absent. Its admitted
constricted-hallway model also scores 53 against the base's 57 on request 22.
The subset establishes neither equivalence nor absence of harm.

\FloatBarrier
\subsection{Rechecking the hallway model rejected under sampled arrivals}
\label{sec:audit-nav}

The fixed hallway model rejected under sampled arrivals is re-evaluated
against the base on two paired blocks. One block lies within the permitted
training index range; the other lies outside it, although its base results
already existed. Both yield modest positive swept gains
(Table~\ref{tab:appendix-C4}). Using a base score from another block would
overstate the effect. These are two evaluations, not new training runs or
permanent bounds on improvement.

\begin{table}[H]
\centering\FTCRAppTableStyle
\caption{Matched rechecks of the hallway model rejected under sampled arrivals. Gained/lost cases preserve the pairing; scores use saved outcomes.}
\label{tab:appendix-C4}\label{tab:ec-audit}
\begin{tabular*}{\linewidth}{@{\extracolsep{\fill}}lrrrrr@{}}
\toprule
Case block & Base & Candidate & Gain & Gained cases & Lost cases\\
\midrule
Within permitted training range & 32/100 & 39/100 & $7$ & 13 & 6\\
Outside permitted training range & 28/100 & 34/100 & $6$ & 15 & 9\\
\bottomrule
\end{tabular*}
\end{table}

\FloatBarrier
\subsection{Placement admission and model selection}
\label{app:chip-gates}

\paragraph{Admission on fixed candidates.}
\label{app:chip-admission-rule-audit}
We compare the recorded rule with a rule requiring only 5\% observed
median-HPWL improvement on the same eight trained candidates.
The recorded rule also requires a passing paired statistical check.
Removing that check admits one additional candidate, \texttt{adaptec2},
whose 7.04\% observed gain has $p=0.079$. The comparison identifies a
change in admission decisions. Neither rejected candidate has a later
candidate-service evaluation, so the comparison cannot establish the
service effect of admitting it. The margin remains fixed throughout the
audit, and a passing check does not establish a population lower bound of
5\% gain. Appendix~\ref{app:admission-rules} defines the recorded rule.

\begin{table}[H]
\centering\FTCRAppDenseStyle
\caption{All placement candidates under an effect-only rule and the recorded
rule. Checks use 30 paired trials. Later gains use the admitted model;
rejected candidates were not evaluated in later service.}
\label{tab:appendix-C5}\label{tab:e2-gates}\label{tab:chip-admission-rule-audit}
\begin{tabular*}{\linewidth}{@{\extracolsep{\fill}}rlrrrllr@{}}
\toprule
Slot & Circuit & \shortstack{Exam gain\\(\%)} & \shortstack{One-sided\\$p$}
& \shortstack{Paired\\wins} & \shortstack{Effect\\only}
& \shortstack{Recorded\\rule} & \shortstack{Later gain (\%)\\visit 2 / visit 3}\\
\midrule
0 & ibm07 & $11.53$ & $9.96\times10^{-7}$ & 27/30 & Accept & Accept & $16.27 / 12.31$\\
1 & ibm18 & $19.78$ & $1.77\times10^{-8}$ & 28/30 & Accept & Accept & $17.57 / 17.32$\\
2 & ibm15 & $30.56$ & $1.95\times10^{-5}$ & 25/30 & Accept & Accept & $36.69 / 21.65$\\
3 & ibm13 & $19.24$ & $1.86\times10^{-9}$ & 29/30 & Accept & Accept & $20.71 / 16.01$\\
4 & bigblue1 & $10.38$ & $9.31\times10^{-10}$ & 30/30 & Accept & Accept & $9.48 / 15.22$\\
9 & ibm10 & $12.83$ & $1.18\times10^{-5}$ & 24/30 & Accept & Accept & $9.78 / 8.90$\\
10 & adaptec3 & $4.60$ & $0.657$ & 12/30 & Reject & Reject & Not evaluated\\
14 & adaptec2 & $7.04$ & $0.079$ & 17/30 & Accept & Reject & Not evaluated\\
\bottomrule
\end{tabular*}
\end{table}

\paragraph{Model selection on arriving circuits.}
Table~\ref{tab:chip-scope-check} shows four consecutive requests for which
RS uses its \texttt{bigblue1} specialist and \sys{} retrieves each circuit's
earlier validated repair.

The complete RS trace in Table~\ref{tab:rs-full-trace} includes every
request. All seven requests with HPWL more than 5\% worse than Frozen use
a model on a different circuit from its admission circuit. Ten other such
uses improve on Frozen, so the record does not show that all reuse across
circuits is harmful.

\begin{table}[H]
\centering\FTCRAppDenseStyle
\caption{Complete RS trace. Positive HPWL change is worse than the matched base. Same/Other refers to the model's earlier admission family.}
\label{tab:appendix-C6}\label{tab:rs-full-trace}
\begin{tabular*}{\linewidth}{@{\extracolsep{\fill}}rllrlrl@{}}
\toprule
Slot & Requested & Model origin & Admit slot & Scope & HPWL increase & Harm $>5\%$\\
\midrule
0 & ibm07 & base & --- & Base & $0.00\%$ & No\\
1 & ibm18 & base & --- & Base & $0.00\%$ & No\\
2 & ibm15 & base & --- & Base & $0.00\%$ & No\\
3 & ibm13 & ibm15 & 2 & Other & $-8.20\%$ & No\\
4 & bigblue1 & ibm13 & 3 & Other & $-6.48\%$ & No\\
5 & ibm13 & bigblue1 & 4 & Other & $-0.39\%$ & No\\
6 & ibm18 & bigblue1 & 4 & Other & $20.73\%$ & Yes\\
7 & bigblue1 & bigblue1 & 4 & Same & $-9.33\%$ & No\\
8 & ibm07 & bigblue1 & 4 & Other & $7.61\%$ & Yes\\
9 & ibm10 & ibm07 & 8 & Other & $-7.95\%$ & No\\
10 & adaptec3 & ibm10 & 9 & Other & $15.41\%$ & Yes\\
11 & ibm13 & ibm10 & 9 & Other & $-3.91\%$ & No\\
12 & ibm18 & ibm10 & 9 & Other & $9.75\%$ & Yes\\
13 & ibm10 & ibm10 & 9 & Same & $-31.23\%$ & No\\
14 & adaptec2 & ibm10 & 9 & Other & $1.63\%$ & No\\
15 & bigblue1 & adaptec2 & 14 & Other & $-0.79\%$ & No\\
16 & adaptec2 & bigblue1 & 15 & Other & $-4.26\%$ & No\\
17 & ibm15 & adaptec2 & 16 & Other & $-11.34\%$ & No\\
18 & adaptec3 & adaptec2 & 16 & Other & $-0.97\%$ & No\\
19 & ibm15 & adaptec3 & 18 & Other & $6.49\%$ & Yes\\
20 & adaptec2 & adaptec3 & 18 & Other & $8.19\%$ & Yes\\
21 & ibm10 & adaptec2 & 20 & Other & $3.08\%$ & No\\
22 & adaptec3 & ibm10 & 21 & Other & $-8.36\%$ & No\\
23 & ibm07 & adaptec3 & 22 & Other & $11.00\%$ & Yes\\
\bottomrule
\end{tabular*}
\end{table}

\paragraph{Trace verification and scope.}
\label{sec:audit-rs}
Every non-base route has a 16-hex hash prefix matching the preceding accepted
candidate in the logs. The trace contains three base routes, two routes on
the admission circuit, and 19 routes on other circuits. Adding a scope check
could change later training and model selection, and that alternative was
not run. The trace therefore does not isolate the causal benefit of the
check. Hash agreement verifies recorded identities without supplying missing
checkpoint files.

\FloatBarrier
\subsection{Development records}
\label{app:navigation-development}\label{app:training-recipes}
\label{app:adapter-checks}\label{app:anchor-check}
\label{app:routing-check}\label{app:checkpoint-review}\label{app:penalty-check}
\label{app:action-check}\label{app:scene-screening}
\label{app:chip-development}\label{app:chip-screening}\label{app:synthetic-chip}
\label{app:early-chip-transfer}\label{app:synthetic-overwrite}
\label{app:own-vs-imported}\label{app:chip-qualification}

Earlier recipe, scene-screening, implementation, and synthetic-placement
studies remain together in \path{analysis/appendix_original_upload.tex}.
The archived block retains successful, failed, incomplete, and unfavorable
results under their original metrics and protocols. They are development
records and do not add independent runs to the benchmark comparisons.

Navigation recipe screening selected the local trainer and used separate
confirmation seeds. Its historical goal-reaching criterion differs from
the main swept-success metric. A three-seed base-penalty study reduced
hallway performance on every seed. Routing tests, unsuccessful action
diagnostics, and exploratory scene grids are retained in the same archive.

Earlier synthetic placement studies used different service order, baselines,
and score definitions. Their transfer checks include both useful and harmful
outcomes. A later overwrite study did not confirm the predicted home-task
degradation. Earlier circuit-specific qualification runs used different
candidates from the benchmark studies, so their outcomes are kept separate.

\FloatBarrier
\section{Measurement definitions and statistical analyses}
\setcounter{table}{0}\setcounter{figure}{0}
\label{app:measurement-details}

This section specifies what is counted and what variation means. Calculations use retained execution records, without new policy training or simulation. Training runs, requests, evaluation trials, and files are different units and are not added into one sample count.

\FloatBarrier
\subsection{Repair-time and checkpoint-storage accounting}
\label{app:cost-accounting}

\paragraph{Timing boundaries.}
Navigation repair time sums training, admission, and extension process times, including failed checks; ordinary service and driver work are excluded. The detailed tables contain ten original method--stream totals; the unchanged archive also preserves the four totals from the additional arrival. Placement training time includes all recorded attempts and evaluation inside the training process. Separate evaluation-trial time starts after model loading and trainer construction: seeding, placement, and the recorded \path{wall_s} are timed, with values rounded to .01 seconds. These timers omit Python startup, deserialization, device setup, overwritten evaluation outputs, and interrupted work absent from retained files. No complete cold/hot loading, peak CPU/GPU memory, throughput, or end-to-end latency profile was collected.

\begin{table}[H]
\centering\FTCRAppTableStyle
\caption{Exact-phase CARVE navigation totals, rounded after summation. Checks include failed extensions; repair excludes ordinary service.}
\label{tab:appendix-D1}\label{tab:nav-cost-components}
\begin{tabular*}{\linewidth}{@{\extracolsep{\fill}}lrrrrr@{}}
\toprule
Stream & Trainings & Train (h) & Checks (h) & Repair (h) & Checks/repair\\
\midrule
Fixed arrivals & 2 & 1.394 & 0.340 & 1.734 & 19.6\%\\
Sampled arrivals & 2 & 1.301 & 0.294 & 1.596 & 18.4\%\\
Caching and admission & 3 & 2.235 & 0.729 & 2.964 & 24.6\%\\
\bottomrule
\end{tabular*}
\end{table}

\begin{table}[H]
\centering\FTCRAppTableStyle
\caption{Placement training and separately timed gate/extension work. Totals include all recorded training attempts; shared base evaluation files are counted once per study.}
\label{tab:appendix-D2}\label{tab:chip-cost-components}
\begin{tabular*}{\linewidth}{@{\extracolsep{\fill}}lrrr@{}}
\toprule
Study / arm & Training (h) & Gate trials (min) & Extension trials (min)\\
\midrule
\ChipStudy{} B1 & 0.000 & 0.00 & 0.00\\
\ChipStudy{} OV & 19.556 & 0.00 & 0.00\\
\ChipStudy{} RS & 31.933 & 24.58 & 0.00\\
\ChipStudy{} B3 & 10.534 & 0.00 & 0.00\\
\ChipStudy{} CARVE & 9.657 & 9.44 & 32.81\\
\ReuseStudy{} & 4.005 & 6.06 & 8.46\\
\TransferStudy{} receiver & 0.000 & 0.00 & 20.71\\
\bottomrule
\end{tabular*}
\end{table}

\paragraph{Shared files and internal evaluation.}
E2's 48 extension comparisons use eight shared base and 48 specialist files, not 96 unique runs. E2-X uses seven base/specialist pairs; E-T uses eight base plus 48 import files. Base files are counted once within each study, never merged across studies. E-T service consists of 240 base and 180 selected-import trials, not 420 end-user requests. E2's eight random-anchor files add 240 diagnostic trials totaling 19.69 seconds and one metadata row per file; those are not assigned to an individual method. Fifty successful E2 training processes include 100 internal evaluation placements, and E2-X adds six. Their time is already inside training and is not added again. The sealed lost-interruption field is zero, separate from the trial timers.

\paragraph{Storage boundaries.}
Checkpoint storage sums logical sizes of final registered serving specialists from cache registries and archive-member indices. It includes navigation extension copies, but excludes the shared base, rejected candidates, training intermediates, task data, and logs. E2 extension credentials add no extra model file; its six specialists total 234,776,244 bytes (39,129,374 each), or 223.90 MiB with $1\,\mathrm{MiB}=2^{20}$ bytes. Names and sizes are listed in \path{e2_admitted_checkpoint_storage.csv}. Both cached navigation methods have a six-entry caching and admission limit. These sizes are neither measured RAM nor GPU memory.

\begin{table}[H]
\centering
\FTCRTableStyle
\caption{Final registered checkpoint files. Extra storage uses CARVE as the denominator.}
\label{tab:appendix-D4}\label{tab:cache-storage}
\begin{tabular*}{\linewidth}{@{\extracolsep{\fill}}lrrrrr@{}}
\toprule
& \multicolumn{2}{c}{Cache-only} & \multicolumn{2}{c}{CARVE} & \shortstack{Cache-only extra\\storage vs. CARVE}\\
\cmidrule(lr){2-3}\cmidrule(lr){4-5}
Experiment & Files & MiB & Files & MiB & (\%)\\
\midrule
Navigation: caching and admission & 5 & 1.872 & 4 & 1.498 & $25.0$\\
Chip E2 & 8 & 298.533 & 6 & 223.900 & $33.3$\\
\bottomrule
\end{tabular*}
\par\vspace{2pt}
{\fontsize{7}{8}\selectfont\raggedright Extra storage is $100(B_{\mathrm{Cache}}/B_{\mathrm{CARVE}}-1)$, calculated from exact byte counts.
Registered duplicate extension copies are included. The common base, rejected candidates,
training intermediates, and logs are excluded. Sizes are logical files, not RAM or GPU memory.\par}
\end{table}

\FloatBarrier
\subsection{Paired effects and uncertainty across evaluation seeds}
\label{app:paired-variation}

\paragraph{Placement summary columns.}
Each request's gain is the reduction of its 30-trial median HPWL relative to the matched Frozen median, not classification accuracy. Mean gain and sample SD use all 24 requests, including first arrivals and base fallbacks. Rank is one plus the number of methods with strictly lower median HPWL on that request; ties share a rank. Mean rank must be recomputed when the comparison set changes. A more-than-5\%-worse request has median HPWL greater than 1.05 times its matched base median. Raw HPWL is not pooled across circuits.

\paragraph{Navigation effects.}
For request $r$, $d_r=100(\widehat p_{\mathrm{CARVE},r}-\widehat p_{\mathrm{reference},r})$. Each request uses 100 paired cases, so this equals the success-count difference. The original recomputation covers the complete archive of 420 row files and 42,000 outcomes. The detailed tables use the three displayed streams, matching family, scene parameters, and case range. For each 30-request stream, sample SD is
\begin{equation}
 s_d=\sqrt{\frac{1}{29}\sum_{r=1}^{30}(d_r-\bar d)^2}.
\end{equation}

SD describes variation across requests sharing models and training history, not independent training seeds or a significance test. The Overwrite and Cache-only contrasts for caching and admission share CARVE observations and are not pooled. The same case index under different scene parameters remains a different case. Tables~\ref{tab:appendix-D5} and~\ref{tab:appendix-D6} retain the displayed stream and family effects.

\begin{table}[H]
\centering\FTCRAppTableStyle
\caption{CARVE minus reference, swept-success difference (pp), mean \ensuremath{\pm} sample SD across 30 matched requests. Positive favors CARVE; --- means not run.}
\label{tab:appendix-D5}\label{tab:nav-paired-effects}
\begin{tabular}{@{}lrr@{}}
\toprule
Stream & Overwrite & Cache-only\\
\midrule
Fixed arrivals & $8.20 \pm 12.47$ & ---\\
Sampled arrivals & $1.40 \pm 14.07$ & ---\\
Caching and admission & $1.90 \pm 19.40$ & $1.93 \pm 10.35$\\
\bottomrule
\end{tabular}
\end{table}

\begin{table}[H]
\centering\FTCRAppTableStyle
\caption{Family differences in swept success (pp). Positive favors CARVE.
H/T/L counts requests with higher, tied, or lower CARVE success;
--- means not run. These are the original batches, not the separate-case
supplementary comparison.}
\label{tab:appendix-D6}\label{tab:nav-family-overwrite-effects}
\label{tab:appendix-D7}\label{tab:nav-family-cache-effects}
\begin{tabular*}{\linewidth}{@{\extracolsep{\fill}}llrrrrrrr@{}}
\toprule
& & & \multicolumn{3}{c}{CARVE $-$ Overwrite} & \multicolumn{3}{c}{CARVE $-$ Cache-only}\\
\cmidrule(lr){4-6}\cmidrule(l){7-9}
Stream & Family & Requests & Mean & Range & H/T/L & Mean & Range & H/T/L\\
\midrule
Fixed arrivals & F-CO & 14 & $11.57$ & $[-5, 61]$ & 11/2/1 & --- & --- & ---\\
Fixed arrivals & F-CS & 7 & $0.86$ & $[-16, 6]$ & 6/0/1 & --- & --- & ---\\
Fixed arrivals & F-HO & 9 & $8.67$ & $[0, 16]$ & 8/1/0 & --- & --- & ---\\
\addlinespace[3pt]
Sampled arrivals & F-CO & 16 & $5.06$ & $[-8, 42]$ & 9/4/3 & --- & --- & ---\\
Sampled arrivals & F-CS & 7 & $-2.86$ & $[-46, 25]$ & 4/0/3 & --- & --- & ---\\
Sampled arrivals & F-HO & 7 & $-2.71$ & $[-14, 9]$ & 3/1/3 & --- & --- & ---\\
\addlinespace[3pt]
Caching and admission & F-CO & 8 & $-4.50$ & $[-84, 26]$ & 4/2/2 & $4.25$ & $[-2, 14]$ & 5/2/1\\
Caching and admission & F-CS & 5 & $3.60$ & $[-1, 19]$ & 1/3/1 & $0.00$ & $[0, 0]$ & 0/5/0\\
Caching and admission & F-HO & 6 & $6.00$ & $[-7, 23]$ & 4/0/2 & $4.50$ & $[-15, 49]$ & 2/1/3\\
Caching and admission & F-HS & 5 & $13.20$ & $[-3, 31]$ & 4/0/1 & $0.00$ & $[0, 0]$ & 0/5/0\\
Caching and admission & F-SQ & 6 & $-4.50$ & $[-15, 7]$ & 1/0/5 & $-0.50$ & $[-6, 5]$ & 2/1/3\\
\bottomrule
\end{tabular*}
\end{table}

\paragraph{Reuse versus own-family training.}
Ten second/third arrivals of five extension-served families use the same 30 base, own-specialist, and reuse seeds per request (300 matched triples). For method $a$, let $m_{a,r}$ be its request median and $b_r$ the matched base median. The gain is $g_{a,r}=100(1-m_{a,r}/b_r)$; the gap is $g_{\mathrm{reuse},r}-g_{\mathrm{own},r}$. Negative gaps mean reuse gains less, even when both improve on the base. Two visits per family are observations, not an interval. Seeds, circuit IDs, and medians are checked against the slot summaries; neither five new circuits nor ten independent training runs are bootstrapped.

\begin{table}[H]
\centering\FTCRAppDenseStyle
\caption{All ten paired revisits. The gap favors reuse when positive. Lower/equal/higher compare reused HPWL with own-specialist HPWL; they need not order marginal medians.}
\label{tab:appendix-D8}\label{tab:e2x-all-paired-effects}
\begin{tabular*}{\linewidth}{@{\extracolsep{\fill}}rlrrrr@{}}
\toprule
Slot & Family & Own gain (\%) & Reuse gain (\%) & Gap (pp) & Lower/equal/higher\\
\midrule
5 & ibm13 & $20.71$ & $13.21$ & $-7.50$ & 10/0/20\\
6 & ibm18 & $17.57$ & $13.27$ & $-4.30$ & 6/0/24\\
7 & bigblue1 & $9.48$ & $1.96$ & $-7.52$ & 0/0/30\\
11 & ibm13 & $16.01$ & $6.94$ & $-9.07$ & 8/0/22\\
12 & ibm18 & $17.32$ & $12.65$ & $-4.67$ & 11/0/19\\
13 & ibm10 & $9.78$ & $3.03$ & $-6.75$ & 10/0/20\\
15 & bigblue1 & $15.22$ & $6.13$ & $-9.08$ & 0/0/30\\
17 & ibm15 & $36.69$ & $29.89$ & $-6.80$ & 13/0/17\\
19 & ibm15 & $21.65$ & $24.39$ & $2.73$ & 17/0/13\\
21 & ibm10 & $8.90$ & $7.64$ & $-1.26$ & 13/0/17\\
\bottomrule
\end{tabular*}
\end{table}

\paragraph{Conditional transfer-service intervals.}
For fixed target $c$, the selected import and base use the same 30 unique service seeds, disjoint from the target's recorded exam seeds. The estimate is
\begin{equation}
 \widehat g_c=100\left(1-\frac{\operatorname{median}_{i}S_{ci}}{\operatorname{median}_{i}B_{ci}}\right),
\end{equation}
not the median of individual percentage reductions. Each bootstrap draws 30 indices with replacement and applies them to both vectors before recomputing the medians and ratio. Each target uses 50,000 resamples from NumPy PCG64 seed 20260910; linear interpolation at percentiles 2.5/97.5 gives the interval in Table~\ref{tab:appendix-B12}. The seed belongs to analysis, not environment execution.

The earliest-admission selector is held fixed and service is not used to choose a different import. Base routes have zero effect by identity and no specialist interval. Under independent paired draws from that target's generator, the intervals describe evaluation variation on the fixed target/model; they omit training, library selection, and between-circuit uncertainty. They are pointwise, not simultaneous, and crossing zero does not undo an admission on another block.

The analysis inputs in \path{data/statistics/} retain six stream contrasts, 26 family summaries, ten revisits, 300 matched triples, and six receiver intervals. Original routes, point estimates, and admission decisions remain unchanged.

\FloatBarrier
\subsection{Interpreting variation in the recorded results}
\label{sec:audit-variation}

Request variability, local-training replication, and conditional evaluation uncertainty are distinct. The original navigation streams and E2/E2-X revisits share evolving state; request counts do not replace independent runs \citep{agarwal2021precipice}. Recipe repetitions in \ref{app:navigation-development} repeat local training, not the deployment system. E-T instead conditions on a fixed target and selected import. Its intervals neither repeat Holm admission nor measure selector regret. These variability analyses were added retrospectively; the two intervals containing zero and the limitations of the original caching and admission study remain reported \citep{cawley2010overfitting}.

\FloatBarrier
\subsection{Repair triggers on benchmark circuits}
\label{sec:audit-alarm}

Benchmark-circuit repair is triggered by model/family identity. The earlier random-placement diagnostic did not flag those circuits in the pre-execution check; its logged scores are not the evidence triggering repair. The placement studies evaluate training, admission, scope, reuse, and fallback, not a validated OOD detector. Navigation instead triggers repair from measured swept success.

\FloatBarrier
\subsection{Recomputation and evidence boundaries}
\label{app:recomputation}

The source archive retains all 420 navigation method--request records,
including the additional history in Appendix~\ref{app:additional-arrival}.
The detailed displays use 300 records. Chip medians, gates, and the complete
RS trace remain unchanged. File manifests and hashes check consistency
of the supplied files, not outcomes of unexecuted methods. The audits do
not load policies, tune rules, or issue credentials. Rerunning policies
requires the corresponding checkpoints and environments. Per-file timing
records remain in the source archive.

\FloatBarrier
\endgroup

\clearpage
\section{Proofs and validation requirements}
\label{app:theory}

The analysis concerns the expected loss on a declared family distribution.
It separates the evidence needed to authorize a specialist from the rules
that preserve and reuse that specialist. All logarithms below are natural.

\subsection{Proof of Proposition~\ref{prop:retained-risk}}
\label{app:theory-retention}

Fix one repair epoch. For each in-scope family $f$, the distribution $P_f$
and loss function $\ell_f$ remain unchanged. The base and each stored
specialist retain their complete inference behavior, including any
randomization procedure. The expected loss $R_f(M)$ and improvement
$\Delta_f(M)$ are defined in Section~\ref{sec:system}.
The probability in Proposition~\ref{prop:retained-risk} concerns the
training and validation history. It does not assert that a specialist
beats the base on every sampled instance.

\paragraph{Account for each new credential.}
Index validation rounds by $k=1,2,\ldots$. The history
$\mathcal{G}_{k-1}$ contains all information available before round $k$,
including the chosen candidates and their target families. Formally,
$\mathcal{G}_{k-1}$ is a sigma-field, meaning a collection of events
whose outcomes are known at that time. The allowance $\alpha_k$ is
chosen from that history before the new validation data are examined.
The event $B_k$ occurs if round $k$ issues at least one credential
$(j,f)$ with $\Delta_f(M_j)<0$. If several pairs are examined together,
$B_k$ includes all credentials issued in that round.
The epoch starts with no credentials, or includes every initial
credential in the same error accounting. After validation stops, take
$B_k$ to be the empty event and $\alpha_k=0$ for unused rounds.

\paragraph{Bound the probability of any false credential.}
The conditional error assumption gives
\begin{equation}
 \Pr(B_k)
 =\mathbb{E}\!\left[\Pr(B_k\mid\mathcal{G}_{k-1})\right]
 \leq\mathbb{E}[\alpha_k].
\end{equation}
The equality averages the conditional probability over all possible
histories. For any finite number $N$ of rounds, the union bound gives
\begin{align}
 \Pr\!\left(\bigcup_{k=1}^{N}B_k\right)
 &\leq\sum_{k=1}^{N}\Pr(B_k) \notag\\
 &\leq\mathbb{E}\!\left[\sum_{k=1}^{N}\alpha_k\right]
 \leq\delta.
\end{align}
The last inequality follows from the assumed total allowance
$\sum_{k\geq1}\alpha_k\leq\delta$ on every execution except a set of
probability zero. The events on the left grow as $N$ increases, so their
probabilities converge to the probability of their union. Therefore,
\begin{equation}
 \Pr\!\left(\bigcup_{k\geq1}B_k\right)\leq\delta.
 \label{eq:proof-all-credentials}
\end{equation}
Neither the union bound nor the averaging step requires independence
between validation rounds. Candidates and families may depend on earlier
results, provided the conditional error assumption continues to hold.

\paragraph{Apply credential validity to every service decision.}
On the complementary event in equation~\ref{eq:proof-all-credentials},
every issued credential has nonnegative population improvement. Consider
any in-scope service request $t$. If $g_t=M_0$, its expected loss equals
the base loss. Otherwise, the routing rule supplies a credential
$(j,f_t)$ for $g_t=M_j$, and hence
\begin{equation}
 R_{f_t}(M_0)-R_{f_t}(g_t)
 =\Delta_{f_t}(g_t)\geq0.
\end{equation}
The model, loss, and family distribution remain fixed, so later training
cannot alter the population comparison recorded for that credential.
The argument holds simultaneously for every service request on the same
event of probability at least $1-\delta$. Out-of-scope requests use
$M_0$ by construction and therefore have exact equality with the base.
The service bound in Proposition~\ref{prop:retained-risk} follows.

The proof identifies why preserving a model and restricting its use to
the evaluated family work together. A new credential requires error
accounting, whereas another use of an existing credential does not add a
new validation event. A change in $P_f$ or inference behavior invalidates
the preservation step. A common family name alone does not establish
that the distribution remained unchanged.

\subsection{Proof of Theorem~\ref{thm:local-reuse}}
\label{app:theory-reuse}

Fix a target distribution $P$, a bounded loss $\ell$, and a library
$M_0,M_1,\ldots,M_J$. Define
\begin{equation}
 R_P(M)=\mathbb{E}_{Z\sim P}[\ell(M,Z)],
 \qquad
 \Delta_j=R_P(M_0)-R_P(M_j),
 \qquad \Delta_0=0.
\end{equation}
The validator receives $n$ independent observations from $P$ and the
paired losses of all models on those observations. Its target information
is limited to that sample, and it may inspect the fixed models. The
library and sample size are fixed before the sample is drawn. They may
have been chosen using an earlier history, on which this analysis is
conditioned. The quantity $n^\star$ is the smallest fixed sample size
for which some validator satisfies equation~\ref{eq:reliable-reuse}
uniformly over all allowed libraries and target distributions.

\paragraph{Estimate population improvement.}
For candidate $j$ and observation $i$, set
\begin{equation}
 D_{j,i}=\ell(M_0,Z_i)-\ell(M_j,Z_i),
 \qquad
 \widehat\Delta_j=\frac1n\sum_{i=1}^{n}D_{j,i}.
\end{equation}
Both losses lie in $[0,1]$, so $D_{j,i}\in[-1,1]$ and
$\mathbb{E}[D_{j,i}]=\Delta_j$. For each fixed $j$, the variables
$D_{j,1},\ldots,D_{j,n}$ are independent because the observations are
independent. Paired losses for different models on the same observation
may be dependent.

\paragraph{Control all estimation errors.}
Hoeffding's inequality states that the average of $n$ independent
variables in an interval of width $b-a$ satisfies
\begin{equation}
 \Pr\!\left(\left|\overline X-\mathbb{E}[\overline X]\right|>u\right)
 \leq 2\exp\!\left(-\frac{2nu^2}{(b-a)^2}\right),
 \qquad u>0.
\end{equation}
Here $\overline X$ is their sample average. Applying the inequality with
$a=-1$, $b=1$, and
$\varepsilon_n=\sqrt{2\log(2J/\delta)/n}$ gives
\begin{equation}
 \Pr\!\left(|\widehat\Delta_j-\Delta_j|>\varepsilon_n\right)
 \leq2\exp(-n\varepsilon_n^2/2)=\frac{\delta}{J}.
\end{equation}
The union bound over the $J$ candidates therefore gives
\begin{equation}
 \Pr(G)\geq1-\delta,
 \qquad
 G=\bigcap_{j=1}^{J}
 \left\{|\widehat\Delta_j-\Delta_j|\leq\varepsilon_n\right\}.
 \label{eq:proof-confidence-event}
\end{equation}
Dependence between different candidates' estimates does not affect the
union bound.

\paragraph{Verify acceptable selection.}
Approve candidate $j$ when
$\widehat\Delta_j-\varepsilon_n\geq0$. Select a passing
candidate according to a fixed order, or the base if none passes.
On $G$, every passing candidate satisfies
\begin{equation}
 \Delta_j\geq\widehat\Delta_j-\varepsilon_n\geq0.
\end{equation}
Fallback also has nonnegative improvement because $\Delta_0=0$.
Thus the selected model meets the first requirement in
equation~\ref{eq:reliable-reuse} on $G$.

\paragraph{Recognize an available improvement.}
Suppose some candidate has $\Delta_j\geq\gamma$. On $G$, its lower
confidence bound satisfies
\begin{equation}
 \widehat\Delta_j-\varepsilon_n
 \geq\Delta_j-2\varepsilon_n
 \geq\gamma-2\varepsilon_n.
\end{equation}
One copy of $\varepsilon_n$ accounts for possible underestimation of
$\Delta_j$, and the gate subtracts the other copy. The candidate passes
whenever $2\varepsilon_n\leq\gamma$, which is equivalent to
\begin{equation}
 n\geq\frac{8\log(2J/\delta)}{\gamma^2}.
\end{equation}
At least one specialist is then selected. Both requirements hold on the
same event $G$, proving
\begin{equation}
 n^\star\leq
 \left\lceil\frac{8\log(2J/\delta)}{\gamma^2}\right\rceil.
 \label{eq:proof-reuse-upper}
\end{equation}

\paragraph{Construct fixed models and unknown targets.}
To prove a worst-case lower bound, it suffices to construct allowed
problems that every reliable validator must handle. Let an observation
be $Z=(Z_1,\ldots,Z_J)\in\{0,1\}^J$, and fix the losses
\begin{equation}
 \ell(M_0,Z)=\frac12,
 \qquad
 \ell(M_j,Z)=1-Z_j,
 \qquad 1\leq j\leq J.
\end{equation}
Under $P_0$, all coordinates are independent Bernoulli variables with
probability $1/2-\gamma$ of taking the value one. Under $P_j$, only
coordinate $j$ changes that probability to $1/2+\gamma$.
For a coordinate with probability $p$ of being one, the corresponding
specialist has expected loss $1-p$ and improvement $p-1/2$.
Consequently,
\begin{align}
 P_0 &: \quad \Delta_i=-\gamma\quad\text{for }1\leq i\leq J, \notag\\
 P_j &: \quad \Delta_j=\gamma,\qquad
                 \Delta_i=-\gamma\quad\text{for }1\leq i\leq J,\ i\ne j.
\end{align}
The subscript on $P_0$ labels a target distribution, not the base model.
All models and loss functions are identical across these problems.
Only the unknown target distribution differs, so inspecting the models
cannot distinguish the problems. Observing every paired loss reveals
the full vector $Z$ and supplies no information beyond that vector.

\paragraph{Identify decisions that must differ.}
Let $A_j=\{\widehat j=j\}$ be the event that the validator selects
specialist $j$. Write $\Pr_i$ for probability when the sample comes
from $P_i$, including any randomization by the validator. Under $P_0$,
all specialists are harmful. Reliable reuse therefore requires
\begin{equation}
 \sum_{j=1}^{J}\Pr_0(A_j)\leq\delta.
\end{equation}
The events are disjoint because the validator selects one model.
At least one index $j^\star$ satisfies
$\Pr_0(A_{j^\star})\leq\delta/J$; otherwise their sum would exceed
$\delta$. Under $P_{j^\star}$, the only acceptable specialist is
$j^\star$, and its improvement equals $\gamma$. Fallback violates
the recognition requirement, so
\begin{equation}
 p:=\Pr_{j^\star}(A_{j^\star})\geq1-\delta,
 \qquad
 q:=\Pr_0(A_{j^\star})\leq\frac{\delta}{J}.
 \label{eq:proof-decision-probabilities}
\end{equation}

\paragraph{Measure information in the observations.}
For discrete distributions $Q$ and $R$, their KL divergence, also called
relative entropy, is
\begin{equation}
 \operatorname{KL}(Q\Vert R)
 =\sum_y Q(y)\log\frac{Q(y)}{R(y)}.
\end{equation}
Terms with $Q(y)=0$ contribute zero; a term with $Q(y)>0$ and
$R(y)=0$ makes the divergence infinite. KL divergence is nonnegative,
as follows by applying $-\log u\geq1-u$ to $u=R(y)/Q(y)$ and
summing with weights $Q(y)$. For independent coordinates, the log
likelihood ratio is a sum and its expectation is the sum of their KL
divergences. Only coordinate $j^\star$ differs between the two targets.
Writing $\operatorname{kl}$ for the divergence between Bernoulli
distributions gives
\begin{align}
 d_\gamma
 &:=\operatorname{KL}(P_{j^\star}\Vert P_0) \notag\\
 &=\left(\frac12+\gamma\right)
       \log\frac{\frac12+\gamma}{\frac12-\gamma}
   +\left(\frac12-\gamma\right)
       \log\frac{\frac12-\gamma}{\frac12+\gamma} \notag\\
 &=2\gamma\log\frac{1+2\gamma}{1-2\gamma}.
\end{align}
The same sum argument for $n$ independent observations gives
\begin{equation}
 \operatorname{KL}(P_{j^\star}^{\otimes n}\Vert P_0^{\otimes n})
 =n d_\gamma,
\end{equation}
where $P_i^{\otimes n}$ denotes the distribution of the full sample.

Randomizing the validator does not increase the available information.
For a realized sample $z$, let $w(a\mid z)$ be its probability of
selecting index $a\in\{0,\ldots,J\}$. The rule $w$ is the same under
both targets. Let $Q$ and $R$ be the respective joint distributions of
the sample and selected index. Their likelihood ratio cancels $w$,
and summing over $a$ gives
\begin{equation}
 \operatorname{KL}(Q\Vert R)
 =\operatorname{KL}(P_{j^\star}^{\otimes n}\Vert P_0^{\otimes n})
 =n d_\gamma.
 \label{eq:proof-joint-information}
\end{equation}
Deterministic selection is the special case where each $w(a\mid z)$
is either zero or one.

\paragraph{Compare observation and decision information.}
Partition the joint outcomes according to
$A=A_{j^\star}$ and its complement $A^c$. Substituting conditional
probabilities into the definition of KL gives
\begin{align}
 \operatorname{KL}(Q\Vert R)
 ={}&\operatorname{kl}(p,q)
    +p\operatorname{KL}(Q(\cdot\mid A)\Vert R(\cdot\mid A)) \notag\\
   &+(1-p)\operatorname{KL}(Q(\cdot\mid A^c)
                         \Vert R(\cdot\mid A^c)).
\end{align}
Terms with zero weight are interpreted by continuity. Dropping the
nonnegative conditional terms yields
$n d_\gamma\geq\operatorname{kl}(p,q)$. The inequality is the
data-processing inequality for the decision event. Recording only
whether the selected index is $j^\star$ cannot increase KL divergence.
In this construction every sample vector has positive probability under
both targets. Thus $p>0$ implies $q>0$, and the logarithms below are
well defined.

To bound the binary divergence, define the binary entropy
$h(p)=-p\log p-(1-p)\log(1-p)$. Differentiation gives its maximum
$\log2$ at $p=1/2$. Rearranging the binary KL formula gives
\begin{align}
 \operatorname{kl}(p,q)
 &=p\log(1/q)+(1-p)\log(1/(1-q))-h(p) \notag\\
 &\geq p\log(1/q)-\log2 \notag\\
 &\geq(1-\delta)\log(J/\delta)-\log2.
\end{align}
The second term in the first line is nonnegative, and the final line
uses equation~\ref{eq:proof-decision-probabilities}. Since
$\delta\leq1/4$ and $J\geq1$, we have $1-\delta\geq3/4$ and
$\log(J/\delta)\geq\log4=2\log2$. Therefore,
\begin{equation}
 n d_\gamma\geq\frac14\log(J/\delta).
 \label{eq:proof-required-information}
\end{equation}

\paragraph{Convert information into sample size.}
For $0<\gamma\leq1/4$, the inequality $\log(1+u)\leq u$ gives
\begin{align}
 d_\gamma
 &=2\gamma\log\left(1+\frac{4\gamma}{1-2\gamma}\right) \notag\\
 &\leq\frac{8\gamma^2}{1-2\gamma}
 \leq16\gamma^2.
\end{align}
Combining with equation~\ref{eq:proof-required-information} yields
\begin{equation}
 n^\star\geq\frac{\log(J/\delta)}{64\gamma^2}.
 \label{eq:proof-reuse-lower}
\end{equation}
Together, equations~\ref{eq:proof-reuse-upper} and
\ref{eq:proof-reuse-lower} prove Theorem~\ref{thm:local-reuse}.
The bounds have the same dependence on library size, confidence, and
detectable improvement. They are worst-case bounds over bounded-loss
problems, so they do not assert that every particular library requires
that many examples.

\Needspace{8\baselineskip}
\subsection{A validation rule across successive requests}
\label{app:theory-validation}\label{app:theory-application}

The upper-bound construction in Theorem~\ref{thm:local-reuse} supplies
the conditional error control
required by Proposition~\ref{prop:retained-risk}. Its proof applies to
any error tolerance in $(0,1)$. Fix an epoch allowance $0<\delta<1$
and consider a round $k$
that evaluates $J_k\geq1$ fixed candidates on one target family.
Choose a desired detectable improvement $0<\gamma_k\leq1/4$ before
sampling, and allocate
\begin{equation}
 \alpha_k=\frac{\delta}{k(k+1)},
 \qquad
 n_k=\left\lceil\frac{8}{\gamma_k^2}
       \log\frac{2J_k k(k+1)}{\delta}\right\rceil.
 \label{eq:proof-round-design}
\end{equation}
The identity $1/(k(k+1))=1/k-1/(k+1)$ gives
$\sum_{k\geq1}\alpha_k=\delta$. Conditional on the complete prior
history, draw $n_k$ fresh independent examples from the declared target
distribution. Use the confidence-bound gate in Appendix~\ref{app:theory-reuse} with $J=J_k$ and
$\delta=\alpha_k$. The upper-bound proof controls every passing
candidate at once, so $\Pr(B_k\mid\mathcal{G}_{k-1})\leq\alpha_k$.
It also ensures that a specialist is selected whenever an available
candidate improves by at least $\gamma_k$, on the same confidence
event. A round that covers several families must include all issued
credentials in its conditional error bound, for example by allocating
separate allowances to the family evaluations.

The sample size in equation~\ref{eq:proof-round-design} is sufficient
for recognition as well as acceptable selection. The confidence-bound
gate controls false approval for any fixed positive sample size, but a
smaller sample may cause fallback even when useful expertise exists.
Evaluating all models on $n_k$ observations can require
$(J_k+1)n_k$ model executions. The theorem therefore does not provide a
logarithmic bound on runtime or guarantee that validation is cheaper
than a particular training procedure.

\paragraph{Relationship to the recorded experiments.}
The experimental contracts in Appendix~\ref{app:domain-contracts}
implement the same separation between retention, admission, and reuse,
but use different statistical gates. Navigation requires an observed
gain of at least 15 successes on 100 paired cases. Placement combines
a reduction in median HPWL with a signed-rank test, with multiplicity
handling for extension exams. Neither rule is the bounded-mean
confidence gate analyzed above. An application to HPWL would need a
fixed bounded loss and a test for its expected difference. Improvement
in a median does not by itself establish improvement in an expectation.

The sampling conditions also require explicit verification. A navigation
family includes settings such as scene size and pedestrian count
(Appendix~\ref{app:terminology}). Its $P_f$ must specify their
distribution together with episode randomness. Evaluation at one
setting does not establish the guarantee for other settings with the
same family key. Appendix~\ref{sec:audit-blocks} records overlap between
the permitted training and admission index ranges in the original
navigation runs. The overlap does not prove training exposure, but
disjointness and conditional independence have not been established by
that record. A retrospective service subset cannot establish validity
of the original admission decision. Finally, correction within an
extension exam does not by itself allocate error across all exams in an
epoch. The results above give conditions for a population guarantee;
they do not retrospectively certify the empirical gates or promise the
same realized gain on each later request.

\clearpage
\section{Recorded macro-placement examples}
\label{app:placement-examples}
\begingroup
\setcounter{figure}{0}
\renewcommand{\thefigure}{\thesection\arabic{figure}}
\renewcommand{\theHfigure}{appendix.\thesection.\arabic{figure}}

Figures~\ref{fig:placement-examples-ibm} and~\ref{fig:placement-examples-mixed}
show archived rollouts for the eight circuits in the revisit study.
Each row compares Random, the frozen base, the final shared Overwrite
checkpoint, and a saved \sys{} specialist or candidate using the same
recorded seed. Each row gives the circuit's total loaded macro count;
this is not the number of rectangles displayed. The macro inventory
shows exported macro shapes in a
schematic arrangement, rather than an initial placement. Placement panels
share the circuit's canvas scale; colors encode log macro area and identify
the same macro across methods, while gray marks fixed nodes and ports.
Overwrite uses the same final checkpoint throughout, rather than the
successive checkpoints used to serve the request stream.

\begin{figure}[H]
    \centering
    \includegraphics[page=1,width=\linewidth]{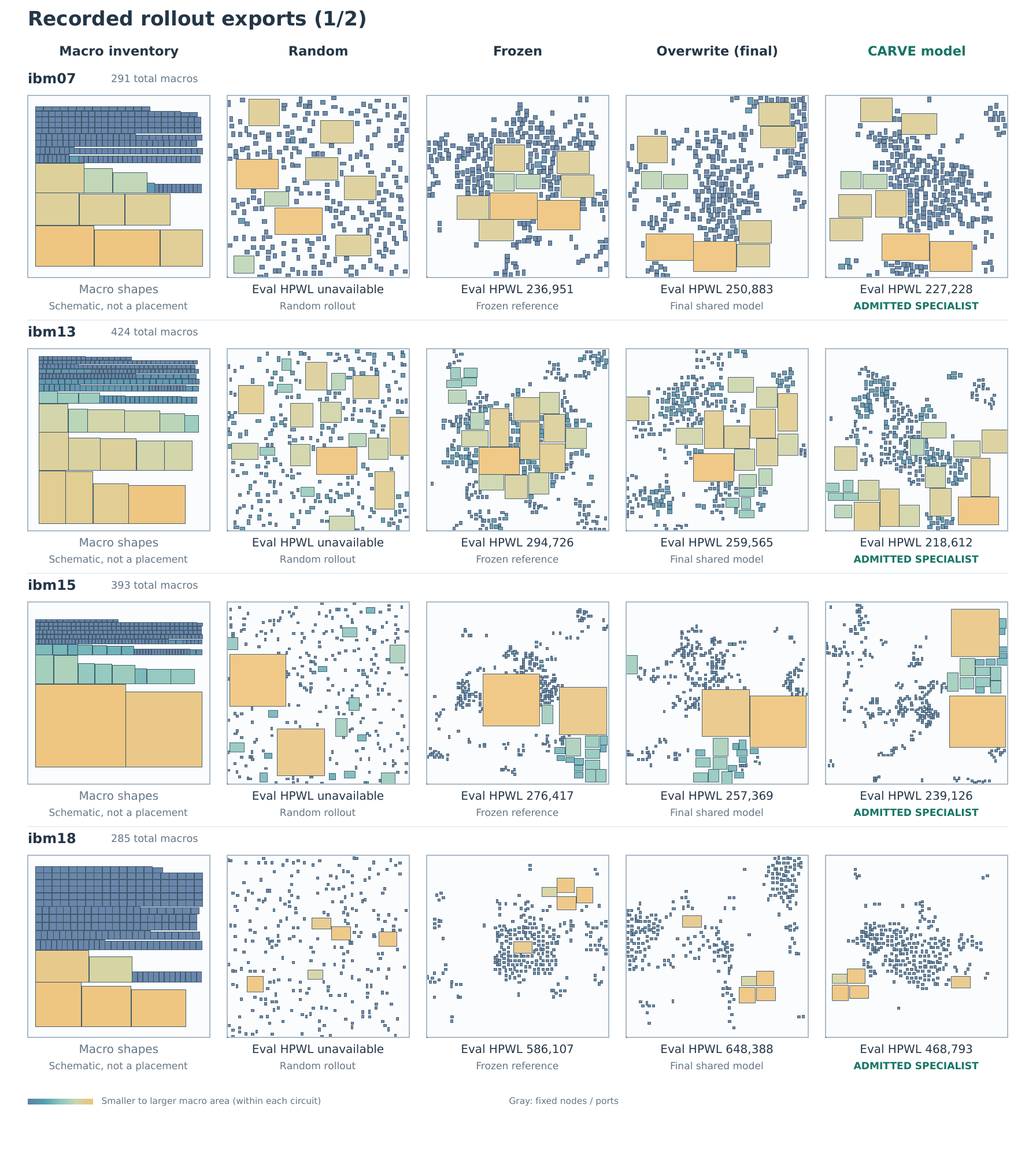}
    \caption{Recorded rollouts for ibm07, ibm13, ibm15, and ibm18.
    Row labels give circuit totals; panels show exported subsets.
    Eval HPWL comes from scoring records. The \sys{} column shows admitted specialists.}
    \label{fig:placement-examples-ibm}
\end{figure}

\clearpage
The adaptec3 row reports the same circuit total as
Figure~\ref{fig:chip-layouts-stream}. The panels show an exported subset
from a different run. Only the bigblue1 model exports include all loaded macros. The archived evaluator continues
the initial rollout for adaptec2 and adaptec3 before computing HPWL;
IBM scoring remains limited to the placed subset.

Eval HPWL comes from the scoring records, not a recomputation from the
displayed exports. Table~\ref{tab:appendix-A12} reports 30-trial medians.
Admission labels come from separate evaluations. The two adaptec
candidates were rejected and not deployed. Their Random exports are
incomplete and have no evaluation HPWL.

Recorded overlaps are retained and shapes are clipped at the canvas.
A full rollout means that all loaded macros are represented; it does not
establish legality. The archived evaluator has no final overlap or boundary
check. The reported HPWL gains therefore describe the archived evaluation
protocol, and full-placement legality remains unverified.

\begin{figure}[H]
    \centering
    \includegraphics[page=2,width=\linewidth]{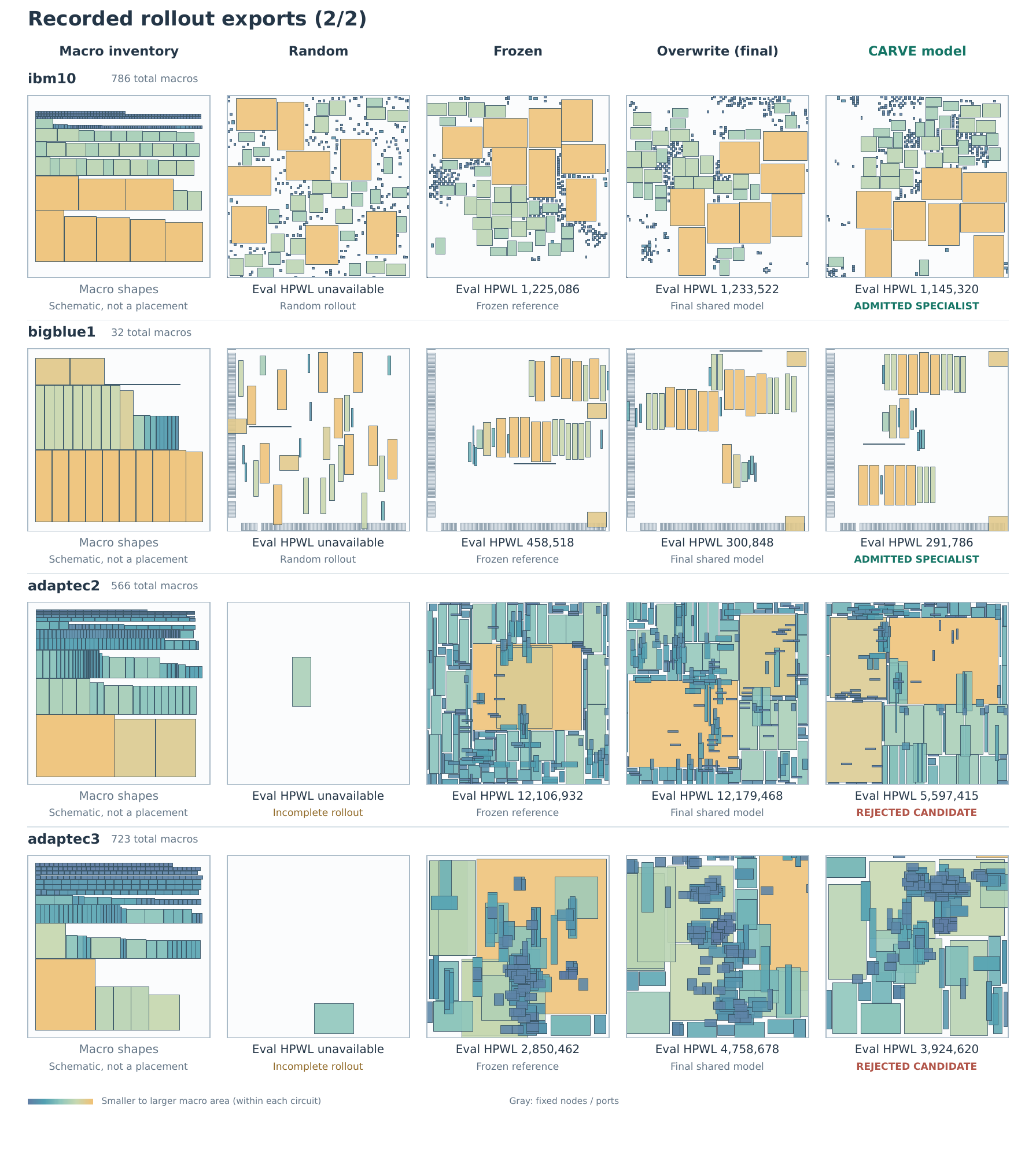}
    \caption{Recorded rollouts for ibm10, bigblue1, adaptec2, and adaptec3.
    Row labels give circuit totals. The ibm10 and bigblue1 models are admitted specialists; the two
    adaptec models are rejected candidates shown for inspection.
    The incomplete Random rollouts are explicitly marked.}
    \label{fig:placement-examples-mixed}
\end{figure}
\endgroup
\end{document}